%% file: conditioned_dfa_arxiv.tex
\documentclass{article}
\def\ARXIV{1}
\input{paper_preamble}
\begin{document}
\input{paper_body}

\end{document}

%% file: paper_preamble.tex
\ifdefined\ARXIV
\usepackage[letterpaper,margin=1in]{geometry}
\usepackage{times}
\usepackage{natbib}
\else
\usepackage{iclr2027_conference,times}
\fi
\usepackage{amsmath,amssymb,mathtools,bm}
\usepackage{booktabs}
\usepackage{placeins}
\usepackage{graphicx}
\usepackage{hyperref}
\usepackage{xcolor}
\usepackage{microtype}
\ifdefined\ARXIV
\hypersetup{
    hidelinks,
    pdftitle={Conditioned Direct Feedback Alignment via Activity and Error Geometry},
    pdfauthor={Houman Safaai, Varun Reddy, Bernardo L. Sabatini}
}
\else
\hypersetup{
    hidelinks,
    pdftitle={Conditioned Direct Feedback Alignment via Activity and Error Geometry},
    pdfauthor={}
}
\fi

\graphicspath{{figures/}{./}}

\newcommand{\E}{\mathbb{E}}
\newcommand{\R}{\mathbb{R}}
\newcommand{\Loss}{\mathcal{L}}
\newcommand{\Cov}{\operatorname{Cov}}
\newcommand{\tr}{\operatorname{tr}}
\newcommand{\diag}{\operatorname{diag}}

\title{Conditioned Direct Feedback Alignment\\via Activity and Error Geometry}

\ifdefined\ARXIV
\author{%
Houman Safaai$^{1}$\thanks{Correspondence: \href{mailto:houman_safaai@harvard.edu}{houman\_safaai@harvard.edu}} \quad
Varun Reddy$^{1}$ \quad Bernardo L.\ Sabatini$^{1,2}$ \\[0.4em]
{\normalfont\small $^{1}$Kempner Institute for the Study of Natural and Artificial Intelligence at Harvard University} \\
{\normalfont\small $^{2}$Howard Hughes Medical Institute; Department of Neurobiology, Harvard Medical School}%
}
\date{}
\else
\fi


\ifdefined\ARXIV
\renewcommand{\topfraction}{0.9}
\renewcommand{\textfraction}{0.1}
\fi

%% file: paper_body.tex
\maketitle

\begin{abstract}
Direct feedback alignment (DFA) trains hidden layers with fixed random
projections of the output error. Even when this feedback provides useful
credit, unequal scales across activity or error directions can distort the
local update. We study conditioned DFA (nDFA), a family that adapts established
inverse-moment preconditioning to either side of this update. An aligned
linear analysis describes how activity conditioning changes spectral learning
rates and early-stopping risk. Synthetic experiments show gains from both
stabilization and changes in update direction. Confirmatory CIFAR-10
experiments show that activity conditioning, including its FOOF formulation,
improves tuned DFA at matched measured training time. Controls support a role
for centered correlations beyond the tested mean and diagonal alternatives,
while early-only conditioning retains most of the benefit with less training
work. Conditioning also benefits backpropagation, and timing interventions
do not establish a mechanism specific to random feedback. Error conditioning
improves short-budget models and changes update directions, but its small
additional improvement under stable training does not survive correction for
multiple comparisons. Predicted benefits on image-background benchmarks are
not confirmed. These findings establish practical benefits and important
limits of conditioning learning with fixed random feedback.
\end{abstract}

\section{Introduction}

Credit assignment asks how a synapse should change when an outcome depends
on many neurons. Backpropagation (BP) assigns this credit by propagating output errors through the
transpose of each forward weight matrix. Feedback alignment (FA) and direct
feedback alignment (DFA) relax this requirement by replacing transported errors
with fixed random feedback signals \citep{lillicrap2016feedback,nokland2016dfa}.
These rules connect to biological credit-assignment models, including segregated
dendritic compartments \citep{lillicrap2020brain,guerguiev2017dendrites,sacramento2018dendritic}.
Their learning can nevertheless be brittle: their
success depends on whether the random feedback produces a usable hidden-layer
update \citep{bartunov2018scalability,launay2020dfa,garg2022randomfeedback}.

Many analyses of DFA focus on the feedback pathway: its rank, scale,
alignment, and compatibility with the network architecture. Here we study a complementary
failure mode that can occur even when feedback provides a useful descent component: the
local DFA update is an outer product with the presynaptic activity, so
high-variance activity directions that carry no task information can dominate
the update with nuisance structure rather than useful credit.

We ask whether reweighting directions in the local activity and error signals
improves learning while leaving feedback fixed. We call this reweighting
conditioning. Anisotropy means unequal second-moment energy across directions.
Activity anisotropy weights combinations of presynaptic inputs; error
anisotropy weights combinations of the teaching signals delivered to hidden
units. Conditioning can act on either side of the local outer product.

Our contribution is a controlled study of established conditioning operators
under imperfect fixed feedback. The activity operator is the Fast First-Order Optimizer (FOOF) update
applied to DFA errors \citep{benzing2022foof}; its matrix inverse is not new.
We separate activity and error factors, test which activity statistics matter,
and compare conditioning schedules against tuned baselines at matched measured
work. A joint confirmation supports activity conditioning and an early-only
efficiency tradeoff, but limits the additional error-side benefit and the
transfer of synthetic nuisance predictions. Distinguishing stabilization,
faster optimization and improved generalization is central to these conclusions.

\section{Conditioned DFA}
\label{sec:method}

We now turn the geometric failure mode into a learning rule.
For layer $\ell$, let $W_\ell$ be its weight matrix, $h_{\ell-1}$ its input
activity, $a_\ell$ its preactivation, and $\phi$ its nonlinearity.
The output error $e$ is the derivative of the per-example loss with respect
to the output preactivation. A fixed matrix $B_\ell$ maps this error to the
hidden-layer dimension. For an elementwise nonlinearity, DFA uses
\begin{equation}
    \delta_\ell^{\mathrm{DFA}}
    =
    (B_\ell e)\odot \phi'(a_\ell),
    \qquad
    G_\ell^{\mathrm{DFA}}
    =
    \delta_\ell^{\mathrm{DFA}}h_{\ell-1}^{\top}.
    \label{eq:dfa}
\end{equation}
Here $\odot$ denotes elementwise multiplication and $\eta>0$ the learning
rate. The raw SGD step is $-\eta G_\ell^{\mathrm{DFA}}$; minibatches average
these outer products. Batch-normalized layers also apply the
local normalization derivative.

\paragraph{Conditioning the local outer product.}
The minibatch update combines local errors with presynaptic activities.
To condition either factor, we estimate the uncentered moments
\begin{equation}
    C_{A,\ell}=\E[h_{\ell-1}h_{\ell-1}^{\top}],
    \qquad
    C_{E,\ell}=\E[\delta_\ell^{\mathrm{DFA}}
                            \delta_\ell^{\mathrm{DFA}\top}],
    \label{eq:secondmoments}
\end{equation}
where the expectation averages examples, including means as well as
fluctuations. Separate damping $\lambda_A,\lambda_E>0$ stabilizes each
inverse.\footnote{Terminology: except when
explicitly discussing the power-$1/2$ ZCA (zero-phase component analysis)
whitening control, ``inverse-second-moment preconditioning'' refers to the
power-$1$ updates below. ``Activity/input side'' refers to right multiplication
in weight-gradient coordinates; ``error/output side'' refers to left
multiplication.}
With $I$ denoting the identity matrix, define
\[
    P_{A,\ell}=(C_{A,\ell}+\lambda_A I)^{-1},
    \qquad
    P_{E,\ell}=(C_{E,\ell}+\lambda_E I)^{-1}.
\]
The conditioned-DFA family defines the pseudo-gradients
\begin{equation}
\begin{aligned}
    U_\ell^{A\text{-}\mathrm{nDFA}}
        &= G_\ell^{\mathrm{DFA}}P_{A,\ell},\\
    U_\ell^{E\text{-}\mathrm{nDFA}}
        &= P_{E,\ell}G_\ell^{\mathrm{DFA}},\\
    U_\ell^{K\text{-}\mathrm{nDFA}}
        &= P_{E,\ell}G_\ell^{\mathrm{DFA}}P_{A,\ell}.
\end{aligned}
    \label{eq:ndfa}
\end{equation}
For norm-matched comparisons, rescale the selected pseudo-gradient $U_\ell$ as
\begin{equation}
\widetilde G_\ell=
\begin{cases}
\|G_\ell^{\mathrm{DFA}}\|_F\,U_\ell/\|U_\ell\|_F,&\|U_\ell\|_F>0,\\
0,&\|U_\ell\|_F=0.
\end{cases}
\label{eq:normmatched}
\end{equation}
Plain SGD uses $\Delta W_\ell=-\eta\widetilde G_\ell$, or $-\eta U_\ell$ without
norm matching. In AdamW \citep{loshchilov2019adamw}, this matrix replaces the
hidden-weight gradient before the optimizer's moment updates. Norm matching
and direction diagnostics concern this pre-optimizer matrix, not the final
parameter step. Bias, classifier and normalization parameters retain the
underlying learner's updates.
We use ``nDFA'' without a prefix for the activity-only rule, and ``K-nDFA'' for
the two-sided Kronecker form; A, E, and K abbreviate these three rules.
The ``n'' denotes inverse-second-moment
normalization, not an exact natural-gradient method. All three updates are
block-local after output-error broadcast: they require minibatch-wide layer
statistics and damped solves, but not backpropagated hidden errors. They are not
synaptically local. A direct dense solve requires $O(d^2)$ state and $O(d^3)$ factorization.
When a minibatch has fewer examples than units, exact sample-space identities
replace this solve by a smaller Gram-system solve without changing the update
(Appendix~\ref{app:sample_space}); K-nDFA still applies both factors.
The activity rule shares FOOF's core operator \citep{benzing2022foof}; here
the error source is DFA and moment estimation and norm matching follow the
stated protocol. The two-sided form resembles KFAC, but a DFA pseudo-error
moment is not a Fisher factor.

\paragraph{Candidate sources of benefit.}
Activity conditioning is relevant when high-variance presynaptic directions are
nuisance-dominated. Error conditioning targets a different pathology: unequal
second-moment energy across local credit coordinates, which can arise from class imbalance,
unequal residual scales, saturation/gating, or anisotropic random-feedback
routing even when the input is not noisy. Combining the factors may help when
each improves the task-relevant part of the update. Estimating $C_E$ is especially
fragile because gated DFA errors can be low-rank; under-damped inversion can
amplify poorly estimated directions. The rule uses layer statistics without nuisance labels or a known task
subspace. Equal algebraic roles do not imply equal damping or robustness.

\paragraph{Error scaling and update magnitude.}
Error moments use per-example signals. A local error computed from a mean
minibatch loss is multiplied by the batch size before forming its moment.
The factor-separated comparisons use these per-example moments, independently
validation-selected damping, and the layerwise norm matching in Eq.~\ref{eq:normmatched}.

\subsection{Linearized analysis: inverse-second-moment conditioning}
\label{sec:linearized}
To isolate the input-covariance effect, take Gaussian inputs
$x\sim\mathcal N(0,\Sigma)$ with $\Sigma\succ0$, target $W^\star x$,
hidden and output weights $W^{(1)},W^{(2)}$, and fixed feedback $B$.
The residual operator is $\varepsilon=W^{(2)}W^{(1)}-W^\star$; here the presynaptic second moment
$C_{A,\ell}$ in Eq.~\ref{eq:secondmoments} specializes to the input covariance
$\Sigma$. The expected first-layer updates are
$g^{\mathrm{BP}}=-(W^{(2)})^\top\varepsilon\Sigma$,
$g^{\mathrm{DFA}}=-B\varepsilon\Sigma$, and
$g^{\mathrm{nDFA}}=-B\varepsilon\Sigma(\Sigma+\lambda_A I)^{-1}$.
Here $g$ denotes a descending update per unit learning rate, whereas $G$ in
Eq.~\ref{eq:dfa} is gradient-like: the corresponding direction is $g=-G$.
The linear-Gaussian loss Hessian has Kronecker form
$(W^{(2)\top}W^{(2)})\otimes\Sigma$. Right-multiplication by
$(\Sigma+\lambda_A I)^{-1}$ is the damped input-side inverse-second-moment
factor used by Kronecker-factored approximate curvature (KFAC)
and related Newton or natural-gradient approximations
\citep{amari1998natural,martens2015kfac}. In the nonlinear experiments below,
we use this curvature-motivated preconditioner on the DFA update. The analogous
factorization motivates separate left and right inverses, but is not an exact Fisher
identity: DFA errors depend on random feedback, gating and residuals
(Appendix~\ref{app:theory}).

\begin{figure}[t]
    \centering
    \includegraphics[width=\textwidth]{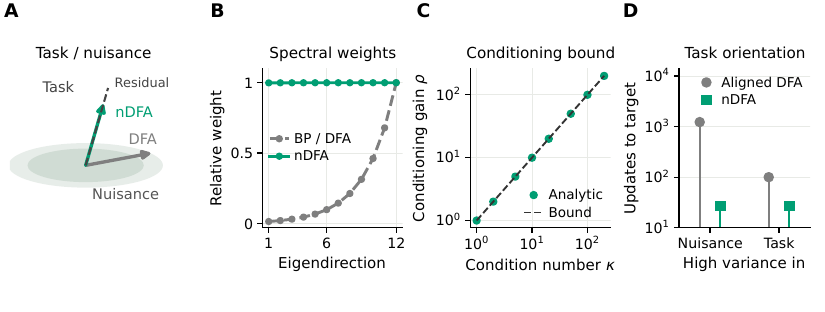}
    \caption{\textbf{Conditioning reweights a local update.}
    \textbf{A,} A mixed residual (dashed direction) has task and nuisance components. An anisotropic
    activity moment weights its nuisance component strongly; inverse-moment
    conditioning reduces that imbalance. \textbf{B,} In the aligned linear
    model, raw BP/DFA weight eigendirections by $\lambda_i$, while activity
    conditioning gives $\lambda_i/(\lambda_i+\lambda_A)$ (relative weights shown).
    \textbf{C,} The analytic gain in input conditioning,
    $\rho=\kappa(\Sigma)/\kappa_{\mathrm{nDFA}}$, approaches its bound
    $\kappa(\Sigma)$ under weak damping. Panels B--C use $\lambda_A=10^{-3}$
    and $\lambda_{\min}=1$. \textbf{D,} Population linear simulations at condition
    number 50 and fixed alignment, averaged over three realizations. With separately selected learning rates,
    updates to a target loss depend on task orientation (Appendix~\ref{app:linear_simulation}).}
    \label{fig:theory_conditioning}
\end{figure}

\paragraph{Spectral prediction.}
In the post-alignment linear-Gaussian regime with population second moments and
damping $\lambda_A>0$, let $\lambda_i$ be the eigenvalues of $\Sigma$.
nDFA replaces the per-eigendirection input factor $\lambda_i$ by $\lambda_i/(\lambda_i+\lambda_A)$. The associated input-side
condition number is
\[
    \kappa_{\mathrm{nDFA}}
    =
    \kappa(\Sigma)\,
    \frac{\lambda_{\min}+\lambda_A}{\lambda_{\max}+\lambda_A},
\]
where $\kappa(\Sigma)=\lambda_{\max}/\lambda_{\min}$ and the extrema
refer to the input eigenvalues (Fig.~\ref{fig:theory_conditioning}B--C).
Realized learning also depends on the residual, initialization, step size,
sampling noise and task orientation. Our empirical hypothesis is narrower:
nDFA should help most when otherwise active high-variance directions are
task-irrelevant nuisance, should be redundant under isotropy, and should offer
little advantage over a well-tuned exact-gradient method when high-variance
directions carry the task. The analogous left-side prediction is conditional:
error nDFA may help when high-variance local-error directions obscure useful
credit, but can hurt when those directions carry the task or when $C_E$ is
poorly estimated or under-damped. Figure~\ref{fig:theory_conditioning}A
illustrates the geometry; panel D shows why the realized gain depends on
task orientation even at the same condition number.

Appendix~\ref{app:theory} gives the exact input-factor identity and alignment
assumptions. Its separate two-block regression calculation illustrates
how re-timing task and nuisance fitting changes early-stopping risk, a
mechanism studied previously for preconditioned and differently scaled
gradient dynamics \citep{amari2021preconditioning,heckel2021earlystopping}.
It does not explain the acquisition of DFA alignment or establish a new
general principle of early stopping.

A complementary construction holds the raw update fixed while varying
activity and error orientations (Appendix~\ref{app:factor_mechanism}). It
isolates how either factor can improve or worsen approximate credit at the
same moment spectrum (Appendix Fig.~\ref{fig:factor_mechanism}A--C),
extending the geometric question beyond the aligned input-side model.

\section{Experimental setup}
\label{sec:setup}

The experiments distinguish directional effects from stabilization, then test
whether those effects remain useful with stronger baselines. An eight-class
synthetic suite varies task geometry, label noise and sample size. Historical
MNIST/Fashion-MNIST \citep{xiao2017fashion} studies separate the activity and error factors at short
training horizons. Their protocols and test-selection limitations remain
explicit (Appendices~\ref{app:synthetic}--\ref{app:threefactor}).

The primary practical study uses CIFAR-10 \citep{krizhevsky2009tiny} ReLU MLPs with batch
normalization (BN) in the core arms, per-method validation selection and equal measured
learning time on the same accelerator. It compares tuned DFA and BP,
forward-decorrelated DFA (FD-DFA), activity conditioning, early-only activity
conditioning, and FOOF with either credit rule. FOOF uses averaged activity
moments and periodic inverse updates without norm matching; A uses current
minibatch moments and Eq.~\ref{eq:normmatched}. Separate interventions examine
centered covariance, conditioning windows at constant learning rate, and
temporary error conditioning. Image-background tasks test the proposed
transfer from synthetic nuisance geometry to real images.

After development, a joint protocol fixes settings, endpoints and ten paired
training seeds. Work-budget studies select the checkpoint with lowest
validation cross-entropy (CE); fixed-epoch interventions use the final
checkpoint. All confirmation training completes before joint test evaluation.
The standard CIFAR/MNIST/Fashion test sets had been used in earlier project
studies; new seeds do not make these datasets previously unseen. The primary
family contains 21 paired tests with global Holm correction. Intervals are
individual Student $t$ intervals over training seeds, conditional on the data
split and selection procedure. Appendix~\ref{app:joint_confirmation} reports
every contrast, protocol deviation and secondary analysis. Historical
experiments retain their original replication units and analysis families.

\section{Results}

\subsection{Synthetic controls separate stabilization from geometry}
\label{sec:results_synthetic}

\begin{figure}[t]
    \centering
    \includegraphics[width=\textwidth]{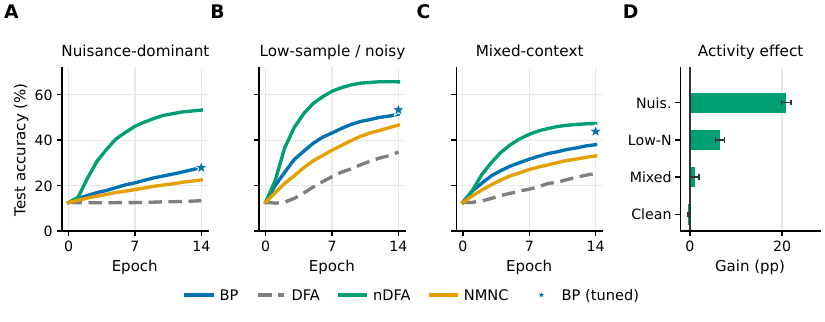}
    \caption{\textbf{Conditioned DFA and the activity-side regime.}
    \textbf{A--C,} Fixed-full-rank nDFA separates from raw DFA in the three
    stressed regimes; stars mark BP endpoints from a separate per-regime
    learning-rate search selected on test accuracy. Raw DFA's shared-rate
    high-loss instability contributes to the displayed gaps.
    The 14-epoch curves average the fixed designed grid; inferential bands are omitted.
    NMNC: neural-manifold noise correlation \citep{kang2026nmnc}, a perturbation-based local rule.
    \textbf{D,} Activity-only nDFA minus DFA after 100 epochs in the focused
    nuisance, low-sample/noisy, mixed and clean task-aligned cells:
    $+20.9$, $+6.5$, $+1.1$ and $-0.3$ percentage points (pp), respectively.
    Bars show mean $\pm$ SEM over five data seeds, each averaging three feedback seeds.}
    \label{fig:rule_positive}
\end{figure}

The synthetic suite tests the predicted dependence on whether high-variance
directions carry task signal or nuisance (Fig.~\ref{fig:rule_positive}A--C;
per-method results in Table~\ref{tab:infodfa_synthetic_noise_split}).

The norm-matched controls provide the clearest evidence for directional
benefit beyond update scale (Fig.~\ref{fig:controls_composite}A). The broader
shared-rate sweep also strongly favors nDFA, but raw DFA has extreme finite
training losses in nearly all stressed runs. Its large accuracy gaps therefore
combine stabilization with any change in useful credit
(Table~\ref{tab:infodfa_synthetic_stability}). All five global seeds favor nDFA
after averaging the fixed grid. These comparisons do not establish an
advantage over individually tuned, near-converged learners.

The focused factor ablation shows an activity-side benefit in the
nuisance-dominant and low-sample/noisy cells (Fig.~\ref{fig:rule_positive}D).

In the short-horizon task-aligned sweep, conditioning improves DFA but
tuned BP remains stronger. The focused 100-epoch clean, norm-matched control
shows no activity benefit (Table~\ref{tab:infodfa_normmatch_factor_controls}).
Thus the broad short-horizon gain should not be read as a sustained
generalization advantage when high-variance directions carry task signal.

In noisy-label vision MLPs, conditioning lifts raw DFA close to BP on
Fashion-MNIST and CIFAR-10 (Table~\ref{tab:infodfa_vision_noise_split}). These
sweeps use fixed full-rank feedback. Their shared-rate, short-horizon
comparisons do not establish parity with individually tuned BP.

On a ColoredMNIST-style task \citep{arjovsky2019irm}, nDFA has lower seed variability
than DFA and approaches BP under a reversed color--label correlation. The
paired accuracy gain does not survive Holm correction at eight runs,
supporting a descriptive stability result rather than superiority
(Appendix~\ref{app:coloredmnist}).

Across cells where raw DFA learns, conditioning also reduces variation across
compound feedback/order seeds (Appendix~\ref{app:feedback_variance}). This is
a descriptive stability observation: feedback and minibatch order vary
together. The early alignment trajectories do not isolate an alignment-rescue
mechanism, and the controlled timing tests below do not establish DFA
specificity.

\subsection{Error conditioning: short-budget gains and a stable-training limit}
\label{sec:results_error_side}

\begin{figure}[t]
    \centering
    \includegraphics[width=\textwidth]{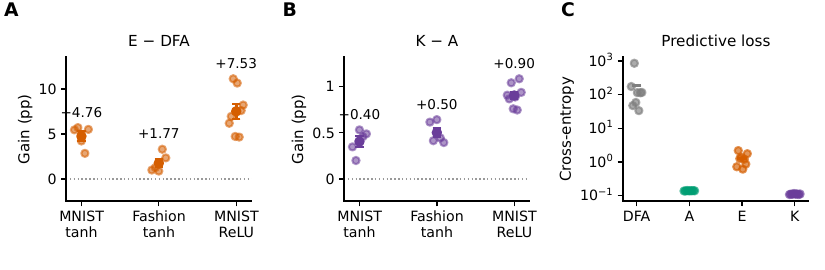}
    \caption{\textbf{Historical short-budget error-side gains.}
    \textbf{A,} E-nDFA improves raw DFA in both tanh studies and the
    ReLU/softmax confirmation. \textbf{B,} Adding the error factor
    after activity conditioning improves all seed-level pairs in all three
    settings. Dots are model/data-order seeds after averaging three feedback
    seeds; large points are mean $\pm$ SEM. \textbf{C,} Final test cross-entropy
    in the ReLU/softmax confirmation (log scale); dots show the eight seed means
    and horizontal marks their average. These fixed-rate endpoints do not
    establish a stable-training advantage (Appendix~\ref{app:joint_error}). Original
    MNIST test exposure is disclosed in Appendix~\ref{app:threefactor}.}
    \label{fig:error_kndfa_replication}
\end{figure}

The left factor changes a different part of the local update, so an activity
benefit does not establish an error benefit. Historical tanh MNIST and
Fashion-MNIST experiments select the two damping values independently and
combine them without a joint search. Error conditioning improves raw DFA,
and adding it to A improves the short-budget endpoint
(Fig.~\ref{fig:error_kndfa_replication}). A ReLU/softmax MNIST replication
preserves these effects. All conditioned hidden matrices are norm-matched
to raw DFA. These fixed-rate studies test early optimization, not superiority
to tuned, converged baselines. The original MNIST seeds had already been
evaluated on test data before validation development; the separate Fashion
and ReLU protocols and later post-hoc extensions are distinguished in
Appendix~\ref{app:threefactor}.

The stable-training CIFAR-10 confirmation gives a more limited result.
Adding E during the first quarter of training changes update direction
substantially and has small favorable mean accuracy and loss effects, but
neither survives the global correction. It also fails the required
performance comparisons with diagonal E. Thus it fails the prespecified
five-part criterion for an additional error-side benefit. The complete
result and scalar-control identity are reported in
Appendix~\ref{app:joint_error}; the failure is not evidence that E is exactly
ineffective.

\subsection{Controls separate scale, covariance, and normalization}

\label{sec:results_controls}

We next test regularization, scalar rescaling and diagonal adaptation as
alternative explanations. On the nuisance-dominant synthetic cell,
regularization and early stopping provide much smaller BP gains than activity
conditioning (Table~\ref{tab:infodfa_controls}). Matching raw DFA's hidden
gradient norms to BP does not recover the benefit. On noisy-label Fashion-MNIST,
however, this scalar correction reaches slightly higher mean accuracy than
nDFA (Table~\ref{tab:infodfa_controls}): scale can suffice when the local
direction is already useful. BP norms are used only in
this diagnostic; the factor confirmations match raw DFA.

\begin{figure}[t]
    \centering
    \includegraphics[width=\textwidth]{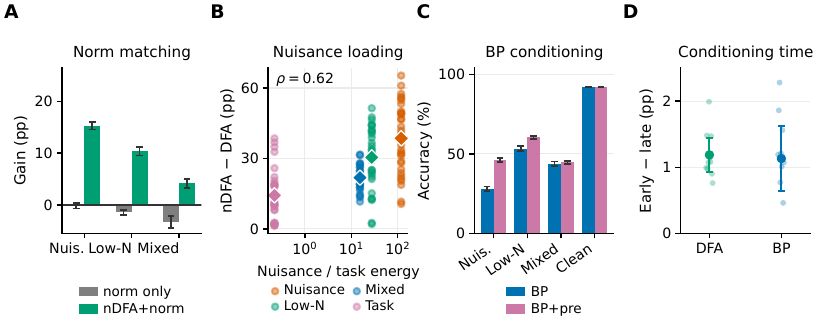}
    \caption{\textbf{Directional controls and conditioning time.}
    \textbf{A,} Activity gains survive instantaneous BP-norm matching of
    hidden updates (five synthetic training seeds, mean $\pm$ SEM).
    \textbf{B,} Gain versus designed nuisance/task energy in the exploratory
    rank-selected sweep (diamonds: regime means; descriptive).
    \textbf{C,} The same activity operator also helps BP in stressed synthetic
    conditions (mean $\pm$ SEM over five global seeds after test-based
    learning-rate selection; descriptive). \textbf{D,} Early-minus-late conditioning at constant learning
    rate helps both DFA and BP on CIFAR-10; their interaction is not
    significant. Dots show ten paired training seeds and bars individual
    95\% $t$ intervals. Panels A--C are historical controls; D is the joint
    confirmation.}
    \label{fig:controls_composite}
\end{figure}

Applying the same BP-norm matching \emph{after} conditioning preserves the
activity-side advantage in all three hard cells. The supported effect is
therefore directional reweighting beyond a scalar step-size change
(Fig.~\ref{fig:controls_composite}A).
Across the designed cells, the gain also grows with nuisance energy relative
to task energy (Fig.~\ref{fig:controls_composite}B); this is a descriptive
association, not an independently validated predictor.

\paragraph{Is BatchNorm already the fix?}
BatchNorm standardizes per-unit activity \citep{ioffe2015batchnorm}.
Raw DFA+BN matches or exceeds unnormalized A-nDFA in the low-sample/noisy
and mixed regimes, but falls short in nuisance-dominant cells
(Tables~\ref{tab:infodfa_synthetic_noise_split} and \ref{tab:infodfa_bn_baseline}).
Within the normalized recipe, adding activity conditioning helps in the nuisance and low-sample
regimes but hurts in the mixed and task-aligned regimes
(Table~\ref{tab:infodfa_bn_baseline}).

\paragraph{Inverse versus inverse-square-root conditioning.}
To distinguish inverse-moment conditioning from whitening
\citep{desjardins2015naturalnn,huang2018decorrelated}, a matched control on
the same 128-cell suite preconditions the DFA update by
$(C_{\ell-1}+\lambda I)^{-1/2}$ instead of nDFA's full inverse
(Table~\ref{tab:infodfa_bpwhiten}). Inverse-square-root conditioning captures
most of the input-side gain. The full inverse adds
more in the nuisance-dominant and low-sample/noisy regimes than in the
mixed-context or clean task-aligned controls. This is consistent with the
spectral motivation, but all comparisons come from the same designed suite.

\begin{figure}[t]
\centering
\includegraphics[width=\textwidth]{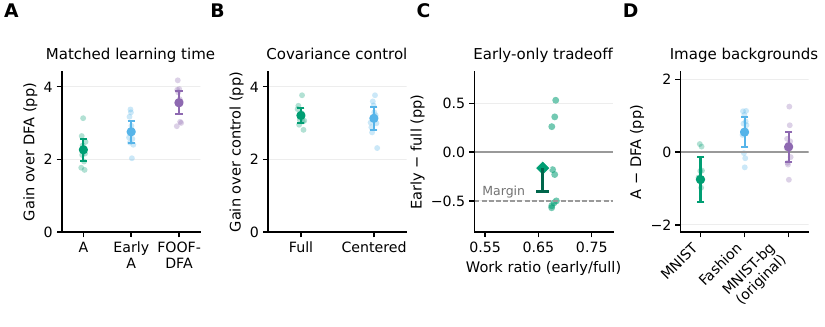}
\caption{\textbf{Confirmed benefits and their limits.}
\textbf{A,} Activity, early activity and FOOF-DFA improve tuned DFA at matched
measured learning time on CIFAR-10. \textbf{B,} In a fixed-epoch experiment,
full and centered conditioning outperform the mean-preserving diagonal-covariance control.
\textbf{C,} In a separate fixed-epoch cohort, early conditioning retains
accuracy within the prespecified margin while reducing measured learning work.
The vertical segment marks the mean's one-sided 95\% lower bound, above the
$-0.5$\,pp margin; it is not a criterion for each seed.
\textbf{D,} Activity-minus-DFA accuracy effects on constructed MNIST/Fashion
backgrounds and the original MNIST-background-images task do not establish
the predicted gains. Points show ten paired seed contrasts; bars show means
and individual, unadjusted 95\% $t$ intervals in A, B and D. C shows seed-level work ratios and accuracy
differences, with the mean marked separately. Superiority tests in A, B and D
use the Holm-adjusted comparisons in Table~\ref{tab:joint_primary}. C uses the
separately prespecified noninferiority and work-reduction criteria in
Appendix~\ref{app:joint_criteria}.}
\label{fig:bn_optimizer_main}
\end{figure}

\paragraph{Centered correlations beyond mean activity.}
Full uncentered moments combine covariance and the activity mean:
$C_A=\Cov(h)+\mu\mu^\top$, where $\mu=\E[h]$. A full-versus-diagonal
comparison alone cannot separate these contributions. We therefore compare
full and centered conditioning with a control that preserves the mean outer
product but removes off-diagonal covariance,
$\diag(\Cov(h))+\mu\mu^\top$. Both richer operators outperform this control
in every confirmation seed, with corrected evidence for both contrasts
(Fig.~\ref{fig:bn_optimizer_main}B). This supports a useful role for centered
correlations under the tested recipe, beyond a mean-only explanation; it
does not exclude other coordinatewise methods. Hidden-weight Adam also fails
to reproduce the synthetic activity gain in the exploratory settings \citep{kingma2015adam}
(Table~\ref{tab:infodfa_adam_control}).

\paragraph{Input reconditioning in exact BP.}
The same right preconditioner also improves BP in the nuisance-dominant and
low-sample/noisy regimes, with little change in the mixed and clean controls
(Fig.~\ref{fig:controls_composite}C), after separate tuning over the same
learning-rate grid on the reported test data. This exploratory control suggests
a benefit shared with ordinary preconditioning, with unresolved tuning effects. nDFA remains
higher in the nuisance regime, but the tuning and regularization contributions
to that difference are unresolved (Appendix Table~\ref{tab:infodfa_bpwhiten}).
A separate validation-selected CIFAR-10 protocol also favors conditioned BP,
while changing the ordering of DFA variants (Appendix~\ref{app:boundary}).

\subsection{Activity gains survive stronger baselines and matched work}
\label{sec:matched_work_main}

\begin{table}[t]
\centering\ifdefined\ARXIV\normalsize\else\small\fi
\ifdefined\ARXIV\setlength{\tabcolsep}{6pt}\else\setlength{\tabcolsep}{4pt}\fi
\begin{tabular}{@{}lrr@{}}
\toprule
Rule & Test accuracy (\%) & Test CE\\
\midrule
DFA (AdamW) & $65.66\pm0.25$ & $0.991\pm0.003$\\
BP (AdamW) & $66.88\pm0.25$ & $0.963\pm0.006$\\
FD-DFA & $67.01\pm0.25$ & $0.960\pm0.006$\\
A-nDFA & $67.92\pm0.27$ & $0.943\pm0.004$\\
Early A-nDFA & $68.41\pm0.34$ & $0.945\pm0.006$\\
FOOF-DFA & $69.21\pm0.30$ & $0.954\pm0.007$\\
FOOF-BP & $70.03\pm1.17$ & $0.954\pm0.008$\\
\bottomrule
\end{tabular}
\caption{\textbf{Joint CIFAR-10 confirmation against tuned baselines.} Mean $\pm$ sample SD over ten paired training seeds. Each run receives 200 seconds of synchronized learning work on an H100; the checkpoint is selected by validation CE. Calibration, conditioning and inverse refresh are charged; validation and export are excluded. The 3072--1024--512--10 MLP and data are shared. Core arms use BN; FD-DFA uses forward decorrelation instead. Settings are selected separately. Appendix~\ref{app:joint_confirmation} gives all arms, estimated FLOPs, update counts and corrected paired tests.}
\label{tab:practical_main}
\end{table}


Practical value requires gains that remain after tuning the baseline and
charging for conditioning. In the ten-seed CIFAR-10 confirmation, activity
conditioning, its early-only variant and FOOF-DFA all improve tuned AdamW DFA
in accuracy, without increasing mean CE (Table~\ref{tab:practical_main};
Fig.~\ref{fig:bn_optimizer_main}A). Every paired seed favors each method, and
all three pass the prespecified practical and corrected statistical criteria.
FOOF-DFA has the highest mean accuracy among the tested DFA methods;
activity conditioning has lower mean CE. FOOF-BP remains the stronger
accuracy reference. These tests establish gains over tuned DFA, not
superiority to every BP or decorrelation method. The credit-rule-by-FOOF
interaction does not establish a preferential benefit for DFA. FOOF also
uses substantially more estimated arithmetic work, so matching learning time
does not establish a FLOP advantage.

\paragraph{When is conditioning needed?}
At constant learning rate, conditioning during the first quarter of training
outperforms conditioning during the last quarter, but the corresponding
interaction does not establish a difference between BP and DFA
(Fig.~\ref{fig:controls_composite}D). Shared-checkpoint development
interventions also show that switching conditioning on late can improve both
credit rules. Thus the evidence supports an early-window advantage, not
irreversible nuisance capture or a DFA-specific timing mechanism.
In a separate fixed-epoch cosine-schedule experiment, early-only activity
conditioning meets the prespecified accuracy noninferiority criterion and
uses about one-third less measured learning work than persistent conditioning
(Fig.~\ref{fig:bn_optimizer_main}C). This tradeoff is conditional on that
recipe and differs from the time-matched comparison in Table~\ref{tab:practical_main}.

\paragraph{Where does the nuisance prediction transfer?}
The synthetic model motivates a directional prediction, not a guarantee
for any noisy input. We test clean images, training-label noise, and images
mixed with independent backgrounds on MNIST and Fashion-MNIST, plus the
original MNIST-background-images benchmark. The predicted background gains
are not confirmed (Fig.~\ref{fig:bn_optimizer_main}D). In both constructed
background tasks, activity conditioning worsens CE after correction, despite
mixed accuracy effects. Input nuisance alone therefore does not identify a
beneficial conditioning regime.

\paragraph{Normalization and scope.}
Earlier width and work studies support activity benefits but show that
retuned error damping can nearly suppress E
(Appendices~\ref{app:scale}--\ref{app:revision_followups}). A separate
unnormalized protocol produces extreme predictive losses for several
conditioned rules (Appendix~\ref{app:boundary}). Under a fixed-recipe intervention, BN restores ordinary
loss scales for both raw and conditioned DFA; reducing feedback scale alone
does not (Appendix~\ref{app:bn_scale_intervention}; Fig.~\ref{fig:revision_followups}). These controls explain why both
accuracy and loss, and the normalization recipe, are necessary to interpret
a conditioning benefit.

\section{Related Work}
\label{sec:related}

Conditioned DFA connects fixed feedback with update preconditioning.
FA replaces exact weight symmetry with fixed random feedback, and DFA broadcasts
the output error directly to hidden layers
\citep{lillicrap2016feedback,nokland2016dfa,refinetti2021align}. Later work
scaled DFA to Transformers but exposed harder vision limits
\citep{bartunov2018scalability,launay2020dfa,wang2026optical}. We condition its
local update; GrAPE instead aligns feedback with forward-mode Jacobian estimates
and BP anchors \citep{caillon2026grape}, a complementary nonlocal axis.

The activity factor uses FOOF's core operator with DFA errors
\citep{benzing2022foof}. The two-sided form resembles KFAC, but its local-error
moment need not estimate the Fisher matrix \citep{martens2015kfac}.
Related normalization and decorrelation methods act on the forward representation
\citep{desjardins2015naturalnn,huang2018decorrelated,huang2019iternorm}.
Task-dependent preconditioning benefits and mode-dependent early stopping
predate this study \citep{amari2021preconditioning,heckel2021earlystopping}.
Covariance-power and spectral-gradient analyses show task-dependent effects
\citep{yoshida2025preconditioning,braun2026spectral}; we apply the factors under
fixed DFA feedback. Shampoo also applies structured matrix preconditioning,
but builds accumulated gradient-derived factors rather than the current-batch
activity and local-error moments used here \citep{gupta2018shampoo}.
Our contribution is the controlled factor study under imperfect fixed feedback.
Update orthogonalization and activity normalization can likewise change effective
FA update rank \citep{boeshertz2026rankcollapse}; their mechanisms overlap with
conditioning even though invertible factors preserve algebraic rank.

Further comparisons with alternative feedback pathways, credit signals,
and conditioning methods appear in Appendix~\ref{app:related_expanded}.

\section{Discussion and Limitations}
\label{sec:discussion}

Activity conditioning improves learning with random feedback beyond a scalar
step-size change. With stronger baselines and a frozen joint confirmation,
its practical benefit survives, centered correlations contribute beyond the
tested mean-preserving control, and early-only conditioning offers a useful
accuracy--work tradeoff. These are properties of established conditioning
operators applied to approximate credit, not a new matrix inverse or a
general advantage over backpropagation.

The two factors have unequal empirical support. Error conditioning changes
directions and improves historical short-budget models, yet its small
additional stable-training gain fails the declared corrected criterion.
This outcome is consistent with the importance of the activity factor in
FOOF/KFAC comparisons \citep{benzing2022foof}, without implying that DFA's
empirical error moment is a Fisher factor or that error conditioning is
universally ineffective. Likewise, the failed image-background predictions
show that the synthetic geometry cannot be inferred from a dataset's noise
label. Our aligned theory does not explain alignment acquisition, and the
timing interventions do not establish a DFA-specific mechanism.

The main practical evidence concerns MLPs and fixed data splits. Training-seed
uncertainty does not include dataset or hyperparameter-selection uncertainty;
standard test sets had historical exposure, and one benchmark cohort changed
accelerator before confirmation. Equal measured learning time excludes
validation and export and does not imply equal FLOPs. Exact sample-space
solves reduce the cost of wide layers, but moments and solves remain
layer-wide operations, not synaptically local plasticity. Within these
limits, activity geometry provides a reproducible way to improve the use of
fixed random feedback.\label{sec:main_text_end}

\section*{Reproducibility Statement}

\ifdefined\ARXIV
Code, frozen experiment configurations, and seed-level confirmation results are available at
\href{https://github.com/KempnerInstitute/conditioned-dfa}%
{\nolinkurl{github.com/KempnerInstitute/conditioned-dfa}}.
\else
The appendix is self-contained and gives the protocols, selected settings,
all primary confirmation contrasts and their seed-level values, assumptions,
and deviations needed to assess the main claims. Code and machine-readable
experimental records are publicly available. No external code attachment
is required to read the supporting analyses.
\fi

\ifdefined\ARXIV\else
\section*{AI Use Statement}

Generative AI assisted research planning and derivations, coding and
synthetic-task implementation, experiment execution and analysis, figure
preparation, literature search, and manuscript drafting and editing. The
authors take responsibility for the numerical results, references, manuscript,
and accompanying artifacts.
\fi

\ifdefined\ARXIV
\section*{Acknowledgements}

This work has been made possible in part by a gift from the Chan Zuckerberg
Initiative Foundation to establish the Kempner Institute for the Study of
Natural and Artificial Intelligence at Harvard University. The computations in
this paper were run on the Kempner Institute AI cluster at Harvard University.
\fi

\input{supplement}

%% file: supplement.tex
\appendix
\setcounter{topnumber}{3}
\setcounter{totalnumber}{4}
\renewcommand{\topfraction}{0.95}
\renewcommand{\textfraction}{0.05}
\renewcommand{\floatpagefraction}{0.7}
\section*{Supplementary material}
The supplement begins with the joint confirmation
(Appendix~\ref{app:joint_confirmation}), including all primary comparisons,
selected settings, seed-level values and deviations. The remaining appendices
provide the aligned theory and numerical protocols
(Appendix~\ref{app:theory}), implementation and historical study provenance
(Appendix~\ref{app:implementation}), synthetic controls
(Appendix~\ref{app:synthetic}), and short-budget factor studies
(Appendix~\ref{app:threefactor}). Appendices~\ref{app:scale}--\ref{app:boundary}
document earlier CIFAR-10 scale, work, damping and stability comparisons.
The orientation construction and related mechanisms conclude the supplement
(Appendices~\ref{app:factor_mechanism}--\ref{app:related_expanded}).
Unless a caption specifies sample SD, $\pm$ denotes SEM over the stated
replication unit. Each study retains its own analysis family. Earlier
studies are not pooled with the joint confirmation. Exploratory architectures
that do not support the present claims are omitted.

\FloatBarrier
\section{Joint confirmation: protocol and complete evidence}
\label{app:joint_confirmation}

The joint confirmation asks which conditioning benefits survive stronger
baselines and prospectively fixed comparisons. Development, model selection
and scientific interpretation are distinguished from the subsequent test
evaluation. This appendix reports the complete primary family, including
failed predictions; the historical studies below retain their original
protocols and do not enlarge this family retrospectively.

\subsection{Design, selection and replication}

The final training inventory contains 880 models in eight cohorts
(Table~\ref{tab:joint_ledger}). Each selected setting uses the same ten model
seeds, 940101--940110; feedback seeds add 1000. Pairing is within a task and
cohort, not across independently selected studies. The real-data splits are
fixed, so uncertainty describes training randomness conditional on those
splits and choices. Synthetic seeds additionally regenerate the task data.
All retained models complete with finite endpoints. The preceding expanded
development includes 882 runs: 878 complete and four fail numerically, all
at the largest tested FOOF-BP learning rate on CIFAR-10. Failed candidates
are ineligible for selection; no selected family is substituted after testing.

Expanded development uses two paired seeds per candidate. Selection minimizes
mean best validation CE, then breaks ties by validation accuracy and fixed
case order. Work-budget evaluations occur at approximately 5\% budget
increments, including initialization and the final step. The early-activity
CIFAR candidate comes from the preceding matched-work development, selected
on validation CE at the same budget. Geometry settings and fixed-epoch
comparisons retain their previously frozen validation choices. The
constant-rate timing comparison fixes the declared learning rate rather than
choosing a favorable rate after inspection. These studies inform one joint
confirmation, not independent sequences of test-guided tuning.

For the practical CIFAR factorial, raw SGD searches learning rates
$\{0.0003,0.003,0.03,0.3,3,30\}$; FOOF crosses rates $\{0.3,3,30\}$ with
absolute damping $\{0.03,0.3,3\}$ for both BP and DFA. AdamW references search
$\{10^{-4},3\times10^{-4},10^{-3}\}$; FD-DFA searches
$\{3\times10^{-5},10^{-4},3\times10^{-4}\}$. A crosses
$\{3\times10^{-4},10^{-3},3\times10^{-3}\}$ with relative damping
$\{1/3,1,3\}$. Thus BP/DFA factorial grids are balanced, but optimizer families
do not have identical numbers of candidates. Selected A and FD settings lie
on boundaries; these finite searches do not establish optimal convergence
envelopes. The error studies search the common A-selected rate over
$\{10^{-4},3\times10^{-4},10^{-3}\}$ and relative E damping over
$\{1,10,100\}$, without enlarging an edge selection. Image-benchmark raw
AdamW and FD grids use those respective rate sets; A crosses the AdamW grid
with $\rho_A\in\{0.3,3\}$, and FOOF crosses rates $\{0.03,0.3,3\}$ with
absolute damping $\{0.3,3\}$.

The common seed list, cases, source and analysis are frozen before confirmation
evaluation. All included training finishes and passes the global audit before
any joint test evaluation. A separate interpretation map is recorded after
development but before test access; it does not change the tests or thresholds
and is not represented as a preregistration preceding development. The
standard CIFAR-10, MNIST and Fashion-MNIST test sets had historical project
exposure. Fresh training seeds do not make those datasets untouched. The
original MNIST-background-images test member is first read after the joint
gate. No test score selects a setting, horizon, model seed or replacement run.

\begin{table}[!htbp]
\centering\small
\setlength{\tabcolsep}{4pt}
\begin{tabular}{@{}llrll@{}}
\toprule
Cohort & Budget & Models & Test endpoint & GPU\\
\midrule
CIFAR practical / FOOF & 200 s & 100 & Best validation CE & H100\\
Synthetic FOOF & 20 s & 140 & Best validation CE & H200\\
CIFAR temporary E & 200 s & 40 & Best validation CE & H100\\
MNIST temporary E & 60 s & 40 & Best validation CE & H200\\
Image benchmarks & 60 s & 420 & Best validation CE & RTX PRO 6000\\
CIFAR moment controls & 200 epochs & 70 & Final & H100\\
CIFAR constant-rate timing & 200 epochs & 40 & Final & H200\\
CIFAR early-A efficiency & 200 epochs & 30 & Final & H100\\
\bottomrule
\end{tabular}
\caption{\textbf{Joint confirmation ledger.} Time budgets measure synchronized
learning work, not total elapsed experiment time. Every case has ten paired
seeds. The final endpoint for the efficiency criterion is specified in the
analysis, separately from the validation-selected work-budget endpoint.}
\label{tab:joint_ledger}
\end{table}

\subsection{Architecture, operators and work accounting}

CIFAR-10 uses a 3072--1024--512--10 ReLU MLP, 45,000 training and 5,000
validation images (split seed 925005), batch size 256, and feedback scale 0.1.
Input channel moments are estimated on training images only. Training uses
reflected-padding-four random crops and horizontal flips; validation and
test do not. Core comparisons share BN before ReLU. FD-DFA uses learned
forward decorrelation instead of BN, with decorrelation rate $10^{-5}$.
AdamW uses $(\beta_1,\beta_2)=(0.9,0.999)$ and $\epsilon=10^{-8}$; weight
decay is $10^{-4}$ on weight matrices and zero on biases and BN parameters.
The raw SGD/FOOF factorial uses SGD without momentum. Thus the tuned AdamW
DFA baseline prevents interpreting an advantage over weak SGD alone as the
practical result. The factorial interaction uses each credit rule's SGD
reference, not its AdamW reference.

For A, $\lambda_A=\max(\rho_A\,\tr(C_A)/d,10^{-6})$, where $d$ is the
presynaptic width. Current-minibatch moments condition hidden weight matrices,
followed by norm matching to each matrix's raw pseudo-gradient. The resulting
matrices enter AdamW before its moment updates; their matched norms do not
imply matched parameter-step norms. Biases, readout and BN-affine
parameters retain their learner's updates. Centered and mean-preserving
controls change only the denominator moment; the raw numerator and
uncentered-energy damping convention remain fixed. FOOF instead uses
normalized exponential moving moments with decay 0.95, 50 calibration
minibatches, and inverse refresh every 100 updates. It conditions every
weight matrix, including the readout, without norm matching; bias and
BN-affine updates remain SGD. Its absolute damping applies to mean moments.
This adapts the published FOOF recipe to the present biased/BN architecture;
it is not a reproduction of the original paper's experimental setting.

Work-budget cosine schedules warm up for the first 2.5\% of measured work
and decay to 1\% of the peak rate. Fixed-epoch cosine schedules warm up for
five epochs. The timing and temporary-E studies use constant learning rates
without warmup or decay. Early and late conditioning occupy the first and
last quarters of the declared work or update budget, respectively.

Learning time includes synchronization, sampling and augmentation, FOOF
calibration, moment/inverse updates, gradient and optimizer work, and numerical
checks. Validation, fixed-state probes, hashing and exports are excluded.
Estimated leading-order linear-algebra FLOPs count matrix products, Gram
systems, solves, FOOF calibration and inverse refresh, and FD transforms,
with fused multiply--add counted as two operations. They exclude pointwise
activation, BN reductions, optimizer arithmetic, data transforms and exports.
These estimates are not hardware-counter FLOPs, and time matching does not
imply arithmetic matching. All training uses float32 with TF32 disabled;
evaluation losses use float64 reductions (PyTorch 2.9.1, CUDA 12.8).

\subsection{Hardware deviation and interruption handling}

Benchmark settings were selected on H200 GPUs. Before any benchmark
confirmation run began, queue constraints led us to move the complete
420-model cohort, including all baselines, to RTX PRO 6000 Blackwell GPUs.
Selected settings, seeds and the nominal 60-second budget were retained,
and the amendment was recorded prospectively. Across 40 eligible case
groups with uninterrupted records, confirmation performs about 35\% more
updates on average than development. Different seeds and case-dependent
rates preclude interpreting that ratio as a controlled hardware speedup.
Early checkpoint selection does not remove this change in training horizon.
All comparisons within the benchmark confirmation use the same GPU model;
the main H100 CIFAR comparison is unaffected.

Five benchmark runs were interrupted and resumed by the scheduler, which
violated the uninterrupted-work requirement. Before test access, all five
were selected for one cold replay solely from interruption metadata: A on
the MNIST mixture at seed 940110, and FOOF-DFA on that mixture at seeds
940106--940109. Original records and checkpoints were archived. Each replay
used identical settings, seed, source, data, initialization, feedback and
GPU type, and completed in one attempt. The failed audit and replay rule
remain in the provenance record; these are replacements for interrupted
attempts, not additional independent observations. Fixed-update geometry
and timing experiments permit state-preserving resumption and are not used
to claim uninterrupted timing efficiency.

\subsection{Statistical criteria fixed before testing}
\label{app:joint_criteria}

The primary family consists of 21 two-sided paired $t$ tests, with Holm
correction across the entire family. Table~\ref{tab:joint_primary} gives
individual 95\% $t$ intervals and Holm-adjusted $p$-values; the intervals do
not provide simultaneous familywise coverage. A practical CIFAR claim also
requires a mean gain of at least one accuracy percentage point and no
increase in mean CE. The three practical comparisons pass these criteria.
Nonsignificance is not evidence of equivalence.

Early-only efficiency is a separate prespecified noninferiority question at
the final fixed-epoch endpoint. It requires a one-sided 95\% lower bound for
early-minus-full accuracy above $-0.5$ percentage points and a mean paired
work ratio at most 0.8. The observed difference is $-0.164$ points, the lower
bound is $-0.401$, and the work ratio is 0.658; this criterion passes. The
individual two-sided interval is $[-0.456,0.128]$ points. This is a conditional
accuracy--work tradeoff, not equal performance or a schedule-independent law.

Clean and label-noise benchmark predictions use secondary individual 90\%
paired intervals and a $\pm1$-point equivalence margin. MNIST clean and both
Fashion conditions meet the margin; MNIST label noise does not. These are
not a familywise equivalence claim. All 21 primary seed contrasts are given
at the end of this appendix, including effects opposite to prediction.

\begin{table}[!htbp]
\centering\small
\setlength{\tabcolsep}{4pt}
\begin{tabular}{@{}rlrrr@{}}
\toprule
ID & Primary contrast & Mean & Individual 95\% CI & $p_{\rm Holm}$\\
\midrule
1 & A -- tuned DFA & $+2.263$ & $[1.958, 2.568]$ & $6.85\!\times\!10^{-7}$\\
2 & Early A -- tuned DFA & $+2.756$ & $[2.448, 3.064]$ & $1.37\!\times\!10^{-7}$\\
3 & FOOF-DFA -- tuned DFA & $+3.554$ & $[3.240, 3.868]$ & $1.95\!\times\!10^{-8}$\\
4 & FOOF gain: DFA -- BP & $-0.229$ & $[-1.148, 0.690]$ & $1.000$\\
5 & Full -- mean-preserving control & $+3.205$ & $[3.006, 3.404]$ & $8.88\!\times\!10^{-10}$\\
6 & Centered -- mean-preserving control & $+3.126$ & $[2.817, 3.435]$ & $5.03\!\times\!10^{-8}$\\
7 & Early -- late A (DFA) & $+1.186$ & $[0.926, 1.446]$ & $4.07\!\times\!10^{-5}$\\
8 & Early--late effect: DFA -- BP & $+0.055$ & $[-0.455, 0.565]$ & $1.000$\\
9 & Early K -- A: accuracy & $+0.384$ & $[0.018, 0.750]$ & $0.315$\\
10 & Early K -- A: CE improvement & $+0.0071$ & $[0.0019, 0.0122]$ & $0.155$\\
11 & Early K -- diagonal E: accuracy & $+0.413$ & $[0.025, 0.801]$ & $0.315$\\
12 & Early K -- diagonal E: CE improvement & $+0.0086$ & $[0.0020, 0.0153]$ & $0.159$\\
13 & Error direction change above 0.001 & $+0.3773$ & $[0.3749, 0.3796]$ & $1.04\!\times\!10^{-18}$\\
14 & MNIST mixture: accuracy & $-0.755$ & $[-1.373, -0.137]$ & $0.198$\\
15 & MNIST mixture: CE improvement & $-0.0303$ & $[-0.0473, -0.0134]$ & $0.038$\\
16 & MNIST: nuisance -- clean gain & $-0.320$ & $[-0.941, 0.301]$ & $1.000$\\
17 & Fashion mixture: accuracy & $+0.548$ & $[0.145, 0.951]$ & $0.155$\\
18 & Fashion mixture: CE improvement & $-0.0153$ & $[-0.0230, -0.0076]$ & $0.020$\\
19 & Fashion: nuisance -- clean gain & $+0.467$ & $[0.028, 0.906]$ & $0.315$\\
20 & Original backgrounds: accuracy & $+0.143$ & $[-0.268, 0.554]$ & $1.000$\\
21 & Original backgrounds: CE improvement & $+0.0033$ & $[-0.0049, 0.0115]$ & $1.000$\\
\bottomrule
\end{tabular}
\caption{\textbf{All 21 primary confirmation contrasts.} Each uses ten paired training seeds. Accuracy effects are percentage points; positive CE effects mean lower loss. The direction row is $1-\cos(K,A)-0.001$. Holm correction covers all rows together. Individual intervals are not simultaneous intervals. Negative significant loss effects contradict the positive nuisance prediction.}
\label{tab:joint_primary}
\end{table}

\FloatBarrier

\subsection{Interpretation of the selected comparisons}

The primary practical comparisons favor A, early A and FOOF-DFA over tuned
DFA, with favorable accuracy differences in all ten seed pairs and lower
mean CE. Table~\ref{tab:joint_practical_all} includes the SGD factorial,
unnormalized baseline and arithmetic costs omitted from the compact main
table. The FOOF-by-credit interaction is the difference between the FOOF
gain over SGD in DFA and that in BP. Its interval spans zero. The data
therefore establish that FOOF works with random feedback, without establishing
preferential benefit for DFA. No equivalence test was specified for the
FOOF-DFA/FOOF-BP gap.

\begin{table}[!htbp]
\centering\small
\ifdefined\ARXIV\setlength{\tabcolsep}{6pt}\else\setlength{\tabcolsep}{4pt}\fi
\begin{tabular}{@{}llrr@{}}
\toprule
Rule & Optimizer & $\eta$ & $\rho_A$ or $\lambda$\\
\midrule
BP (SGD) & SGD & $0.03$ & --\\
FOOF-BP & SGD & $3$ & $0.3$\\
BP (AdamW) & AdamW & $0.0001$ & --\\
DFA (SGD) & SGD & $0.03$ & --\\
FOOF-DFA & SGD & $3$ & $0.3$\\
DFA (AdamW) & AdamW & $0.0001$ & --\\
FD-DFA & AdamW & $0.0003$ & --\\
A-nDFA & AdamW & $0.0003$ & $3$\\
DFA without BN & AdamW & $0.0001$ & --\\
Early A-nDFA & AdamW & $0.0005$ & $1$\\
\bottomrule
\end{tabular}
\caption{\textbf{Selected settings for the practical CIFAR-10 cohort.} A uses relative activity damping; FOOF uses absolute damping of mean moments. FD-DFA uses decorrelation rate $10^{-5}$. Early A uses conditioning for the first quarter of learning work. These choices precede test evaluation.}
\label{tab:joint_practical_settings}
\end{table}

\begin{table}[!htbp]
\centering\small
\setlength{\tabcolsep}{4pt}
\begin{tabular}{@{}lrrrr@{}}
\toprule
Rule & Accuracy (\%) & CE & Updates & FLOPs ($10^{12}$)\\
\midrule
BP (SGD) & $65.70\pm0.32$ & $0.9999$ & 70,269 & 283.1\\
FOOF-BP & $70.03\pm1.17$ & $0.9537$ & 48,830 & 1476.7\\
BP (AdamW) & $66.88\pm0.25$ & $0.9631$ & 61,787 & 249.0\\
DFA (SGD) & $65.11\pm0.35$ & $1.0108$ & 70,839 & 266.8\\
FOOF-DFA & $69.21\pm0.30$ & $0.9540$ & 49,006 & 1469.2\\
DFA (AdamW) & $65.66\pm0.25$ & $0.9913$ & 62,035 & 233.7\\
FD-DFA & $67.01\pm0.25$ & $0.9602$ & 42,177 & 437.8\\
A-nDFA & $67.92\pm0.27$ & $0.9435$ & 35,333 & 226.2\\
DFA without BN & $61.30\pm0.60$ & $1.1444$ & 89,829 & 338.3\\
Early A-nDFA & $68.41\pm0.34$ & $0.9449$ & 55,343 & 231.8\\
\bottomrule
\end{tabular}
\caption{\textbf{Complete practical CIFAR-10 cohort.} Accuracy is mean $\pm$ sample SD over ten paired seeds; CE is the mean. Each run receives 200 seconds of measured learning work. Test metrics use the validation-CE-selected checkpoint. Update counts and estimated FLOPs cover the complete budget, including work after that checkpoint; selected checkpoints need not occur at equal times or FLOP counts.}
\label{tab:joint_practical_all}
\end{table}

\FloatBarrier

\subsection{Activity statistics}

The geometry experiment compares $C_A$, $\Cov(h)$,
$\diag(C_A)$, $\diag(\Cov(h))+\mu\mu^\top$, and
$\tr(\Cov(h))I/d+\mu\mu^\top$, as well as raw DFA and a full-moment
sample-space implementation. Full and centered moments both improve over
the mean-preserving diagonal-covariance control in every seed. The controls
share the raw numerator, norm matching and damping convention, but use
separately validation-selected damping. Their comparison supports a role
for centered correlations beyond the tested mean-only explanation; it does
not show that every diagonal optimizer or mean-based method must fail.

\begin{table}[!htbp]
\centering\small
\ifdefined\ARXIV\setlength{\tabcolsep}{6pt}\else\setlength{\tabcolsep}{4pt}\fi
\begin{tabular}{@{}lrrrr@{}}
\toprule
Rule & Accuracy (\%) & CE & $\eta$ & $\rho_A$\\
\midrule
Centered covariance & $67.90\pm0.30$ & $0.9577$ & $0.001$ & $1$\\
Diagonal moment & $64.64\pm0.22$ & $1.0390$ & $0.001$ & $0.1$\\
Diagonal covariance + mean & $64.77\pm0.34$ & $1.0410$ & $0.001$ & $30$\\
Full moment & $67.98\pm0.24$ & $0.9569$ & $0.001$ & $1$\\
Isotropic covariance + mean & $64.70\pm0.32$ & $1.0429$ & $0.001$ & $30$\\
Raw DFA & $64.78\pm0.30$ & $1.0343$ & $0.001$ & --\\
Full (sample-space solve) & $67.98\pm0.27$ & $0.9573$ & $0.001$ & $1$\\
\bottomrule
\end{tabular}
\caption{\textbf{Final fixed-epoch geometry endpoints.} Accuracy is mean $\pm$ sample SD over ten paired seeds; CE is the mean. Final checkpoints after 200 epochs. Relative activity damping $\rho_A$ uses the uncentered activity energy for every geometry control. A dash denotes an inactive factor.}
\label{tab:joint_geometry}
\end{table}

\FloatBarrier

\subsection{Timing and efficiency}

The constant-rate timing interaction compares early-minus-late gains in DFA
and BP. The early-window DFA effect survives correction; its extra benefit
over BP does not. Shared-state interventions use four development seeds,
forks at epochs 50 and 150, and equal 50-epoch continuations with the same
future samples and optimizer state. Conditioning can still help after a late
fork. These results do not establish irreversible nuisance capture.

\begin{table}[!htbp]
\centering\small
\ifdefined\ARXIV\setlength{\tabcolsep}{6pt}\else\setlength{\tabcolsep}{4pt}\fi
\begin{tabular}{@{}lrrrr@{}}
\toprule
Rule & Accuracy (\%) & CE & $\eta$ & $\rho_A$\\
\midrule
DFA, early A & $65.76\pm0.27$ & $1.0023$ & $0.0003$ & $1$\\
DFA, late A & $64.58\pm0.30$ & $1.0543$ & $0.0003$ & $1$\\
BP, early A & $67.55\pm0.34$ & $0.9951$ & $0.0003$ & $1$\\
BP, late A & $66.41\pm0.45$ & $1.0756$ & $0.0003$ & $1$\\
\bottomrule
\end{tabular}
\caption{\textbf{Final constant-rate timing endpoints.} Accuracy is mean $\pm$ sample SD over ten paired seeds; CE is the mean. Final checkpoints after 200 epochs at constant learning rate. Relative activity damping $\rho_A=1$ applies during the first quarter (early) or last quarter (late); the credit rule is shown separately.}
\label{tab:joint_timing}
\end{table}

\begin{table}[!htbp]
\centering\small
\ifdefined\ARXIV\setlength{\tabcolsep}{6pt}\else\setlength{\tabcolsep}{4pt}\fi
\begin{tabular}{@{}lrrrr@{}}
\toprule
Rule & Accuracy (\%) & CE & $\eta$ & $\rho_A$\\
\midrule
DFA & $65.01\pm0.25$ & $1.0109$ & $0.0003$ & --\\
A-nDFA & $67.44\pm0.24$ & $0.9647$ & $0.003$ & $0.3333$\\
Early A-nDFA & $67.27\pm0.32$ & $0.9743$ & $0.003$ & $0.3333$\\
\bottomrule
\end{tabular}
\caption{\textbf{Final fixed-epoch efficiency endpoints.} Accuracy is mean $\pm$ sample SD over ten paired seeds; CE is the mean. Final checkpoints after 200 epochs with a cosine schedule. Relative activity damping $\rho_A$ applies throughout or during the first quarter; a dash denotes no conditioning.}
\label{tab:joint_epochs}
\end{table}

\begin{table}[!htbp]
\centering\small
\setlength{\tabcolsep}{4pt}
\begin{tabular}{@{}lllrr@{}}
\toprule
Task & Fork & Prior A & DFA gain (pp) & BP gain (pp)\\
\midrule
CIFAR-10 & 25\% & On & $2.51\pm0.92$ & $1.49\pm0.61$\\
CIFAR-10 & 25\% & Off & $1.68\pm0.85$ & $1.04\pm0.47$\\
CIFAR-10 & 75\% & On & $2.04\pm0.29$ & $1.10\pm0.29$\\
CIFAR-10 & 75\% & Off & $1.16\pm0.40$ & $0.71\pm0.68$\\
Synthetic & 25\% & On & $1.11\pm0.65$ & $1.82\pm0.29$\\
Synthetic & 25\% & Off & $1.27\pm0.31$ & $2.29\pm0.19$\\
Synthetic & 75\% & On & $0.60\pm0.18$ & $0.61\pm0.08$\\
Synthetic & 75\% & Off & $1.04\pm0.28$ & $0.71\pm0.28$\\
\bottomrule
\end{tabular}
\caption{\textbf{Shared-checkpoint development interventions.} Mean $\pm$ sample SD of validation-accuracy differences over four paired prefix seeds. At each saved state, conditioning-on and conditioning-off branches receive the same 50-epoch continuation, optimizer state, batches and constant learning rate. These are development results, outside the primary family; late gains contradict an irreversible-capture interpretation.}
\label{tab:joint_branches}
\end{table}

\FloatBarrier

\subsection{Temporary error conditioning}

\label{app:joint_error}
The error test uses the common learning rate selected on A alone and holds
it constant. CIFAR uses $\eta=3\times10^{-4}$ and $\rho_A=1$; early full E
selects relative damping $\rho_E=1$, while diagonal E selects $\rho_E=100$.
MNIST uses a 784--256--128--10 ReLU MLP without BN, batch 128, 55,000/5,000
training/validation examples, feedback scale 1, AdamW without weight decay,
$\eta=10^{-3}$, absolute $\lambda_A=0.3$, and $\rho_E=10$ for both E forms.
Its images are scaled by $1/255$ without augmentation. Relative E damping
scales the mean squared per-example gated local error; minibatch mean-loss
scaling is undone before the error moment is formed.

Early K applies E during the first quarter of measured learning work and
then continues A. Persistent K uses the selected early-K damping throughout.
A positive scalar E factor cancels under hidden-matrix norm matching, so A
is also the matched scalar comparator; this identity was numerically
checked, not counted as another independent training arm. The directional
measure is $1-\cos(K,A)$ between conditioned pseudo-gradients before
AdamW, on common states and fixed training batches,
averaged over hidden layers and positive-time probes before the early-window
boundary, then within each seed. It isolates E rather than conflating E
with activity conditioning.

The prespecified E claim requires improved accuracy and CE over both A and
diagonal E, and directional change greater than 0.001; all five tests enter
the global family. Early K has favorable means over A (0.384 accuracy
points and 0.0071 lower CE), but the Holm-adjusted $p$-values are 0.315 and
0.155. Only the directional criterion passes. The gate therefore fails:
the result is a small positive estimate with insufficient corrected evidence,
not a negative effect or an exactly zero effect. The secondary MNIST result
cannot replace this failed CIFAR criterion. No window, seed or learning-rate
extension follows it.

\begin{table}[!htbp]
\centering\small
\ifdefined\ARXIV\setlength{\tabcolsep}{6pt}\else\setlength{\tabcolsep}{4pt}\fi
\begin{tabular}{@{}lrrrrr@{}}
\toprule
Rule & Accuracy (\%) & CE & $\eta$ & $\rho_A$ & $\rho_E$\\
\midrule
A-nDFA & $66.87\pm0.23$ & $0.9777$ & $0.0003$ & $1$ & --\\
Early K-nDFA $\to$ A-nDFA & $67.25\pm0.39$ & $0.9706$ & $0.0003$ & $1$ & $1$\\
Early A + diagonal E $\to$ A & $66.84\pm0.23$ & $0.9793$ & $0.0003$ & $1$ & $100$\\
K-nDFA throughout & $66.47\pm0.34$ & $0.9703$ & $0.0003$ & $1$ & $1$\\
\bottomrule
\end{tabular}
\caption{\textbf{Stable-training CIFAR-10 error-factor endpoints.} Accuracy is mean $\pm$ sample SD over ten paired seeds; CE is the mean. Validation-CE-selected checkpoints at 200 seconds of measured learning work. Activity damping $\rho_A$ and error damping $\rho_E$ are relative to their respective uncentered energies. Early error factors apply during the first quarter; activity conditioning remains active throughout. A is also the matched scalar-error control.}
\label{tab:joint_error_cifar}
\end{table}

\begin{table}[!htbp]
\centering\small
\ifdefined\ARXIV\setlength{\tabcolsep}{6pt}\else\setlength{\tabcolsep}{4pt}\fi
\begin{tabular}{@{}lrrrrr@{}}
\toprule
Rule & Accuracy (\%) & CE & $\eta$ & $\lambda_A$ & $\rho_E$\\
\midrule
A-nDFA & $97.54\pm0.11$ & $0.0834$ & $0.001$ & $0.3$ & --\\
Early K-nDFA $\to$ A-nDFA & $97.77\pm0.15$ & $0.0764$ & $0.001$ & $0.3$ & $10$\\
Early A + diagonal E $\to$ A & $97.53\pm0.13$ & $0.0842$ & $0.001$ & $0.3$ & $10$\\
K-nDFA throughout & $97.73\pm0.12$ & $0.0764$ & $0.001$ & $0.3$ & $10$\\
\bottomrule
\end{tabular}
\caption{\textbf{Secondary MNIST error-factor endpoints.} Accuracy is mean $\pm$ sample SD over ten paired seeds; CE is the mean. Validation-CE-selected checkpoints at 60 seconds of measured learning work. Activity damping $\lambda_A$ is absolute; error damping $\rho_E$ is relative to per-example local-error energy. Early error factors apply during the first quarter, with activity conditioning throughout.}
\label{tab:joint_error_mnist}
\end{table}

\FloatBarrier

\subsection{Image-benchmark predictions}

Constructed MNIST and Fashion conditions share 10,000 training and 2,000
validation foreground identities (training-only split seed 934005). The
clean condition scales pixels by $1/255$. Label noise changes 20\% of training
labels independently to a different class; validation and test labels remain
clean. Input nuisance mixes 0.35 foreground with 0.65 independently assigned
grayscale CIFAR-10 training-image background, center-cropped to $28\times28$.
The background pools for training, validation and test are disjoint and
assignment does not use foreground class. These are paired constructed
variants, not the original MNIST-variations dataset. The original
MNIST-background-images archive \citep{larochelle2007deep} supplies a separate standard task: its first
10,000 training rows train, its remaining 2,000 rows validate, and its
50,000-row test member is evaluated after the global gate.

All image-benchmark conditions use 784--256--128--10 ReLU MLPs, batch 128,
feedback scale 1, no weight decay and no further augmentation. Core arms use
BN; FD uses decorrelation. Before testing we predict accuracy and loss gains
from A under input nuisance, a larger accuracy gain than under clean inputs,
and small clean/label-noise effects. Neither constructed dataset meets the
full prediction. CE worsens significantly after correction in both, and
the original background task provides no corrected benefit. The complete
six-method endpoints below retain these reversals. The task description
alone does not establish the spectral relation assumed by the theory.

\begin{table}[!htbp]
\centering\small
\setlength{\tabcolsep}{4pt}
\begin{tabular}{@{}lrrrrrr@{}}
\toprule
Rule & Accuracy (\%) & CE & $\eta$ & Damp. & Updates & TFLOPs\\
\midrule
\multicolumn{7}{l}{\textit{MNIST clean}}\\
DFA & $94.80\pm0.18$ & 0.1732 & 0.001 & -- & 30,786 & 3.69\\
BP & $95.83\pm0.28$ & 0.1465 & 0.001 & -- & 30,501 & 3.89\\
FD-DFA & $94.83\pm0.24$ & 0.1907 & 0.0003 & -- & 23,146 & 7.68\\
A-nDFA & $94.37\pm0.37$ & 0.1889 & 0.001 & 3 & 17,248 & 3.93\\
FOOF-BP & $95.96\pm0.23$ & 0.1567 & 3 & 3 & 27,915 & 17.90\\
FOOF-DFA & $95.33\pm0.14$ & 0.1721 & 0.3 & 3 & 28,084 & 17.79\\
\multicolumn{7}{l}{\textit{MNIST label noise}}\\
DFA & $90.48\pm0.28$ & 0.4837 & 0.0001 & -- & 29,180 & 3.50\\
BP & $90.02\pm0.71$ & 0.4864 & 0.0003 & -- & 30,700 & 3.91\\
FD-DFA & $91.06\pm0.89$ & 0.4354 & 0.0003 & -- & 21,290 & 7.07\\
A-nDFA & $89.41\pm0.42$ & 0.4966 & 0.0003 & 3 & 18,295 & 4.17\\
FOOF-BP & $90.50\pm0.21$ & 0.4837 & 0.03 & 3 & 28,007 & 17.96\\
FOOF-DFA & $79.02\pm3.17$ & 0.9634 & 3 & 3 & 28,056 & 17.77\\
\multicolumn{7}{l}{\textit{MNIST mixture}}\\
DFA & $79.07\pm0.35$ & 0.6731 & 0.0001 & -- & 30,802 & 3.69\\
BP & $79.18\pm0.37$ & 0.6638 & 0.0001 & -- & 30,968 & 3.95\\
FD-DFA & $76.95\pm1.17$ & 0.7892 & 0.0001 & -- & 22,931 & 7.61\\
A-nDFA & $78.31\pm0.83$ & 0.7035 & 0.001 & 3 & 18,529 & 4.22\\
FOOF-BP & $81.49\pm0.47$ & 0.7268 & 0.3 & 0.3 & 28,091 & 18.01\\
FOOF-DFA & $79.01\pm0.32$ & 0.7072 & 0.03 & 0.3 & 28,160 & 17.84\\
\bottomrule
\end{tabular}
\caption{\textbf{Image-benchmark confirmation.} All six prespecified families and conditions are retained. Accuracy is mean $\pm$ sample SD; CE, update counts and estimated linear-algebra work are means over ten seeds. All runs use the same 60-second RTX PRO 6000 budget and validation-CE selection. Damping is relative for A and absolute for FOOF. FD uses decorrelation rate $10^{-5}$.}
\label{tab:joint_benchmarks_1}
\end{table}

\begin{table}[!htbp]
\centering\small
\setlength{\tabcolsep}{4pt}
\begin{tabular}{@{}lrrrrrr@{}}
\toprule
Rule & Accuracy (\%) & CE & $\eta$ & Damp. & Updates & TFLOPs\\
\midrule
\multicolumn{7}{l}{\textit{Fashion clean}}\\
DFA & $84.91\pm0.25$ & 0.4466 & 0.0001 & -- & 30,857 & 3.70\\
BP & $85.67\pm0.27$ & 0.4263 & 0.0001 & -- & 30,999 & 3.95\\
FD-DFA & $82.39\pm1.59$ & 0.7129 & 0.0003 & -- & 23,178 & 7.70\\
A-nDFA & $84.99\pm0.25$ & 0.4534 & 0.001 & 3 & 18,556 & 4.23\\
FOOF-BP & $85.08\pm0.19$ & 0.4356 & 0.03 & 3 & 28,107 & 18.02\\
FOOF-DFA & $84.44\pm0.41$ & 0.4951 & 0.03 & 0.3 & 26,025 & 16.49\\
\multicolumn{7}{l}{\textit{Fashion label noise}}\\
DFA & $81.73\pm0.39$ & 0.6872 & 0.0001 & -- & 28,558 & 3.42\\
BP & $82.50\pm0.75$ & 0.6704 & 0.0001 & -- & 30,285 & 3.86\\
FD-DFA & $76.16\pm1.40$ & 0.8341 & 0.0003 & -- & 23,130 & 7.68\\
A-nDFA & $81.85\pm0.21$ & 0.6797 & 0.0001 & 0.3 & 18,438 & 4.20\\
FOOF-BP & $82.17\pm0.37$ & 0.6733 & 0.03 & 3 & 27,789 & 17.82\\
FOOF-DFA & $75.82\pm2.91$ & 0.8441 & 0.03 & 0.3 & 27,626 & 17.50\\
\multicolumn{7}{l}{\textit{Fashion mixture}}\\
DFA & $71.47\pm0.41$ & 0.8266 & 0.0001 & -- & 30,695 & 3.68\\
BP & $70.21\pm0.83$ & 0.8660 & 0.0001 & -- & 30,534 & 3.89\\
FD-DFA & $65.50\pm0.91$ & 1.0977 & 0.0003 & -- & 23,022 & 7.64\\
A-nDFA & $72.02\pm0.43$ & 0.8419 & 0.0003 & 3 & 18,564 & 4.23\\
FOOF-BP & $71.30\pm0.29$ & 0.8685 & 0.03 & 0.3 & 27,884 & 17.88\\
FOOF-DFA & $70.87\pm0.75$ & 0.9376 & 0.03 & 0.3 & 28,146 & 17.83\\
\bottomrule
\end{tabular}
\caption{\textbf{Image-benchmark confirmation (continued).} All six prespecified families and conditions are retained. Accuracy is mean $\pm$ sample SD; CE, update counts and estimated linear-algebra work are means over ten seeds. All runs use the same 60-second RTX PRO 6000 budget and validation-CE selection. Damping is relative for A and absolute for FOOF. FD uses decorrelation rate $10^{-5}$.}
\label{tab:joint_benchmarks_2}
\end{table}

\begin{table}[!htbp]
\centering\small
\setlength{\tabcolsep}{4pt}
\begin{tabular}{@{}lrrrrrr@{}}
\toprule
Rule & Accuracy (\%) & CE & $\eta$ & Damp. & Updates & TFLOPs\\
\midrule
\multicolumn{7}{l}{\textit{Original MNIST backgrounds}}\\
DFA & $73.85\pm0.58$ & 0.8137 & 0.0001 & -- & 30,583 & 3.66\\
BP & $72.95\pm0.50$ & 0.8462 & 0.0001 & -- & 31,044 & 3.96\\
FD-DFA & $71.20\pm1.22$ & 1.0307 & 0.0003 & -- & 22,788 & 7.57\\
A-nDFA & $73.99\pm0.14$ & 0.8103 & 0.0001 & 3 & 18,548 & 4.23\\
FOOF-BP & $72.80\pm0.19$ & 0.8506 & 0.03 & 3 & 28,084 & 18.00\\
FOOF-DFA & $71.49\pm0.63$ & 0.9641 & 0.03 & 0.3 & 27,340 & 17.32\\
\bottomrule
\end{tabular}
\caption{\textbf{Image-benchmark confirmation (continued).} All six prespecified families and conditions are retained. Accuracy is mean $\pm$ sample SD; CE, update counts and estimated linear-algebra work are means over ten seeds. All runs use the same 60-second RTX PRO 6000 budget and validation-CE selection. Damping is relative for A and absolute for FOOF. FD uses decorrelation rate $10^{-5}$.}
\label{tab:joint_benchmarks_3}
\end{table}

\FloatBarrier

\subsection{Secondary synthetic factorial}

The secondary synthetic FOOF interaction is larger without BN (11.04 points,
individual 95\% interval $[9.84,12.24]$) than with BN (3.84 points,
$[-0.06,7.73]$); it is outside the primary family and does not establish a
general mechanism specific to DFA.

\begin{table}[!htbp]
\centering\small
\ifdefined\ARXIV\setlength{\tabcolsep}{6pt}\else\setlength{\tabcolsep}{4pt}\fi
\begin{tabular}{@{}llrrrr@{}}
\toprule
Rule & BN & Accuracy (\%) & CE & $\eta$ & $\lambda$\\
\midrule
BP (SGD) & Off & $15.07\pm0.56$ & $2.5119$ & $0.001$ & --\\
FOOF-BP & Off & $23.37\pm1.75$ & $2.3998$ & $0.003$ & $0.3$\\
BP (AdamW) & Off & $15.24\pm0.90$ & $2.5493$ & $3\times10^{-5}$ & --\\
DFA (SGD) & Off & $16.42\pm1.21$ & $2.8564$ & $0.001$ & --\\
FOOF-DFA & Off & $35.75\pm1.89$ & $2.2066$ & $0.03$ & $0.3$\\
DFA (AdamW) & Off & $15.58\pm1.00$ & $2.8502$ & $3\times10^{-5}$ & --\\
BP (SGD) & On & $14.26\pm0.48$ & $2.2351$ & $0.001$ & --\\
FOOF-BP & On & $19.67\pm1.72$ & $2.1597$ & $0.003$ & $0.3$\\
BP (AdamW) & On & $15.19\pm0.93$ & $2.2401$ & $3\times10^{-5}$ & --\\
DFA (SGD) & On & $19.70\pm4.25$ & $2.2323$ & $0.001$ & --\\
FOOF-DFA & On & $28.94\pm3.07$ & $2.1097$ & $0.003$ & $0.3$\\
DFA (AdamW) & On & $15.61\pm1.81$ & $2.2513$ & $0.0001$ & --\\
A-nDFA & Off & $34.39\pm2.02$ & $2.0405$ & $0.0001$ & $0.03$\\
A-nDFA & On & $41.72\pm2.13$ & $1.6317$ & $0.0003333$ & $0.3$\\
\bottomrule
\end{tabular}
\caption{\textbf{Secondary synthetic FOOF factorial.} Accuracy is mean $\pm$ sample SD over ten paired seeds; CE is the mean. The nuisance-dominant synthetic condition uses 20 seconds of measured learning work per run and validation-CE checkpoint selection (Appendix~\ref{app:joint_confirmation}). BN denotes batch normalization. Damping $\lambda$ is absolute; dashes denote inactive factors. A-nDFA uses SGD with momentum 0.9; the SGD/FOOF pairs have no momentum.}
\label{tab:joint_synthetic}
\end{table}

\FloatBarrier

\subsection{Seed-level primary contrasts}

Tables~\ref{tab:joint_seed_0}--\ref{tab:joint_seed_5} list every paired value, with IDs matching Table~\ref{tab:joint_primary}. Display rounding does not enter the statistical calculations.

\begin{table}[!htbp]
\centering\small
\setlength{\tabcolsep}{4pt}
\begin{tabular}{@{}rrrrrr@{}}
\toprule
ID & 940101 & 940102 & 940103 & 940104 & 940105\\
\midrule
1 & $+1.770$ & $+2.640$ & $+2.120$ & $+2.380$ & $+3.130$\\
2 & $+2.600$ & $+3.170$ & $+2.460$ & $+3.370$ & $+3.290$\\
3 & $+2.910$ & $+3.910$ & $+4.170$ & $+3.840$ & $+3.510$\\
4 & $+0.820$ & $+1.100$ & $-0.610$ & $-0.440$ & $-1.370$\\
5 & $+3.380$ & $+3.470$ & $+2.950$ & $+3.080$ & $+3.230$\\
6 & $+3.010$ & $+2.880$ & $+2.740$ & $+3.590$ & $+3.480$\\
7 & $+0.990$ & $+1.480$ & $+0.980$ & $+1.070$ & $+1.990$\\
8 & $-0.200$ & $-0.380$ & $-1.300$ & $+0.300$ & $+0.780$\\
9 & $+0.270$ & $-0.330$ & $+0.460$ & $+0.560$ & $-0.290$\\
10 & $-0.0008$ & $+0.0017$ & $+0.0165$ & $+0.0135$ & $-0.0018$\\
11 & $+0.280$ & $-0.330$ & $+0.370$ & $+0.380$ & $+0.240$\\
12 & $-0.0026$ & $+0.0049$ & $+0.0181$ & $+0.0084$ & $+0.0070$\\
13 & $+0.3771$ & $+0.3770$ & $+0.3774$ & $+0.3723$ & $+0.3795$\\
14 & $-0.700$ & $-2.940$ & $-0.710$ & $+0.220$ & $-0.520$\\
15 & $-0.0238$ & $-0.0921$ & $-0.0381$ & $-0.0068$ & $-0.0277$\\
16 & $-0.450$ & $-2.060$ & $-0.450$ & $+0.430$ & $+0.210$\\
17 & $+0.470$ & $+1.120$ & $+0.780$ & $-0.030$ & $+1.020$\\
18 & $-0.0089$ & $-0.0033$ & $-0.0106$ & $-0.0345$ & $-0.0171$\\
19 & $+0.760$ & $+1.050$ & $+1.320$ & $+0.350$ & $+0.760$\\
20 & $+0.108$ & $+0.364$ & $-0.294$ & $-0.344$ & $+0.216$\\
21 & $-0.0036$ & $+0.0062$ & $-0.0096$ & $-0.0074$ & $+0.0062$\\
\bottomrule
\end{tabular}
\caption{\textbf{Seed-level primary contrasts.} IDs and units match Table~\ref{tab:joint_primary}. These are training replications on fixed real-data splits, not independent datasets.}
\label{tab:joint_seed_0}
\end{table}

\begin{table}[!htbp]
\centering\small
\setlength{\tabcolsep}{4pt}
\begin{tabular}{@{}rrrrrr@{}}
\toprule
ID & 940106 & 940107 & 940108 & 940109 & 940110\\
\midrule
1 & $+2.300$ & $+1.940$ & $+1.710$ & $+2.160$ & $+2.480$\\
2 & $+2.920$ & $+2.030$ & $+2.560$ & $+2.760$ & $+2.400$\\
3 & $+3.930$ & $+3.650$ & $+3.030$ & $+3.590$ & $+3.000$\\
4 & $-2.680$ & $+0.150$ & $+1.400$ & $+0.580$ & $-1.240$\\
5 & $+3.270$ & $+3.760$ & $+3.040$ & $+2.810$ & $+3.060$\\
6 & $+3.290$ & $+3.760$ & $+3.210$ & $+2.990$ & $+2.310$\\
7 & $+1.140$ & $+1.080$ & $+1.470$ & $+0.900$ & $+0.760$\\
8 & $+0.130$ & $-0.480$ & $+1.010$ & $-0.170$ & $+0.860$\\
9 & $+0.730$ & $+0.490$ & $+1.390$ & $-0.020$ & $+0.580$\\
10 & $+0.0119$ & $+0.0148$ & $+0.0072$ & $+0.0097$ & $-0.0019$\\
11 & $+0.700$ & $-0.150$ & $+1.610$ & $+0.200$ & $+0.830$\\
12 & $+0.0172$ & $+0.0068$ & $+0.0258$ & $+0.0040$ & $-0.0032$\\
13 & $+0.3755$ & $+0.3755$ & $+0.3829$ & $+0.3817$ & $+0.3738$\\
14 & $-0.810$ & $-0.980$ & $+0.150$ & $-0.510$ & $-0.750$\\
15 & $-0.0238$ & $-0.0331$ & $-0.0117$ & $-0.0170$ & $-0.0292$\\
16 & $-1.050$ & $-0.380$ & $+0.860$ & $+0.510$ & $-0.820$\\
17 & $-0.420$ & $-0.170$ & $+0.840$ & $+1.130$ & $+0.740$\\
18 & $-0.0323$ & $-0.0167$ & $-0.0037$ & $-0.0158$ & $-0.0101$\\
19 & $-0.580$ & $-0.460$ & $+0.340$ & $+0.830$ & $+0.300$\\
20 & $-0.760$ & $+1.248$ & $+0.736$ & $+0.294$ & $-0.138$\\
21 & $-0.0114$ & $+0.0223$ & $+0.0183$ & $+0.0092$ & $+0.0032$\\
\bottomrule
\end{tabular}
\caption{\textbf{Seed-level primary contrasts (continued).} IDs and units match Table~\ref{tab:joint_primary}. These are training replications on fixed real-data splits, not independent datasets.}
\label{tab:joint_seed_5}
\end{table}

\FloatBarrier

\FloatBarrier
\section{Linear analysis and simulation protocols}
\label{app:theory}

Write $C_{\ell-1}:=C_{A,\ell}$ for the activity second moment.
To make its role in Eq.~\ref{eq:ndfa} precise, consider the
minimal setting in which BP, DFA, and nDFA differ. Let the input
covariance be positive definite and the inputs satisfy
$x \sim \mathcal{N}(0, \Sigma)$, let the target be a linear map
$y = W^\star x$, and consider a single hidden layer
\begin{equation}
    h = W^{(1)} x,
    \qquad
    \hat y = W^{(2)} h,
    \qquad
    \Loss = \tfrac{1}{2}\,\E_x\|\hat y - y\|^2 .
\end{equation}
Write the residual operator $\varepsilon := W^{(2)} W^{(1)} - W^\star$ and the
per-sample output error $e(x) = \varepsilon x$. Let
$B \in \mathbb{R}^{n_h \times n_y}$ be the fixed feedback matrix. The expected
per-step update directions in $W^{(1)}$ are
\begin{align}
    g^{\mathrm{BP}}
    &= -\,\E_x\!\big[(W^{(2)})^\top e(x)\,x^\top\big]
    = -(W^{(2)})^\top \varepsilon\, \Sigma,
    \label{eq:gbp}\\
    g^{\mathrm{DFA}}
    &= -\,\E_x\!\big[B\, e(x)\,x^\top\big]
    = -B\,\varepsilon\,\Sigma,
    \label{eq:gdfa}\\
    g^{\mathrm{nDFA}}
    &= g^{\mathrm{DFA}}\,(\Sigma + \lambda_A I)^{-1}
    = -B\,\varepsilon\,\Sigma\,(\Sigma+\lambda_A I)^{-1}.
    \label{eq:gndfa}
\end{align}
The updates differ in (a) which left projection of $\varepsilon$ is used
($(W^{(2)})^\top$ for BP, $B$ for DFA/nDFA), and (b) which input-statistics
weight is applied on the right ($\Sigma$ for BP and DFA, and
$\Sigma(\Sigma+\lambda_A I)^{-1}$ for nDFA). As $\lambda_A \to 0$ the right
factor of nDFA approaches the identity. In this population linear model,
conditioning removes the input-statistics weighting $\Sigma$ that DFA inherits
from BP while leaving the feedback direction untouched.

\paragraph{The aligned regime.}
At initialization, when the forward weights are independent of the
zero-mean feedback draw, Eq.~\ref{eq:gdfa} gives $\E_B[g^{\mathrm{DFA}}]=0$.
Useful hidden updates can emerge as \emph{feedback alignment} makes the
forward weights correlate with $B$, so that the effective error projection
$B$ acquires a positive component along $(W^{(2)})^\top$
\citep{lillicrap2016feedback,refinetti2021align}. We state the spectral
proposition under the explicit perfect-alignment condition
$B=\alpha(W^{(2)})^\top$ with $\alpha>0$, where
$g^{\mathrm{DFA}}\!\propto g^{\mathrm{BP}}$. Scalar output alone does not imply
this alignment. We hold the readout fixed when deriving the training dynamics.
More general multiclass alignment
can be decomposed into an aligned component plus residual feedback components;
the same input-side preconditioner acts on all components, but descent then
depends on the aligned component being large enough.

\paragraph{What conditioning preconditions.}
The loss is quadratic in $W^{(1)}$ with Hessian (in row-major
$\operatorname{vec}(W^{(1)})$ coordinates, so the activity factor appears on the
right of the Kronecker product)
\begin{equation}
    H \;=\; (W^{(2)\top} W^{(2)}) \otimes \Sigma ,
    \label{eq:hessian}
\end{equation}
so a Kronecker-factored Newton step contains an input-side factor
$\Sigma^{-1}$. Right-multiplication by $(\Sigma+\lambda_A I)^{-1}$ is a damped
approximation to that one factor; nDFA does not estimate or invert the
output-side Hessian factor $W^{(2)\top}W^{(2)}$. The same input factor appears
in a Kronecker-factored natural-gradient construction only under an explicitly
specified likelihood (for example homoscedastic Gaussian regression) and the
usual factorization approximation \citep{amari1998natural,martens2015kfac}.
For the nonlinear softmax classifiers below, nDFA is therefore best described
as a one-sided, curvature-motivated inverse-second-moment preconditioner, not as
an exact Newton or natural-gradient method.

\paragraph{Proposition 1 (input-factor spectral identity).}
Assume the explicitly aligned linear-Gaussian regime above with fixed readout,
population second moments, fixed feedback $B=\alpha(W^{(2)})^\top$ with
$\alpha>0$, and damping $\lambda_A>0$. Then input-side conditioning replaces the
per-eigendirection factor $\lambda_i$ of raw BP/DFA by
$\lambda_i/(\lambda_i+\lambda_A)$, reducing the input-side condition number from
$\kappa(\Sigma)$ to $\kappa_{\mathrm{nDFA}}$ in Eq.~\ref{eq:cond_reduction}. The
condition-number ratio is
$\rho(\Sigma,\lambda_A)=\kappa(\Sigma)/\kappa_{\mathrm{nDFA}}$. This is an exact
spectral identity for the stated population model. It is not an
iteration-complexity result for DFA: realized improvement additionally depends
on the residual feedback component, initialization, step size, stopping rule,
nonlinear gates, and the jointly estimated minibatch gradient and second moment.

\paragraph{Anisotropy controls the gain.}
The consequence is a statement about the \emph{conditioning of the training
dynamics}, not about the single-step inner product with $g^{\mathrm{BP}}$.
Decomposing $\Sigma = U\Lambda U^\top$ with
$\Lambda=\operatorname{diag}(\lambda_1,\dots,\lambda_d)$, the per-direction gain
of the aligned BP/DFA update on coordinate $i$ scales as $\lambda_i$, so the
input-side condition number of the update operator is $\kappa(\Sigma) =
\lambda_{\max}/\lambda_{\min}$. nDFA replaces this gain by
$\lambda_i/(\lambda_i+\lambda_A)$, giving
\begin{equation}
    \kappa_{\mathrm{nDFA}}
    \;=\;
    \frac{\max_i \lambda_i/(\lambda_i+\lambda_A)}{\min_i \lambda_i/(\lambda_i+\lambda_A)}
    \;\xrightarrow[\lambda_A\to 0]{}\; 1 ,
    \qquad
    \rho(\Sigma,\lambda_A)
    :=
    \frac{\kappa(\Sigma)}{\kappa_{\mathrm{nDFA}}}
    \label{eq:cond_reduction}
\end{equation}
the condition-number reduction factor. Since standard first-order GD bounds on
a quadratic scale as $O(\kappa\log\tfrac1\epsilon)$, $\rho(\Sigma,\lambda_A)$ is
the ratio of the corresponding worst-case condition-number bounds. It therefore
summarizes the maximum improvement suggested by that analysis, not an exact
upper bound on the observed iteration ratio. Whether this spectral improvement
translates into faster training depends on \emph{where the task lies} in
$\Sigma$'s spectrum:

\textbf{(i) Input isotropy.} If $\Sigma=\sigma^2 I$ then $\rho=1$:
activity conditioning reduces to scalar rescaling and provides no input-side
condition-number improvement. Readout-side curvature need not be isotropic.

\textbf{(ii) A scoped empirical hypothesis.}
Preconditioning equalizes the per-direction gain, so it accelerates a task direction
in proportion to how \emph{slow} the unconditioned update was there---i.e., to
how small that direction's variance $\lambda_i$ is. Decompose the target into
$\Sigma$'s eigenbasis. If the task lies in the \emph{low-variance} directions of
$\Sigma$---equivalently, the high-variance directions are task-irrelevant
nuisance---the unconditioned update crawls along the task directions and nDFA
can realize a speed-up close to the condition-number ratio $\rho$. If instead the task lies in the
\emph{high-variance} directions, the unconditioned update is already fast there
and nDFA gives little benefit. Holding $\kappa(\Sigma)$ fixed and only moving the
target's energy between low- and high-variance eigendirections changes the
realized nDFA speed-up by more than an order of magnitude
(Fig.~\ref{fig:theory_conditioning}D, at fixed $\kappa=50$). This idealized
calculation motivates the design principle behind
Eq.~\ref{eq:ndfa}: noise and nuisance help conditioning not by raising $\kappa$
per se, but by loading the high-variance directions with task-irrelevant
variance, which pushes the task into the low-variance subspace where preconditioning
pays off. $\kappa(\Sigma)$ bounds the spectral improvement factor; it is
not a general bound on the observed training-speed ratio.

\textbf{(iii) Excessive damping.} As $\lambda_A \to \infty$ the input-statistics
weight $\Sigma(\Sigma+\lambda_A I)^{-1} \to \lambda_A^{-1}\Sigma$, hence
$g^{\mathrm{nDFA}} \to \lambda_A^{-1} g^{\mathrm{DFA}}$ and $\rho \to 1$: heavy
damping reverts nDFA to a scaled DFA step. In population, smaller damping removes
the input factor more completely. In the implemented minibatch algorithm,
damping can still be required because the empirical factor may be singular or
poorly conditioned and because label noise, nonlinear gates, nonstationarity,
and finite precision perturb the idealized model. We therefore treat damping as
an empirical rule parameter rather than claiming a universal optimum;
\S\ref{app:damping_theory} makes this scope explicit.

\paragraph{Population simulation protocol.}
\label{app:linear_simulation}
Figure~\ref{fig:theory_conditioning}D compares task orientations at fixed
$\kappa=50$. Inputs have dimension 12, the hidden layer has 10 units, and
the output has dimension 4. The population moment has eigenvalues spaced
geometrically from 1 to 50 and a random orthogonal eigenbasis. A fixed
readout has independent Gaussian entries with variance $1/10$, and
$B=W^{(2)\top}$. The target is a Gaussian linear map on either the four
lowest-variance or four highest-variance eigenvectors, scaled by $1/\sqrt{12}$.
Initial first-layer weights have independent Gaussian entries with standard
deviation 0.01; the initial residual can therefore occupy other modes too.
Activity damping is $10^{-3}$, and loss is the population quantity
$L=\tr(\varepsilon\Sigma\varepsilon^\top)/2$.

For each rule and initialization we search 24 learning rates spaced
geometrically from $10^{-4}$ to 5, reporting the fewest updates to
$L/L_0<10^{-6}$. Each candidate has a 4,000-update limit and stops if loss is
nonfinite or exceeds $10^3L_0$. The figure averages seeds 0, 1 and 2.
This separately rate-selected illustration shows dependence on task
orientation; it is not a matched-computation comparison or a condition-number
sweep. The generator is included in the public reproduction code.

\paragraph{Interpreting the projected step.}
The projected BP-step ratio
$\Pi(g,g^{\mathrm{BP}})=\langle g,g^{\mathrm{BP}}\rangle/\|g^{\mathrm{BP}}\|^2$
measures instantaneous descent in BP units. At fixed norm it is maximized by
the BP direction, and it also changes under scalar rescaling of an update.
Preconditioning can lower this projection while improving convergence through
more balanced learning rates. We therefore interpret $\Pi$ alongside update
norms and trajectories, rather than as an objective to maximize.

\paragraph{Scope and what this does not show.}
This is a linearized statement about the post-alignment regime with a
Gaussian-input model and population covariances. It does not bound the full
nonlinear trajectory, model forward-weight alignment to fixed $B$, or analyze
convolutional weight sharing; finite-sample $\widehat\Sigma$ estimation and
damping are discussed in \S\ref{app:damping_theory} below. The preceding curvature
calculation applies to the stated fixed-readout regression model and specified
likelihood. In the nonlinear classifiers, the DFA-error second moment is not
generally a Fisher factor, so the conditioned rule is curvature-motivated
rather than an exact natural-gradient update. It does give a falsifiable prediction:
the improvement over raw DFA at a fixed training budget should grow with the fraction
of the layer's second-moment energy that lies in task-irrelevant (nuisance)
directions. The bound $\rho\le\kappa(\Sigma)$ concerns the population
input-factor condition number, not realized nonlinear speed or accuracy, while
any advantage over a tuned BP baseline should
disappear under isotropy or when the high-variance directions are themselves
task-relevant. The synthetic regimes that gain most over raw DFA from
conditioning---nuisance-dominant, low-sample/noisy, mixed-context---are
precisely those constructed so that the high-variance hidden directions carry
nuisance, while the task-aligned control places the task in the high-variance
directions and removes the BP advantage (a slight loss to BP), even though both
have anisotropic hidden statistics (Appendix~\ref{app:anisotropy_predictor}).

\subsection{Finite-sample damping: scope of the implemented estimator}
\label{app:damping_theory}

For a minibatch with activation matrix \(X\) and residual matrix \(E\), the
implemented linear-model update uses the \emph{same samples} in its raw
outer-product gradient and its activity factor. In the aligned noiseless case,
\[
    \widehat G=-B\varepsilon\widehat C,\qquad
    \widehat G(\widehat C+\lambda_A I)^{-1}
    =-B\varepsilon\widehat C(\widehat C+\lambda_A I)^{-1},
    \qquad
    \widehat C=X^\top X/n .
\]
Consequently, at \(\lambda_A=0\) and full-rank \(\widehat C\), the empirical
factor cancels exactly. The estimator is not
\(\Sigma(\widehat C+\lambda_A I)^{-1}\), which would combine a population
gradient numerator with an independently sampled denominator. Covariance
sampling error by itself therefore does not imply inverse blow-up or a
universal interior optimum for the batch algorithm.

Damping nevertheless remains necessary when \(\widehat C\) is singular or
ill-conditioned and can be useful under label or gradient noise, nonlinear
gating, changing representations, finite precision, and stochastic
optimization \citep{vershynin2018hdp}. These effects depend jointly on the gradient and factor
estimates, so we select \(\lambda_A\) empirically and do not claim a closed-form
optimal damping law. Risk and stability bounds for this joint estimator remain open. In particular, this
paper draws no depth-, width-, or ImageNet-scaling conclusion from a covariance-concentration formula.
\subsection{Mode timing: generalization dynamics under conditioning}
\label{app:mode_timing}

The spectral identity above concerns optimization geometry. This subsection
gives a separate, illustrative calculation of test-error trajectories in the
same aligned linear-Gaussian model with label noise. It is not a model of the
pre-alignment DFA trajectory or evidence for the nonlinear-network results.
Within this restricted model, conditioning changes when each eigendirection is
learned without changing the least-squares destination.

\paragraph{Finite-sample setup.}
Keep the model of Eqs.~\ref{eq:gbp}--\ref{eq:gndfa} with scalar output and
explicit alignment $B=\alpha (W^{(2)})^\top$, $\alpha>0$, and add label
noise and a finite training set: $n$ samples $x_\mu\sim\mathcal N(0,\Sigma)$,
labels $y_\mu = w^{\star\top}x_\mu + \xi_\mu$ with
$\xi_\mu\sim\mathcal N(0,\sigma^2)$ i.i.d. Collect
$X\in\mathbb{R}^{n\times d}$, $\widehat\Sigma = X^\top X/n$, and let the
first layer follow gradient flow on the \emph{empirical} loss with $W^{(2)}$
fixed on the timescale analyzed (the post-alignment regime; the acquisition of forward-weight
alignment to fixed $B$ is outside scope, as in Proposition~1). The composite
predictor $w:=(W^{(2)}W^{(1)})^\top$ then obeys
\begin{equation}
    \dot w
    \;=\;
    -\,\eta_{\mathrm{eff}}\;M
    \Big[\widehat\Sigma\,(w-w^\star)-\tfrac{1}{n}X^\top\xi\Big],
    \qquad
    M=
    \begin{cases}
        I & \text{BP, aligned DFA},\\[2pt]
        (\widehat\Sigma+\lambda_A I)^{-1} & \text{nDFA},
    \end{cases}
    \label{eq:modetiming_flow}
\end{equation}
with $\eta_{\mathrm{eff}}=\eta\|W^{(2)}\|^2$ for BP and
$\eta\alpha\|W^{(2)}\|^2$ for aligned DFA: \emph{BP and aligned DFA share the
same clock up to a scalar}. More generally, a scalar-output composite predictor
obeys this flow for arbitrary fixed $B$ when $W^{(2)}B>0$, with
$\eta_{\mathrm{eff}}=\eta W^{(2)}B$; this weaker condition does not make the
first-layer DFA update proportional to BP. ($M$ and $\widehat\Sigma$ commute, so
Eq.~\ref{eq:modetiming_flow} is the composite form of right-multiplying the
first-layer update by the damped inverse, Eq.~\ref{eq:gndfa}, here with the
plug-in $\widehat\Sigma$; it is held fixed here to isolate timing.) For
$n\ge d$, $\widehat\Sigma\succ0$ a.s.\ and every rule has the \emph{same}
unique fixed point, the least-squares estimator
\begin{equation}
    \widehat w
    \;=\;
    w^\star+\widehat\Sigma^{-1}\tfrac{1}{n}X^\top\xi ,
    \qquad
    \operatorname{Cov}\big(\widehat w - w^\star\mid X\big)
    =\tfrac{\sigma^2}{n}\,\widehat\Sigma^{-1}.
    \label{eq:modetiming_fixedpoint}
\end{equation}
In the eigenbasis of $\widehat\Sigma$ (eigenvalues $\hat\lambda_i$), with
$w(0)=0$,
\begin{equation}
    w_i(t) \;=\; \widehat w_i\,\big(1-e^{-r_i t}\big),
    \qquad
    r_i =
    \begin{cases}
        \eta_{\mathrm{eff}}\,\hat\lambda_i & \text{BP/DFA},\\[2pt]
        \eta_{\mathrm{eff}}\,\dfrac{\hat\lambda_i}{\hat\lambda_i+\lambda_A} &
        \text{nDFA}.
    \end{cases}
    \label{eq:modetiming_rates}
\end{equation}
(Units: raw $\eta_{\mathrm{eff}}$ carries $[\mathrm{variance}\cdot
\mathrm{time}]^{-1}$, nDFA's carries $[\mathrm{time}]^{-1}$; the minimum risk along each trajectory and within-rule mode-timing
ratios are invariant under global time rescaling. Absolute flow times use the
stated effective learning rate and are not computational-cost comparisons.)

\paragraph{Exact empirical risk and the population-spectrum surrogate.}
Define the excess test risk $R(t)=\|w(t)-w^\star\|_\Sigma^2$ and
$A_t=\exp(-\eta_{\mathrm{eff}}M\widehat\Sigma t)$. Conditional on the design,
\begin{equation}
\begin{aligned}
\E_\xi[R(t)\mid X]
 &=w^{\star\top}A_t^\top\Sigma A_t w^\star\\
 &\quad+\frac{\sigma^2}{n}\operatorname{tr}
 \!\left[\Sigma(I-A_t)\widehat\Sigma^{-1}(I-A_t)^\top\right].
\end{aligned}
\label{eq:modetiming_exact_risk}
\end{equation}
The shared endpoint is $(\sigma^2/n)\operatorname{tr}(\Sigma\widehat\Sigma^{-1})$.
Averaging over Gaussian designs gives $\sigma^2d/(n-d-1)$ for $n>d+1$
\citep{belkin2020twomodels}. Thus $n\gtrsim d$ alone does not justify replacing
inverse empirical moments by population moments. The simpler calculation below
replaces the empirical eigensystem by that of $\Sigma$; it is a
population-spectrum surrogate, accurate only when the relevant sample moments
are sufficiently concentrated \citep{advani2020highdim}. In this surrogate,
\begin{equation}
    R_{\mathrm{pop}}(t)
    \;=\;
    \underbrace{\sum_i \lambda_i w_i^{\star 2}\,e^{-2 r_i t}}_{\text{signal
    fitting (decreasing)}}
    \;+\;
    \underbrace{\frac{\sigma^2}{n}\sum_i \big(1-e^{-r_i t}\big)^2}_{\text{noise
    fitting (increasing)}} .
    \label{eq:modetiming_risk}
\end{equation}
Here the rates use population eigenvalues. The empirical trajectories in the
numerical comparison instead use Eq.~\ref{eq:modetiming_exact_risk}.
Two readings of Eq.~\ref{eq:modetiming_risk} motivate the two-block calculation.
First, within this surrogate, every fitted mode eventually contributes
$\sigma^2/n$ of excess risk, independent of its eigenvalue. Second, the exact
fixed-point identity shows that all rules share a destination, so conditioning
changes the path rather than the least-squares estimator.

\paragraph{Proposition~2 (mode timing in the aligned linear model).}
Assume the aligned regime above, $n\ge d$, gradient flow on the empirical loss
from $w(0)=0$, Gaussian label noise $\sigma^2$, and the population-spectrum
approximation of Eq.~\ref{eq:modetiming_risk}. Define the time for mode $i$ to
reach a fraction $0<\theta<1$ of its destination,
$t_i(\theta)=\ln\!\big(\tfrac{1}{1-\theta}\big)/r_i$. Then:

\emph{(i) Timing.} BP and aligned DFA fit mode $i$ at rate
$r_i\propto\lambda_i$ (high-variance modes first: spectral bias), with
$t_i/t_j=\lambda_j/\lambda_i$; nDFA fits at rate
$r_i\propto\lambda_i/(\lambda_i+\lambda_A)$, with
\begin{equation}
    \frac{t_i}{t_j}
    =\frac{\lambda_j\,(\lambda_i+\lambda_A)}{\lambda_i\,(\lambda_j+\lambda_A)}
    \;\in\;\Big(1,\;\frac{\lambda_j}{\lambda_i}\Big)
    \quad\text{for }\lambda_j>\lambda_i,
    \label{eq:modetiming_ratio}
\end{equation}
monotonically compressed toward $1$ as $\lambda_A\downarrow0$: the fitting
order is preserved but the timing gaps collapse.

\emph{(ii) Destinations.} All three rules converge to the same
$\widehat w$ and hence the same final risk
(given by Eq.~\ref{eq:modetiming_exact_risk}; $\sigma^2d/n$ in the surrogate):
conditioning re-times learning without
changing what is learned.

\emph{(iii) Trajectory optimum and crossing.} Take the two-block spectrum:
$d_T$ task modes at $\lambda_T$ carrying all the signal
$S:=\sum_{i\in T}\lambda_i w_i^{\star2}$, and $d_N$ noise-only nuisance modes
at $\lambda_N$. Write $N:=\sigma^2 d_N/n$, $N_T:=\sigma^2 d_T/n$,
$N_{\mathrm{tot}}:=N+N_T$, task residual $u:=e^{-r_T t}$, and timing ratio
$\rho:=r_N/r_T$ (not the condition-number ratio of Proposition~1; within this
subsection $\rho$ always denotes the timing ratio), so that
\begin{equation}
    \E_\xi R
    = S u^2 + N_T (1-u)^2 + N\big(1-u^{\rho}\big)^2,
    \qquad
    \rho_{\mathrm{DFA}}=\frac{\lambda_N}{\lambda_T},
    \quad
    \rho_{\mathrm{nDFA}}(\lambda_A)
    =\frac{\lambda_N(\lambda_T+\lambda_A)}{\lambda_T(\lambda_N+\lambda_A)} .
    \label{eq:modetiming_twoblock}
\end{equation}
The best-achievable risk along the trajectory,
$R^\dagger(\rho):=\min_{t\ge0}\E_\xi R(t)$, is \emph{strictly increasing in}
$\rho$ whenever $\sigma>0$ and $S>0$. Consequently, for $\sigma,S>0$ and every
finite $\lambda_A>0$, conditioning strictly lowers $R^\dagger$ iff the task
modes are the slow ones ($\lambda_T<\lambda_N$), strictly raises it iff
$\lambda_T>\lambda_N$, and leaves it unchanged iff $\lambda_T=\lambda_N$ (it
is also unchanged in the degenerate cases $\sigma=0$ or $S=0$). In the strong-separation/strong-compression limit
($\rho_{\mathrm{DFA}}\to\infty$, $\rho_{\mathrm{nDFA}}\to1$),
which is ensured by $\lambda_N/\lambda_T\to\infty$ and
$\lambda_A/\lambda_T\to0$,
\begin{equation}
    R^\dagger_{\mathrm{DFA}}
    =\min\Big\{S,\;N+\tfrac{S N_T}{S+N_T}\Big\},
    \qquad
    R^\dagger_{\mathrm{nDFA}}
    =\frac{S\,N_{\mathrm{tot}}}{S+N_{\mathrm{tot}}},
    \qquad
    \Delta \approx \frac{\min(S,N)^2}{S+N}\ \ (N_T\ll N),
    \label{eq:modetiming_delta}
\end{equation}
To leading order when $N_T\ll N$, the relative improvement
$\Delta/R^\dagger_{\mathrm{DFA}}\le\tfrac12$ is
maximized when the signal power matches the accumulated noise floor,
$S=\sigma^2 d_N/n$. At finite damping, define
$f_\rho(u)=Su^2+N_T(1-u)^2+N(1-u^\rho)^2$. The total risk improvement in
the two-block surrogate is
\begin{equation}
    \Delta(\lambda_A)=
    \min_{0\le u\le1}f_{\rho_{\mathrm{DFA}}}(u)
    -\min_{0\le u\le1}f_{\rho_{\mathrm{nDFA}}(\lambda_A)}(u).
    \label{eq:modetiming_material}
\end{equation}
It includes signal bias, task-noise fitting and nuisance-noise fitting.
We evaluate the two complete minima numerically. Their difference depends
on signal and noise power as well as the rate ratio; the model does not
supply a universal task-eigenvalue threshold for useful damping.

\emph{Proof sketch.} (i)--(ii) are Eqs.~\ref{eq:modetiming_fixedpoint} and
\ref{eq:modetiming_rates} read off directly; the ratio bound in
Eq.~\ref{eq:modetiming_ratio} is monotone algebra. For (iii): in
Eq.~\ref{eq:modetiming_twoblock}, $f(u)=Su^2+N_T(1-u)^2+N(1-u^\rho)^2$ has
$f'(1)=2S>0$ and $f'(0^+)=-2N_T<0$ (for $\rho>1$; the boundary term vanishes).
For $0<\rho\le1$, the right derivative at zero is also negative
(or tends to $-\infty$), so a minimizer is interior for every $\rho>0$.
Since $\partial_\rho f=-2N(1-u^\rho)u^\rho\ln u>0$ there, comparing
$f_{\rho_1}$ and $f_{\rho_2}$ at a minimizer of the latter proves strict
increase whenever $\rho_1<\rho_2$, without assuming uniqueness.
For $\lambda_N>\lambda_T$, $\rho_{\mathrm{nDFA}}$ increases from $1$ to
$\rho_{\mathrm{DFA}}$ as damping increases; when $\lambda_N<\lambda_T$, it
decreases from $1$ to $\rho_{\mathrm{DFA}}$. The sign claims follow.
Eq.~\ref{eq:modetiming_delta} is
direct minimization at $\rho\in\{1,\infty\}$ ($u^{\infty}=0$ for $u<1$);
Eq.~\ref{eq:modetiming_material} compares the two complete minima. \hfill$\square$

\begin{figure}[t]
    \centering
    \includegraphics[width=\textwidth]{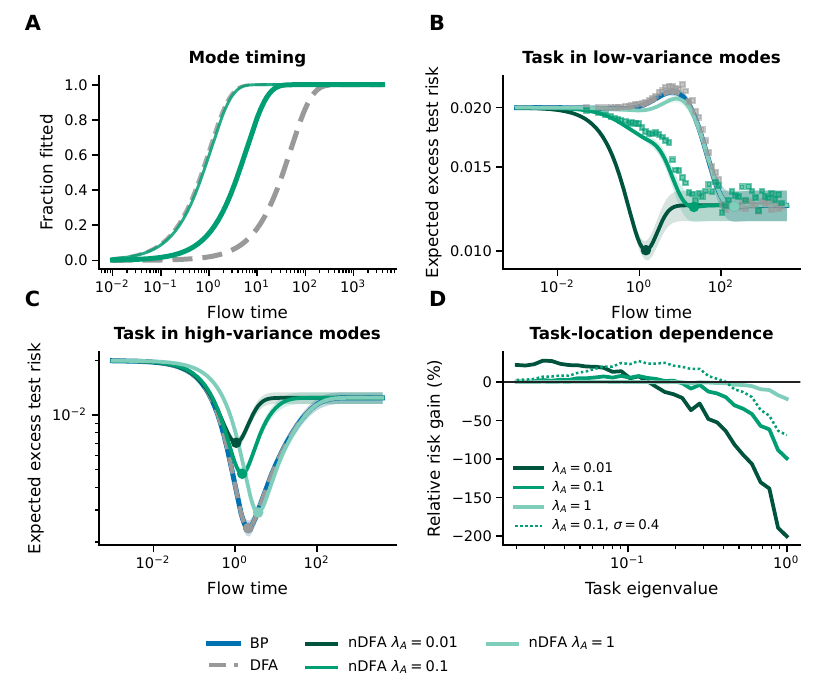}
    \caption{\textbf{Mode-timing illustration (Proposition~2).} \textbf{A,} Per-mode
    fitting curves in the aligned linear model ($d{=}32$, $\kappa{=}50$, log-spaced eigenvalues from 0.02 to 1): raw
    DFA/BP fit the high-variance nuisance mode $\kappa{=}50\times$ earlier
    than the low-variance task mode; conditioning compresses the gap to
    $\rho_{\mathrm{nDFA}}{=}5.5$ at $\lambda_A{=}0.1$ while preserving the order. Thin curves mark nuisance and thick curves
    mark task. \textbf{B,} Expected excess test-risk trajectories with the task on the two
    lowest-variance directions ($\sigma{=}0.2$, $n{=}128$): raw DFA (grey
    dashed; coincides with BP, blue) absorbs the nuisance noise before
    fitting the task and shows no early-stopping gain in this example, while nDFA
    with $\lambda_A{<}\lambda_{\mathrm{task}}$ moves the trajectory optimum
    (dots) lower and, under the chosen clock, earlier; squares: minibatch SGD on the explicit two-layer
    network. \textbf{C,} Task-aligned control: the same re-timing strictly
    hurts as compression strengthens: a task-aligned reversal despite the
    same label noise. \textbf{D,} Relative gain $(\min_t R_{\mathrm{DFA}}-\min_t R_{\mathrm{nDFA}})/\min_t R_{\mathrm{DFA}}$ as the task mode
    sweeps the spectrum: the gain changes sign with task location, and the crossing shifts with
    $\sigma$ and $\lambda_A$.
    Bands in B--C are $\pm1$ SEM over 20 draws; D averages ten draws.}
    \label{fig:mode_timing}
\end{figure}

\paragraph{Interpretation and limitations.}
In the two-block model, conditioning helps the best point on the trajectory when
the task occupies the slower, lower-variance block and can hurt when the task
occupies the faster block. This is a statement about the model's rates and
early-stopping path, not a guarantee for finite-step nonlinear DFA. In
particular, the derivation assumes fixed perfect alignment, \(n\ge d\), an
invertible empirical covariance, Gaussian label noise, and a population-spectrum
approximation for risk. It does not model how feedback alignment is acquired.
The full-spectrum crossing and finite-damping expressions below are thus
simulation checks of the equations, not independent empirical validation of a
network-level law.
\paragraph{Numerical validation.}
Figure~\ref{fig:mode_timing} simulates the model exactly
(\texttt{analysis/validate\_mode\_timing.py}: $d{=}32$, $\kappa{=}50$,
log-spaced spectrum, $\sigma{=}0.2$, $n{=}128$, $S{=}0.02$ on the two
lowest-variance directions; both the closed-form gradient-flow solution on the
empirical loss and minibatch stochastic gradient descent (SGD) on a two-layer linear network with
fixed aligned feedback). The empirical-flow simulation verifies the stated dynamics rather than
providing an independent test of their assumptions. Panel A shows compressed
mode-timing gaps. In the nuisance-dominant example (panel B), strong
conditioning lowers the trajectory minimum (and reaches it earlier under the chosen clock); mild
conditioning provides little gain under this signal/noise balance. Minibatch
SGD on the explicit aligned network follows the same trend. The task-aligned
control reverses the ordering (panel C). Sweeping task location across the
spectrum changes the gain's sign, with a crossing that depends on both noise
and damping (panel D). The empirical full-spectrum trajectories and the
population two-block calculation are distinct models; their quantitative
gains need not coincide. The numerical check minimizes all
three terms of the two-block risk separately for each rate ratio.

\FloatBarrier
\section{Implementation, statistics and study provenance}
\label{app:implementation}
The comparisons share a fixed-feedback pathway but differ in
normalization, damping and selection. We distinguish these choices before
reporting their results; each study states its own departures below.
We first distinguish their credit signals. For plain SGD, backpropagation computes
\begin{equation}
    \delta_\ell^{\mathrm{BP}}
    =
    (W_{\ell+1}^{\top}\delta_{\ell+1}^{\mathrm{BP}})\odot\phi'(a_\ell),
    \qquad
    \Delta W_\ell^{\mathrm{BP}}
    =
    -\eta\,\delta_\ell^{\mathrm{BP}}h_{\ell-1}^{\top}.
\end{equation}
Feedback alignment replaces $W_{\ell+1}^{\top}$ by a fixed random matrix. DFA projects the output error directly as in Eq.~\ref{eq:dfa}. DRTP replaces the output error by a random projection of the target. Local auxiliary losses attach supervised heads to hidden layers. VNC and NMNC estimate credit from perturbation-output correlations; NMNC restricts perturbations to a low-dimensional activity manifold.

MLP activity and error factors use full local blocks. Statistics are
recomputed from the current minibatch, except for the specified FOOF averages. Fixed Gaussian feedback
may be rank-truncated, with its Frobenius norm preserved before applying the
feedback scale. The classifier uses the exact output gradient: softmax
cross-entropy in the broad benchmarks and one-vs-rest binary log loss in the
tanh confirmations. Biases retain their local derivatives.
The original MLPs have no BatchNorm unless specified. Diagnostic spectra use
centered covariances and should be distinguished from the uncentered moments
inverted during training. Damping and solve conventions are specified below.

\begin{center}
\fbox{\begin{minipage}{0.94\textwidth}
\textbf{Algorithm 1: conditioned DFA pseudo-gradient for a hidden layer.}
Given a minibatch, compute DFA hidden errors. Without normalization,
$\delta_{\ell i}=(B_\ell e_i)\odot\phi'(a_{\ell i})$, where $e_i$ is the
derivative of the \emph{unreduced} per-example loss. Include the local
normalization derivative when BN is present. Form
$G_\ell=|{\cal B}|^{-1}\sum_i\delta_{\ell i}h_{\ell-1,i}^{\top}$,
$C_A=|{\cal B}|^{-1}\sum_i h_{\ell-1,i}h_{\ell-1,i}^{\top}$, and
$C_E=|{\cal B}|^{-1}\sum_i\delta_{\ell i}\delta_{\ell i}^{\top}$.
Then choose activity $GP_A$, error $P_EG$, or K-nDFA $P_EGP_A$, with
$P_A=(C_A+\lambda_A I)^{-1}$ and $P_E=(C_E+\lambda_E I)^{-1}$.
When the protocol uses norm matching, rescale the chosen matrix to the
Frobenius norm of $G_\ell$ after all active factors, including the zero case
in Eq.~\ref{eq:normmatched}. Supply this pseudo-gradient to the base optimizer:
Plain SGD multiplies it by $-\eta$, whereas AdamW first updates its moments and
then applies its adaptive step and any specified decoupled weight decay.
Matching pseudo-gradient norms does not match final AdamW step norms.
The feedback matrix $B_\ell$ is fixed and random; no transposed forward weights
or backpropagated hidden errors are used.
\end{minipage}}
\end{center}

\subsection{Diagnostic definitions}
\label{app:diagnostics}
To distinguish update direction from magnitude, let $g$ denote a
local hidden-weight pseudo-gradient and $g^\star$ its BP counterpart, both
before the base optimizer's transformation. Inner
products and norms are Frobenius products over the weight entries. Then:
\begin{equation}
    \cos(g,g^\star)=
    \frac{\langle g,g^\star\rangle}{\|g\|\|g^\star\|},
    \qquad
    \Pi(g,g^\star)=
    \frac{\langle g,g^\star\rangle}{\|g^\star\|^2}.
\end{equation}
The projected ratio measures a first-order descent component in BP units;
it depends on magnitude as well as direction and is not a training objective.

Feedback rank measures the number of output-error directions preserved
by the broadcast matrix. The activity spectra and nuisance/task energy ratio
characterize the geometry on which that feedback acts
(Appendix~\ref{app:anisotropy_predictor}).

\subsection{Moments, damping, and optimization}

The scale of a moment determines how strongly its inverse changes an update.
We therefore specify the loss normalization and damping conventions before
describing the implementation.

For each hidden layer, activities and local errors are obtained from the
same minibatch and forward state. The output error is the derivative of the
unreduced per-example loss; DFA forms local teaching signals from this error.
A local signal computed from a mean loss is multiplied
by the actual batch size before forming its second moment. This convention
also applies to partial batches. We use uncentered moments unless a control
explicitly specifies centering. The A/E/K conditioning operation affects hidden
weight matrices only; the separately specified FOOF recipe also conditions
the readout. Biases and BN affine parameters retain their local
derivatives, while the output classifier receives its exact supervised
gradient. Fixed feedback matrices are not optimized.

The original MNIST/Fashion-MNIST studies use absolute damping, with separate
values in the units of each moment. The CIFAR-10 A/E/K studies use
\begin{equation}
\lambda_A=\max\{\rho_A\tr(C_A)/d_{\rm in},10^{-6}\},\qquad
\lambda_E=\max\{\rho_E\tr(C_E)/d_{\rm out},10^{-6}\},
\end{equation}
where $\tr$ denotes the matrix trace, $\rho_A,\rho_E$ are relative damping coefficients, and
$d_{\rm in},d_{\rm out}$ are the layer's input and output widths.
Writing $G$ for the raw DFA gradient and $U$ for its conditioned form,
we rescale each nonzero $U$ once, after all active factors, by $\widetilde G=\|G\|_F U/\|U\|_F$. Matching is to the raw DFA gradient
computed at that model's current state, not to a different model's BP gradient.
It controls the instantaneous pre-optimizer hidden-matrix magnitude,
not the AdamW parameter-step norm or the full training trajectory.

The original dense implementation symmetrizes moment blocks, solves in
float32, imposes a minimum ridge, and can increase the ridge or use least
squares after a numerical failure. The larger-network implementation uses
exact feature- or minibatch-space solves at the declared ridge, without
adaptive ridge escalation or pseudoinverse fallback. Numerical failures
remain failed cases rather than silently changing the algorithm.
The factor-transfer confirmation reports no such failures. Training uses
float32, with higher-precision reductions where specified in the study protocols.

\subsection{Exact solves in minibatch space}
\label{app:sample_space}

Conditioning wide layers would be costly if every update required solving
a system with one coordinate per unit. The following identities allow the
same update to be computed from a smaller system with one coordinate per
example when the minibatch is smaller than the layer width.

Let the $b$ rows of $A\in\R^{b\times d_{\rm in}}$ and
$D\in\R^{b\times d_{\rm out}}$ contain the minibatch activities and local
errors. Then $C_A=A^\top A/b$, $C_E=D^\top D/b$, and $G=D^\top A/b$.
Define the minibatch Gram matrices
$K_A=AA^\top/b$ and $K_E=DD^\top/b$. The push-through identity gives
\begin{align}
G(C_A+\lambda_A I)^{-1}
 &=b^{-1}D^\top(K_A+\lambda_A I)^{-1}A,\\
(C_E+\lambda_E I)^{-1}G
 &=b^{-1}D^\top(K_E+\lambda_E I)^{-1}A,\\
(C_E+\lambda_E I)^{-1}G(C_A+\lambda_A I)^{-1}
 &=b^{-1}D^\top(K_E+\lambda_E I)^{-1}
                 (K_A+\lambda_A I)^{-1}A.
\end{align}
The two Gram factors need not commute. These identities preserve the
current-batch moments and the learning rule; they change only the solve
coordinates. The implementation chooses the smaller system separately
for each factor. When both use minibatch space, the cost is
$O(b^3+b^2(d_{\rm in}+d_{\rm out})+b\,d_{\rm in}d_{\rm out})$, with
$O(b^2+b(d_{\rm in}+d_{\rm out}))$ auxiliary storage beyond the parameters
and returned gradient. The dense gradient still requires
$O(d_{\rm in}d_{\rm out})$ storage. These identities apply to current-batch
moments, including sample-dependent gates; moving averages require a
separate representation or approximation.

\subsection{Replication units, selection, and statistical comparisons}
\label{app:statistics}

Repeated training runs measure how sensitive a result is to training
randomness. To interpret that uncertainty, we distinguish independent runs
from repeated feedback draws and from changes in the experimental condition.

For the original MNIST and Fashion-MNIST confirmations, the unit of
replication is a model/data-order seed after averaging three feedback seeds.
Tables report means and standard errors of the mean (SEM). For the additional CIFAR-10 studies,
each replicate jointly varies initialization, sampling, augmentation, and
feedback as applicable; tables report means and sample standard deviations
(SD). Methods are paired within a study and condition. Reusing seed labels
across widths or training pools does not make the conditions independent
replicates, and earlier development seeds are never pooled with confirmation.

Hyperparameters in these additional studies are selected on validation data. Those earlier studies retain their
declared final endpoint. The joint confirmation separately prespecifies
validation-CE checkpoint selection for its work-budget comparisons. When the best tested setting lies at the edge of a search grid, we report
it as a boundary selection rather than assuming that the optimum was found.
The original factor experiments combine separately selected A and E
damping in K. The distinct full-pool boundary study uses a joint K grid
and is identified separately for this reason.

CIFAR-10 paired intervals are individual two-sided 95\% Student $t$
intervals, computed from the seed-level differences. Positive accuracy
differences and negative cross-entropy (CE) differences favor the first method. Intervals
are conditional on the benchmark, reused split, and development procedure;
they do not describe dataset uncertainty or hyperparameter-selection
uncertainty. Exact sign-flip tests additionally assume exchangeability of
paired signs. The earlier four-study evaluation prespecifies two accuracy comparisons:
A--DFA in the optimizer study and A--FD-DFA, where FD-DFA uses
forward decorrelation, in the equal-budget decorrelation study. Holm adjustment uses this fixed family of two. All other comparisons in that earlier evaluation, including factor-transfer
contrasts, are descriptive. The later joint confirmation uses the distinct
21-test family in Appendix~\ref{app:joint_confirmation}.

\subsection{Final evaluation and reproducibility scope}

Final testing assesses the selected models without using their test scores
to revise settings or training duration. The retained prediction records
also allow the reported metrics to be checked without retraining.

The first four shared-BN studies evaluate 280 final models: 96 from the
scale study, 128 from factor transfer, 24 from optimizer confirmation, and
32 from the forward-decorrelation comparison. The follow-ups and
separate boundary study contribute a further 170 and 50 models, respectively
(Table~\ref{tab:study_ledger}). The first four studies reproduce their
saved validation predictions using training-derived preprocessing and
frozen normalization state. No test result selects a model, seed, horizon,
or setting. Scale-study A checkpoints are at 5,000 updates; BP/DFA
checkpoints are at 8,500. Raw-method checkpoints at 5,000 retain metrics
only, so that equal-update comparison is reported on validation.

The official test set was used in earlier project studies. Final model
selection was fixed before this evaluation, but the dataset itself was
not previously unseen. Prediction-based reproduction checks verify
reported accuracy and losses without repeating training or checkpoint
inference. Compact records retain exact predicted classes and labels and
float64 per-example CE; reconstructing CE directly from logits requires
the full archived evidence. Hardware timing and the original order of
test access cannot be established from those compact records alone.

\subsection{Study protocols and cohorts}
\begin{table}[!htbp]
    \centering
    \small
    \setlength{\tabcolsep}{4pt}
    \begin{tabular}{@{}p{0.18\textwidth}p{0.76\textwidth}@{}}
        \toprule
        Tier & Protocol summary \\
        \midrule
        Synthetic stress suite &
        Eight-class latent-circle task with input dimension 64 and nuisance dimension 24. Regimes are nuisance-dominant $(s_{\rm task},s_{\rm nuis})=(0.45,2.0)$, low-sample/noisy $(0.7,1.5)$, mixed-context $(0.75,1.2)$ with task--nuisance interaction, and task-aligned $(1.3,0.25)$. The 128 cells are 4 regimes $\times$ train sizes 512/1024/2048/4096 $\times$ input noise 0.05/0.15 $\times$ train-label noise 0/0.1/0.2/0.4, evaluated on clean test labels with 4096 test examples. MLP hidden widths are 256--128, trained for 14 epochs plus an epoch-0 evaluation, with batch size 128 and learning rate 0.08. The primary comparison fixes full-rank feedback, damping $\lambda=0.3$, 5 data seeds, and 3 feedback seeds; BP uses the same data seeds. Rank-restricted and test-selected summaries are historical sensitivity analyses only. \\
        \addlinespace
        Vision MLP noise &
        Fashion-MNIST and CIFAR-10 MLP sweeps use train sizes 1000/3000/10000 and train-label noise 0/0.1/0.2/0.4, with clean test labels (2000 Fashion-MNIST and 3000 CIFAR-10 test examples in the aggregate). The BP, FA/DFA, DRTP, VNC/NMNC, and nDFA methods are compared; local-feedback methods aggregate over 5 data seeds and 3 feedback seeds, while BP aggregates over data seeds only. The saved sweep fixes full-rank feedback and contains one configuration per method and cell; its common short training recipe is exploratory. \\
        \addlinespace
        Synthetic controls &
        BP regularization controls use matched seeds for BP, BP+L2, BP+label smoothing, early stopping, raw DFA, DFA+BP-norm matching, and conditioned DFA. \\
        \addlinespace
        Clean three-factor MLPs &
        The tanh studies use three hidden width-300 layers for 1,000 updates (batch 128, learning rate $10^{-3}$). A fixed 5,000-example split selects $\lambda_A/\lambda_E$ independently: MNIST uses development seeds 42--44, frozen $0.3/10$, and confirmation seeds 50--54; preregistered Fashion-MNIST uses seeds 60--62, frozen $0.03/30$, and confirmation seeds 70--74. The architectural replication uses a 256--128 ReLU MLP with softmax cross-entropy (learning rate $0.03$), development seeds 0--2, frozen $3/0.1$, and fresh confirmation seeds 100--107. All cross model seeds with three feedback seeds and norm-match conditioned hidden gradients to DFA. Their later runners evaluate test at the last update; the earlier MNIST test exposure is disclosed in Appendix~\ref{app:threefactor}. \\
        \bottomrule
    \end{tabular}
    \caption{\textbf{Protocol summary.} Compact details for the main
    experimental tiers; code availability is stated in the Reproducibility
    Statement.}
    \label{tab:experiment_protocol}
\end{table}

\begin{table}[!htbp]
\centering\small
\setlength{\tabcolsep}{4pt}
\begin{tabular}{@{}p{0.25\textwidth}p{0.31\textwidth}p{0.27\textwidth}l@{}}
\toprule
Study / artifact ID & Development and selection & Final cohort & GPU\\
\midrule
Scale / S1 & Earlier validation choices transferred & 8 seeds, 4 cells, 3 rules; 96 models & H100\\
Error transfer / S2 & 2 seeds, 7 E dampings; mean CE & 8 seeds, 4 cells, 4 rules; 128 models & H100\\
Optimizer / S3 & 2 seeds, 3 rates per rule; mean CE & 8 seeds, 3 rules; 24 models & H100\\
Decorrelation / S4 & 2 seeds, 9 settings per rule; mean CE & 8 seeds, 4 rules; 32 models & H100\\
Longer work / S5 & S4 settings transferred & 5 seeds, 2 budgets, 4 rules; 40 models & H100\\
Width retuning / S6 & 2 seeds, 4 dampings, 2 operators and widths; mean CE & 5 seeds, 2 widths, 5 rules; 50 models & H100\\
BN intervention / S7 & Fixed common recipe & 5 seeds, 4 conditions, 4 rules; 80 models & H100\\
Separate protocol / S8 & 24 trials per rule; final accuracy, CE tie-break & 5 seeds, 10 rules; 50 models & H200\\
\bottomrule
\end{tabular}
\caption{CIFAR-10 study ledger. Each row is a separate cohort. S1--S4
account for the original 280 final models; S5--S7 add 170, and S8 adds 50.
The artifact IDs identify distinct study cohorts; exact configurations and
source hashes are archived with the reproduction records. The public entry
point identifies the released summaries and the complete joint-confirmation
records; historical checkpoints and per-example predictions are not distributed. Selection uses validation data;
confirmation reports the declared final test endpoint. CE denotes
cross-entropy. H100 measurements use the 80GB HBM3 model.}
\label{tab:study_ledger}
\end{table}

\subsection{Computational scope}
\label{app:compute}
The direct implementation adds a damped moment solve per active factor and
layer. For width $d$ and batch size $b$, forming and solving a dense factor
costs $O(d^2b+d^3)$ time and $O(d^2)$ storage. The exact sample-space identities
above reduce this cost when the batch is narrower than the layer, using the
same low-rank inversion principle as SKFAC \citep{tang2021skfac}. K-nDFA
charges both solves. The measured comparisons in
Appendices~\ref{app:scale}--\ref{app:revision_followups} include these operations
and report their timing scopes; equal update counts alone are not equal work.

\FloatBarrier
\section{Synthetic evidence and mechanism controls}
\label{app:synthetic}
The synthetic suite tests whether conditioning helps when high-variance
directions contain nuisance. Table~\ref{tab:infodfa_synthetic_noise_split}
gives the full-rank endpoints; the paired analysis below treats each global
seed, rather than each designed cell, as a replicate.
\begin{table}[!htbp]
    \centering
    \small
    \setlength{\tabcolsep}{4pt}
    \begin{tabular}{@{}lrrrrrr@{}}
            \toprule
            Regime & BP & DFA & nDFA & Best NC & $\Delta$ nDFA & Wins \\
            \midrule
            Nuisance-dominant & 27.9 & 13.4 & 53.3 & 26.8 & +39.9 & 32/32 \\
            Low-sample/noisy & 51.4 & 34.8 & 65.7 & 50.0 & +30.9 & 32/32 \\
            Mixed-context & 38.0 & 25.3 & 47.3 & 36.2 & +22.0 & 32/32 \\
            Task-aligned & 92.0 & 74.5 & 90.4 & 91.7 & +15.9 & 32/32 \\
            \bottomrule
    \end{tabular}
    \caption{Primary fixed-full-rank synthetic summary at the shared learning
    rate. BP comes from this sweep; the separately test-tuned BP references
    in Fig.~\ref{fig:rule_positive} and Table~\ref{tab:infodfa_bpwhiten} use a different study.
    ``Best NC'' is the best
    noise-correlation baseline in the archived sweep. Cells are designed
    conditions rather than independent population samples.}
    \label{tab:infodfa_synthetic_noise_split}
\end{table}

\begin{table}[!htbp]
\centering\small
\setlength{\tabcolsep}{4pt}
\begin{tabular}{@{}lrrrr@{}}
\toprule
Regime & BP loss & DFA loss & A loss & DFA loss $>10$ (\%) \\
\midrule
Nuisance-dominant & 0.77 & 86,170.82 & 0.41 & 100.0 \\
Low-sample/noisy & 0.68 & 1,350.76 & 0.38 & 97.1 \\
Mixed-context & 0.93 & 1,195.00 & 0.52 & 100.0 \\
Task-aligned & 0.79 & 1.66 & 0.60 & 25.0 \\
\bottomrule
\end{tabular}
\caption{\textbf{Predictive instability in the shared-rate synthetic sweep.} Median final training loss and the fraction of raw-DFA runs with loss above 10. BP uses 160 runs per regime; local rules use 480 crossed data/feedback runs. These distribution summaries do not treat crossed runs as independent replications. No activity-nDFA run exceeds the threshold. Large accuracy differences in this sweep therefore combine stabilization with directional effects; norm-matched controls address the latter separately.}
\label{tab:infodfa_synthetic_stability}
\end{table}

\label{app:stat_tests}

The suite crosses designed conditions with five global data seeds and
repeated feedback draws. We average the full designed grid within each
global seed before estimating uncertainty. Crossed feedback draws are not
additional independent replicates. Table~\ref{tab:infodfa_seedlevel_stats}
reports this conditional seed-level comparison.

\begin{table}[!htbp]
    \centering
    \small
    \setlength{\tabcolsep}{4pt}
    \begin{tabular}{@{}llrrr@{}}
        \toprule
        Regime & Comparison & Mean $\Delta$ (pp) & 95\% seed $t$ interval & Wilcoxon $p$ (seeds) \\
        \midrule
        Nuisance-dominant & nDFA $-$ DFA & $+39.86$ & $[+37.42,\,+42.29]$ & $0.0625$ \\
        Low-sample/noisy & nDFA $-$ DFA & $+30.88$ & $[+27.85,\,+33.90]$ & $0.0625$ \\
        Mixed-context & nDFA $-$ DFA & $+22.02$ & $[+21.45,\,+22.59]$ & $0.0625$ \\
        Task-aligned & nDFA $-$ DFA & $+15.85$ & $[+13.21,\,+18.50]$ & $0.0625$ \\
        \bottomrule
    \end{tabular}
    \caption{Seed-level comparison over the fixed synthetic stress grid.
    Differences are averaged across all 32 designed cells and retained feedback
    draws within each of five global data seeds. Individual Student $t$
    intervals use four degrees of freedom and assume approximately normal,
    independent seed-level differences. Cells and crossed feedback draws
    are not independent replicates; five seeds limit the assessment of coverage. The exact
    two-sided signed-rank test cannot fall below $0.0625$ with five nonzero pairs;
    all five observed differences have the same sign in every regime.}
    \label{tab:infodfa_seedlevel_stats}
\end{table}

\begin{samepage}
All five seed-level differences favor activity nDFA in every regime.
The small number of seeds limits inference: a two-sided signed-rank test
cannot reach the conventional threshold with five nonzero pairs.
The separately tuned BP stars in Fig.~\ref{fig:rule_positive} are descriptive references, not part of
this paired comparison.
\par
\end{samepage}

\subsection{Predicted anisotropy gain}
\label{app:anisotropy_predictor}

The linearized prediction depends on whether high-variance directions carry
nuisance or task signal. Hidden spectra remain anisotropic even in the
task-aligned control, so the trained condition number alone does not identify
the favorable regime. The designed nuisance/task energy ratio instead
describes how the synthetic generator distributes variance before training.
Across 128 cells it correlates with the activity-nDFA gain over DFA
(Spearman $\rho=0.62$; Fig.~\ref{fig:controls_composite}B).
This calculation and Table~\ref{tab:prospective_diagnostic} use the same
exploratory, test-selected feedback-rank summaries; the primary fixed-full-rank
comparison is reported separately above. Because regimes determine much of
the ratio, the association is a descriptive mechanism check, not an
independent-task significance test or a validated predictor for a new task.

\subsection{A prospective estimate of the mechanism variable does not transfer}
\label{app:prospective_diagnostic}

The \texttt{nuisance\_energy\_ratio} above is computed from the synthetic
generator's ground truth, so it cannot be evaluated before training on a new
problem. We tested a purely \emph{prospective} version, estimable from the
training data and a randomly initialized network alone: for each presynaptic
layer that nDFA preconditions, we estimate the activity covariance $C$ from an
untrained forward pass and let $U$ have orthonormal columns spanning
the centered class-conditional mean activations. We define
$\widehat r=(\tr C-\tr U^{\!\top}CU)/\tr U^{\!\top}CU$, combining layers and
five initialization seeds by geometric mean. An empty task subspace has
zero projected energy; the implementation floors the denominator at
$10^{-12}$. The task-blind damped condition
number $\kappa(C)$ serves as a baseline
(\texttt{analysis/compute\_prospective\_diagnostic.py}, artifacts in
\texttt{results/infodfa\_prospective\_diagnostic\_v1/}).

The pre-training estimate does not provide a reliable decision rule
(Table~\ref{tab:prospective_diagnostic}). Its synthetic association is weaker
than that of the designed ratio and no better than the task-blind condition
number, and the association reverses on the vision transfer.
These results limit the proposed diagnostic: the tested initialization
statistics do not capture enough of the subsequent learning dynamics to
predict when conditioning will help.

\begin{table}[!htbp]
    \centering
    \small
    \caption{Prospective nuisance-energy diagnostic vs.\ realized nDFA$-$DFA
    gain. Correlations are Spearman $\rho$. The pre-training estimate
    $\widehat r$ does not exceed the task-blind $\kappa(C)$ baseline on the
    synthetic grid, and reverses sign on the held-out vision transfer.}
    \label{tab:prospective_diagnostic}
    \begin{tabular}{lccc}
        \toprule
        Scope & $\widehat r$ (prospective) & $\kappa(C)$ (task-blind) & designed ratio (post hoc) \\
        \midrule
        Synthetic, all 128 cells      & $0.54$  & $0.59$ & $0.62$ \\
        \midrule
        Vision transfer, all 24 cells & $-0.61$ & --     & -- \\
        \bottomrule
    \end{tabular}
\end{table}

\subsection{Conditioning and feedback-alignment dynamics}
\label{app:alignment_dynamics}

\begin{samepage}
The post-alignment model assumes useful credit and does not explain how it
is acquired. Historical synthetic alignment trajectories used a shared
learning rate and did not separate stabilization from acquisition of useful
credit. They therefore do not establish an alignment-rescue mechanism.
Gradient alignment alone is also insufficient for good learning performance
\citep{chu2023alignment}; initialization can change the order in which
components are learned in linear FA \citep{girotti2021convergence}.
The constant-rate and shared-checkpoint tests in
Appendix~\ref{app:joint_confirmation} provide the controlled timing evidence.
\par
\end{samepage}

\subsection{Reduced variability across feedback and order seeds}
\label{app:feedback_variance}

\begin{figure}[t]
    \centering
    \includegraphics[width=\textwidth]{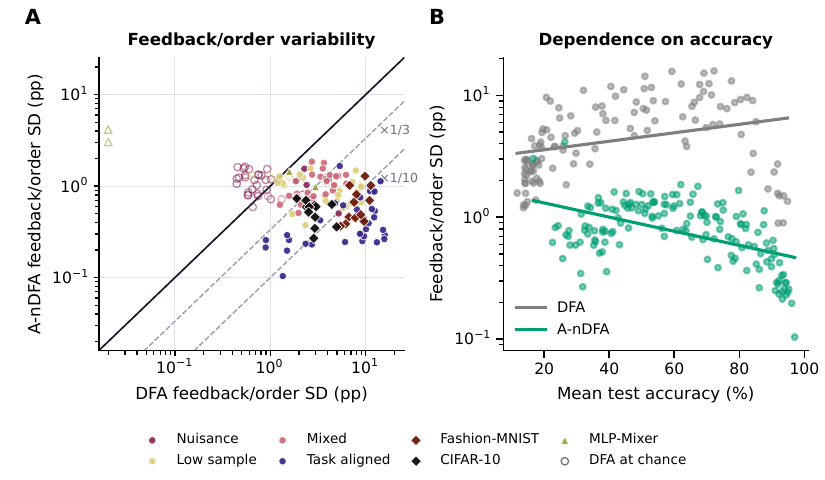}
    \caption{\textbf{Variability across compound feedback/order seeds.} \textbf{A,} Per-cell
    feedback-seed standard deviation of final accuracy, raw DFA versus nDFA
    (log-log; diagonal = no change), colored by regime; vision and MLP-Mixer cells
    are filled diamonds/triangles, and cells where raw DFA sits at the chance
    floor are open circles (excluded from the ratio statistics). Guide lines
    mark $\times\tfrac13$ and $\times\tfrac1{10}$ reductions. \textbf{B,}
    Ceiling control: feedback-seed standard deviation versus mean final
    accuracy for above-chance cells. Fitted method-specific trends show that
    the activity-nDFA reduction persists after accounting descriptively for mean accuracy.
    Only the consistently normalized raw and activity rules are included.}
    \label{fig:feedback_variance}
\end{figure}

To assess sensitivity to random feedback, we calculate the within-data-seed
SD of final accuracy across three compound feedback/order seeds. The
analysis covers 156 synthetic, vision-MLP and MLP-Mixer cells, with 35 chance-floor
DFA cells excluded from ratios because their variance is degenerate.
Among the remaining cells, the median DFA-to-nDFA SD ratio is 6.1.
The reduction appears across regimes and persists descriptively after
accounting for mean accuracy (Fig.~\ref{fig:feedback_variance}).

\begin{samepage}
These are descriptive summaries of a reused grid, not independent-cell
inference. The feedback seed also controls minibatch order, so the result
does not isolate feedback-matrix variability. The four MLP-Mixer cells, two at
floor, provide little separate evidence. Distinguishing feedback and order
effects requires an experiment that varies them independently.
\par
\end{samepage}

\subsection{Regularization, scalar scaling and moment controls}
These controls ask whether stronger BP regularization, a scalar update
rescaling, or coordinatewise normalization explains the activity gain.
They retain their own matched cohorts and selection procedures; their
absolute accuracies should not be pooled across protocols.
\begin{table}[!htbp]
    \centering
    \small
    \setlength{\tabcolsep}{4pt}
    \begin{tabular}{@{}llrrr@{}}
        \toprule
        Task & Method & Test acc & $\Delta$ BP & $\Delta$ DFA \\
        \midrule
        Synthetic nuisance & BP & 58.28 $\pm$ 3.16 & +0.00 & +39.50 \\
         & BP + L2 ($\lambda$=1e-3) & 59.29 $\pm$ 3.09 & +1.01 & +40.51 \\
         & BP + label smoothing (0.1) & 54.90 $\pm$ 3.68 & -3.38 & +36.12 \\
         & BP + early stop & 59.22 $\pm$ 3.53 & +0.94 & +40.44 \\
         & DFA & 18.78 $\pm$ 3.42 & -39.50 & +0.00 \\
         & DFA + norm match & 12.22 $\pm$ 0.41 & -46.06 & -6.56 \\
         & nDFA & 72.45 $\pm$ 0.62 & +14.17 & +53.67 \\
        Fashion-MNIST noisy & BP & 79.19 $\pm$ 0.53 & +0.00 & +17.56 \\
         & BP + L2 ($\lambda$=1e-3) & 79.52 $\pm$ 0.35 & +0.33 & +17.89 \\
         & BP + label smoothing (0.1) & 79.85 $\pm$ 0.28 & +0.66 & +18.22 \\
         & BP + early stop & 79.70 $\pm$ 0.53 & +0.51 & +18.07 \\
         & DFA & 61.63 $\pm$ 1.56 & -17.56 & +0.00 \\
         & DFA + norm match & 77.00 $\pm$ 1.35 & -2.19 & +15.37 \\
         & nDFA & 76.49 $\pm$ 0.58 & -2.70 & +14.86 \\
        \bottomrule
    \end{tabular}
    \caption{\textbf{Regularization and norm-matching controls.} Three BP regularization variants test whether a stronger BP recipe closes the gap to conditioned DFA. \texttt{DFA + norm match} rescales each layer's DFA gradient to match the BP gradient norm on a held-out evaluation batch, isolating the contribution of activity-side second-moment preconditioning (nDFA) from per-layer norm matching. These accuracies come from a single nuisance cell and a single Fashion-MNIST cell run at five matched seeds with a tuned recipe, and are therefore not directly comparable to the sweep-averaged per-regime accuracies of Table~\ref{tab:infodfa_synthetic_noise_split}, which aggregate 32 cells per regime under the shared hyperparameter protocol of Table~\ref{tab:experiment_protocol}.}
    \label{tab:infodfa_controls}
\end{table}
\begin{table}[!htbp]
\centering
\small
\setlength{\tabcolsep}{4pt}
\begin{tabular}{llrr}
\toprule
Cell & Method & Test acc. & $\Delta$ vs.\ DFA+norm \\
\midrule
Nuisance & DFA & 15.3 $\pm$ 0.6 & +0.2 \\
 & DFA + norm & 15.1 $\pm$ 0.3 & +0.0 \\
 & A-nDFA + norm & 30.4 $\pm$ 0.8 & +15.3 \\
\addlinespace[2pt]
Low sample/noisy & DFA & 23.0 $\pm$ 0.8 & +1.4 \\
 & DFA + norm & 21.6 $\pm$ 0.6 & +0.0 \\
 & A-nDFA + norm & 31.9 $\pm$ 0.4 & +10.3 \\
\addlinespace[2pt]
Mixed & DFA & 34.6 $\pm$ 1.0 & +3.3 \\
 & DFA + norm & 31.2 $\pm$ 1.0 & +0.0 \\
 & A-nDFA + norm & 35.4 $\pm$ 1.0 & +4.1 \\
\addlinespace[2pt]
Clean aligned & DFA & 97.2 $\pm$ 0.1 & +0.5 \\
 & DFA + norm & 96.7 $\pm$ 0.1 & +0.0 \\
 & A-nDFA + norm & 96.4 $\pm$ 0.2 & -0.3 \\
\bottomrule
\end{tabular}
\caption{\textbf{Norm-matched factor-control endpoints.} Values are final test accuracy in percent for the focused 100-epoch synthetic cells used in Fig.~\ref{fig:controls_composite}. The norm-matched conditioned rows rescale each hidden local gradient to the matched BP layer norm after applying the conditioner.}
\label{tab:infodfa_normmatch_factor_controls}
\end{table}
\begin{table}[!htbp]
\centering
\small
\setlength{\tabcolsep}{4.5pt}
\begin{tabular}{lrrcrrr}
\toprule
Regime & BP$^{\dagger}$ & BP+precond.$^{\dagger}$ & $\Delta$ & DFA & DFA+$C^{-1/2}$\,$^{\ddagger}$ & nDFA \\
\midrule
Nuisance-dominant & 27.9 & 46.2 & $+18.3$ & 13.5 & 47.5 & 53.3 \\
Low-sample/noisy  & 53.3 & 60.4 & $+7.1$  & 34.7 & 62.0 & 65.7 \\
Mixed-context     & 43.8 & 44.6 & $+0.72$  & 25.4 & 46.7 & 47.5 \\
Task-aligned      & 92.0 & 91.9 & $-0.1$  & 75.0 & 91.0 & 90.4 \\
\bottomrule
\end{tabular}

\vspace{2pt}
{\footnotesize $^{\dagger}$ best of five learning rates per regime, selected by
mean test accuracy. $^{\ddagger}$ DFA, DFA+decorrelation, and the displayed nDFA
column all come from the same matched replication of the 128-cell suite;
decorrelation applies $(C+\lambda I)^{-1/2}$.}
\caption{\textbf{BP-preconditioning control (best learning rate per method).} Applying the same
input-side inverse-second-moment preconditioner used by nDFA to the \emph{exact} BP
gradient. BP and BP+precond.\ are each tuned to their best of five learning rates
($\eta\in\{0.02,0.04,0.08,0.16,0.32\}$) per regime using test accuracy; the
local rules use the fixed sweep rate and receive no analogous rate search. The gain
is regime-specific: large on nuisance-dominant, small on mixed-context (where tuning BP's
own learning rate recovers most of the apparent gain), and $\approx 0$
on the clean task-aligned control. The DFA, decorrelation, and nDFA columns are
the matched replication backing the main-text decorrelation deltas, not the
separate split-table aggregates. Values are final test accuracy in percent.}
\label{tab:infodfa_bpwhiten}
\end{table}

\begin{table}[!htbp]
\centering\small
\setlength{\tabcolsep}{4pt}
\begin{tabular}{@{}lrrrr@{}}
\toprule
Cell & DFA & Hidden Adam & Diagonal $C_A^{-1/2}$ & Activity nDFA \\
\midrule
Nuisance & $17.28\pm1.19$ & $21.03\pm1.04$ & $21.26\pm1.22$ & $45.53\pm0.93$ \\
Low-sample & $33.35\pm0.94$ & $32.33\pm0.92$ & $33.58\pm0.95$ & $40.66\pm0.79$ \\
Mixed & $37.80\pm1.92$ & $36.92\pm1.29$ & $38.78\pm1.66$ & $38.93\pm1.36$ \\
Task-aligned & $97.09\pm0.17$ & $97.05\pm0.19$ & $97.10\pm0.19$ & $96.98\pm0.14$ \\
\bottomrule
\end{tabular}
\caption{\textbf{Exploratory optimizer and diagonal controls.} Test accuracy (\%), mean $\pm$ SEM over five model seeds after averaging two feedback draws per seed. These are single cells with configurations selected on test accuracy, not a validation-selected optimizer benchmark. The diagonal rule uses inverse square roots while A-nDFA uses an inverse; this comparison does not isolate off-diagonal structure at a common power. Adam adapts hidden-weight updates. The original settings and explicit cohort summaries are archived with the reproduction records.}
\label{tab:infodfa_adam_control}
\end{table}

Normalization has long been important in learning with asymmetric feedback
\citep{liao2016symmetry}. We therefore compare activity conditioning and raw
DFA with BN present in both rules, rather than attributing the entire
unnormalized gain to the inverse moment.

\begin{table}[!htbp]
\centering\small
\setlength{\tabcolsep}{4pt}
\begin{tabular}{@{}lrrrr@{}}
\toprule
Regime & DFA+BN & A+BN & $\Delta$ (pp) & Individual 95\% CI \\
\midrule
Nuisance-dominant & 47.37 & 50.72 & $+3.34$ & $[+2.58,+4.11]$ \\
Low-sample/noisy & 66.65 & 67.64 & $+0.99$ & $[+0.26,+1.71]$ \\
Mixed-context & 50.72 & 48.24 & $-2.48$ & $[-3.10,-1.86]$ \\
Task-aligned & 87.46 & 86.48 & $-0.97$ & $[-1.77,-0.18]$ \\
\bottomrule
\end{tabular}
\caption{\textbf{Conditioning within a shared BatchNorm recipe.} Regime means of test accuracy (\%). Both rules use BN before the hidden ReLU under the synthetic sweep's shared learning rate and 14-epoch horizon. Contrasts average all 32 designed cells and three feedback draws within each of five global seeds; intervals use Student $t$ with four degrees of freedom. Conditioning adds a benefit in the nuisance and low-sample regimes but reduces accuracy in the other two. This comparison does not establish the ordering after independent learning-rate tuning or convergence.}
\label{tab:infodfa_bn_baseline}
\end{table}

\FloatBarrier
\begin{table}[!htbp]
    \centering
    \small
    \setlength{\tabcolsep}{4pt}
    \begin{tabular}{@{}lrrrrrr@{}}
            \toprule
            Dataset & BP & DFA & nDFA & NMNC & $\Delta$ nDFA & Wins \\
            \midrule
            Fashion-MNIST & 72.5 & 50.2 & 71.5 & 73.2 & +21.3 & 12/12 \\
            CIFAR-10 & 32.2 & 17.2 & 32.1 & 30.3 & +14.9 & 12/12 \\
            \bottomrule
    \end{tabular}
    \caption{Exploratory noisy-label vision MLP sweep. Feedback rank is fixed
    at full rank; NMNC is the named comparator, not a per-cell selection
    across noise-correlation methods. Training uses the common short recipe.}
    \label{tab:infodfa_vision_noise_split}
\end{table}
\subsection{Transfer to a spurious-correlation task}
\label{app:coloredmnist}

Our ColoredMNIST-style task tests transfer under a controlled change in nuisance structure
\citep{arjovsky2019irm}. The training color--label correlation is reversed
at test, and a grayscale probe removes color. Conditioning reduces seed
variability and brings mean accuracy close to BP
(Table~\ref{tab:infodfa_coloredmnist}). The paired nDFA--DFA accuracy gains
do not survive Holm correction at eight seeds. This supports a descriptive
stability result, without establishing superiority to DFA or BP.

\begin{table}[!htbp]
    \centering
    \small
    \setlength{\tabcolsep}{4pt}
    \begin{tabular}{@{}lrrrrr@{}}
        \toprule
        Method & Test (color-reversed) & Grayscale & $\Delta$ DFA (test) & $\Delta$ BP (test) & $\Delta$ DFA (gray) \\
        \midrule
        BP & 82.81 $\pm$ 0.72 & 91.24 $\pm$ 0.22 & +8.57 & -- & +10.88 \\
        BP + L2 & 82.73 $\pm$ 0.80 & 91.23 $\pm$ 0.22 & +8.49 & -0.09 & +10.87 \\
        DFA & 74.24 $\pm$ 4.64 & 80.36 $\pm$ 4.35 & -- & -8.57 & -- \\
        DFA + norm match & 72.29 $\pm$ 0.27 & 85.46 $\pm$ 0.27 & -1.94 & -10.52 & +5.10 \\
        nDFA & 82.50 $\pm$ 0.32 & 91.87 $\pm$ 0.18 & +8.26 & -0.31 & +11.51 \\
        \bottomrule
    \end{tabular}
    \caption{ColoredMNIST-style spurious-correlation task, 8 seeds. One training environment has 85\% label/color correlation, test has 15\%, and labels remain clean. This differs from the two training environments and label noise of \citet{arjovsky2019irm}. The grayscale probe removes color. Values are mean $\pm$ SEM over eight paired training seeds.
    Differences are descriptive: the nDFA--DFA improvements do not survive
    Holm correction over the tested comparison family.}
    \label{tab:infodfa_coloredmnist}
\end{table}

\FloatBarrier
\section{Independent activity and error confirmations}
\label{app:threefactor}

\begin{figure}[t]
    \centering
    \includegraphics[width=\textwidth]{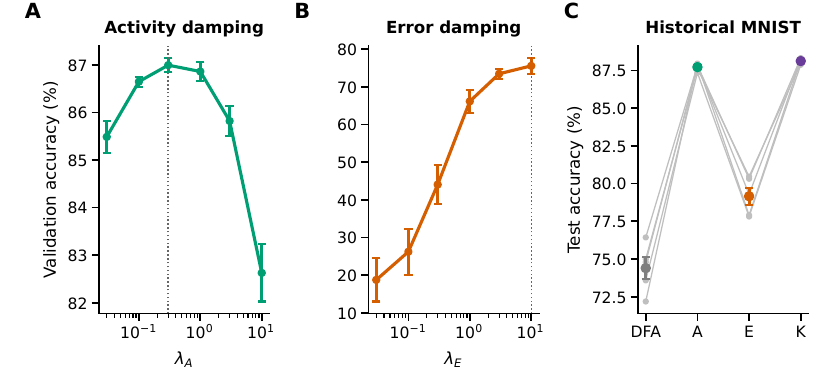}
    \caption{\textbf{Activity, error, and two-sided conditioning.}
    \textbf{A--B,} Development validation accuracy selects activity damping
    $\lambda_A{=}0.3$ and the much heavier error damping $\lambda_E{=}10$
    independently; whiskers are SEM over three development seeds.
    \textbf{C,} Historical MNIST test endpoints. Gray lines connect the four methods
    within each model/data-order seed after averaging three feedback
    seeds; colored points show seed means $\pm$ SEM. Error nDFA improves raw
    DFA without injected input noise, while K-nDFA improves both single-sided
    rules. All conditioned hidden gradients are norm-matched to raw DFA.
    The prior test exposure disclosed in Appendix~\ref{app:threefactor}
    precludes a test-blind confirmation interpretation.}
    \label{fig:threefactor_conditioning}
\end{figure}

To test the two factors separately, we use three width-300 tanh hidden
layers, sigmoid outputs, and one-vs-rest binary log loss. Training uses
1,000 updates, batch size 128, and learning rate $10^{-3}$.
Figure~\ref{fig:threefactor_conditioning}A--B shows independent damping
selection, and panel C shows the original MNIST factor comparison for Eq.~\ref{eq:ndfa}.
Development uses model/data-order seeds 42--44 and one
feedback seed. Activity and error damping are selected independently from
$\{0.03,0.1,0.3,1,3,10\}$ on a fixed 5,000-example split of the MNIST training
set. The selected values are $\lambda_A=0.3$ and $\lambda_E=10$; K-nDFA simply
combines them. The error choice lies at the upper grid boundary and validation
accuracy was still increasing there, so the MNIST error damping is not fully
localized. The original test comparison fixes these choices and uses model/data-order
seeds 50--54 crossed with three feedback seeds. We average feedback seeds
within each model seed before computing the five paired differences. This later
runner evaluates test only at its final step. However, an earlier runner had
already used the MNIST test set for repeated evaluation during training and evaluated activity and
K-nDFA on seeds 50--54 before validation development. Thus these seeds are
not a test-blind confirmation, and subsequent validation selection does
not undo the earlier exposure. The separately preregistered Fashion-MNIST
and ReLU studies provide the confirmatory replications; the later MNIST
seed extension is reported separately as post-hoc evidence.

\begin{table}[!htbp]
    \centering
    \small
    \setlength{\tabcolsep}{5pt}
    \begin{tabular}{@{}lcccc@{}}
        \toprule
        Method & Activity factor & Error factor & Test accuracy (\%) & Test loss \\
        \midrule
        DFA           & -- & -- & $74.40 \pm 0.71$ & $1.844 \pm 0.013$ \\
        A-nDFA & $\checkmark$ & -- & $87.72 \pm 0.11$ & $1.272 \pm 0.007$ \\
        E-nDFA    & -- & $\checkmark$ & $79.16 \pm 0.57$ & $1.620 \pm 0.013$ \\
        K-nDFA        & $\checkmark$ & $\checkmark$ & $\mathbf{88.12 \pm 0.09}$ & $\mathbf{1.206 \pm 0.004}$ \\
        \bottomrule
    \end{tabular}
    \caption{Original MNIST factor comparison, mean $\pm$ SEM over five
    model/data-order seeds after averaging three feedback seeds within each.
    The paired accuracy gains are $+13.32$\,pp for A-nDFA over DFA,
    $+4.76$\,pp for E-nDFA over DFA, and $+0.40$\,pp for K-nDFA over
    A-nDFA. Every contrast agrees in sign across all five seed-level
    pairs; the two-sided Wilcoxon test is floor-limited at $p=0.0625$.
    The prior test exposure is disclosed in Appendix~\ref{app:threefactor}.}
    \label{tab:threefactor_confirmation}
\end{table}

The result supports the algebraic symmetry while also showing an empirical
asymmetry. Activity conditioning has the larger standalone effect and tolerates
moderate damping. Error conditioning has a smaller but independent benefit and
requires substantially heavier damping under this recipe. Fixed feedback and
tanh saturation shape the local errors, but these experiments do not isolate
error anisotropy as the cause of the damping dependence.
K-nDFA is useful when both corrections are active; it should not be interpreted
as automatically superior in regimes where $C_E$ is noisy or nearly singular.

\subsection{Preregistered Fashion-MNIST replication and source-swap diagnosis}
\label{app:fashion_threefactor}

Before reading any development endpoint, we registered the clean Fashion-MNIST
protocol and pass/fail criteria in a timestamped protocol. The architecture, optimizer, loss, step budget, norm
matching, and train/validation/test discipline match the MNIST study, but the
split and model/data-order seeds are fresh. Development seeds 60--62 select
$\lambda_A$ and $\lambda_E$ independently
(Table~\ref{tab:fashion_threefactor}) from
$\{0.03,0.1,0.3,1,3,10,30,100\}$; confirmation freezes the selected
$0.03/30$ pair on seeds 70--74 crossed with three feedback seeds. The activity
optimum lies at the lower grid boundary, so its absolute damping is not fully
localized; the error optimum is interior.

\begin{table}[!htbp]
    \centering
    \small
    \setlength{\tabcolsep}{5pt}
    \begin{tabular}{@{}lcccc@{}}
        \toprule
        Method & Activity factor & Error source & Test accuracy (\%) & Test loss \\
        \midrule
        DFA                    & -- & --  & $61.98 \pm 0.17$ & $1.938 \pm 0.012$ \\
        A-nDFA          & $\checkmark$ & --  & $80.29 \pm 0.10$ & $1.575 \pm 0.007$ \\
        E-nDFA             & -- & DFA & $63.74 \pm 0.41$ & $1.842 \pm 0.010$ \\
        K-nDFA                 & $\checkmark$ & DFA & $\mathbf{80.79 \pm 0.09}$ & $\mathbf{1.500 \pm 0.007}$ \\
        K-nDFA (BP source)     & $\checkmark$ & BP  & $80.29 \pm 0.10$ & $1.576 \pm 0.007$ \\
        \bottomrule
    \end{tabular}
    \caption{Frozen Fashion-MNIST confirmation, mean $\pm$ SEM over five
    model/data-order seeds after averaging feedback seeds. The preregistered
    contrasts are $+1.77$\,pp for E-nDFA over DFA, $+0.50$\,pp for K-nDFA
    over A-nDFA, and $+0.499$\,pp for local K-nDFA over the BP-source
    comparator; each has the same sign in all five pairs.}
    \label{tab:fashion_threefactor}
\end{table}

The frozen confirmation supports the error-only and incremental two-sided
gains (Fig.~\ref{fig:error_kndfa_replication}A--B). The BP-error comparator does not
isolate an error-source effect: its moment is much smaller than the local
DFA-error moment, so the shared damping makes it nearly scalar. Norm matching
then makes its update almost identical to activity nDFA. An audited trajectory
confirms this scale mismatch. The source-swap criterion therefore provides
no evidence for source equivalence or the absence of a locality penalty.

A post-hoc validation sweep of BP-source damping over
$\{0.001,0.003,0.01,0.03,0.1,0.3,1,3,10,30\}$ selects 3, with loss breaking
an accuracy tie. Five fresh model/data-order seeds (80--84), each averaged
over three feedback draws, again favor local K-nDFA over activity nDFA.
The retuned BP-source factor adds little benefit and remains almost parallel
to the activity-only update on the audited trajectory. This comparison is
exploratory and specific to the tested source scales; it does not establish
source equivalence. The ReLU/softmax study below tests a separate question:
whether the two factor gains persist under a different activation and loss.

\subsection{Fresh-seed ReLU/softmax architectural replication}
\label{app:relu_threefactor}

The next confirmation changes the tanh DFA-stall architecture and one-vs-rest
binary log loss simultaneously. A 256--128 ReLU MLP uses multiclass softmax cross-entropy,
learning rate $0.03$, batch size 128, and 1,000 updates. Development seeds 0--2
select $\lambda_A$ from $\{0.03,0.1,0.3,1,3,10,30\}$ and $\lambda_E$ from
$\{0.003,0.01,0.03,0.1,0.3,1,3,10\}$ on a fixed 5,000-example split. The
selected $3/0.1$ pair is interior to both grids.
Table~\ref{tab:relu_threefactor} reports the test endpoints. We recorded the protocol and
pass/fail criteria before test evaluation. Confirmation uses fresh
model/data-order seeds 100--107 crossed with feedback seeds 0--2, averages
feedback seeds within model seed, and evaluates test only after the last update.
As in the tanh studies, conditioned hidden weight gradients are norm-matched to
raw DFA.

\begin{table}[!htbp]
    \centering
    \small
    \setlength{\tabcolsep}{5pt}
    \begin{tabular}{@{}lcccc@{}}
        \toprule
        Method & Activity factor & Error factor & Test accuracy (\%) & Test loss \\
        \midrule
        DFA           & -- & -- & $87.08 \pm 0.79$ & $187.339 \pm 95.506$ \\
        A-nDFA & $\checkmark$ & -- & $95.81 \pm 0.03$ & $0.1385 \pm 0.0005$ \\
        E-nDFA    & -- & $\checkmark$ & $94.61 \pm 0.16$ & $1.242 \pm 0.187$ \\
        K-nDFA        & $\checkmark$ & $\checkmark$ & $\mathbf{96.71 \pm 0.04}$ & $\mathbf{0.1103 \pm 0.0009}$ \\
        \bottomrule
    \end{tabular}
    \caption{Frozen ReLU/softmax confirmation, mean $\pm$ SEM over eight
    model/data-order seeds after averaging three feedback seeds. Error nDFA
    improves DFA by $+7.53\pm0.86$\,pp; K-nDFA improves A-nDFA by
    $+0.90\pm0.04$\,pp and lowers loss by $0.0282\pm0.0011$. Both accuracy
    contrasts have the predicted sign in 8/8 pairs (two-sided Wilcoxon
    $p=0.0078125$).}
    \label{tab:relu_threefactor}
\end{table}

Raw DFA's unusually large cross-entropy reflects rare confident errors from
unstable logits rather than a mismatch in the reported accuracy; all methods
use the same softmax objective and evaluation. Error-only conditioning sharply
reduces that instability, while the two-sided rule gives the best accuracy and
loss. This confirmation supports architecture/loss transfer on MNIST. It does
not erase a negative scope result: in exploratory ReLU/softmax Fashion-MNIST
development, independently selected factors did not preserve an incremental
K-nDFA gain. We therefore do not claim ReLU dataset generality.

\subsection{Post-hoc seed extensions}
The frozen protocols were later repeated with additional model seeds.
Both contrasts retain their sign in the added cohorts
(Table~\ref{tab:factor_seed_extensions}). These extensions improve the
precision of the descriptive comparison but do not address the short training
horizon, establish tuned-baseline superiority, or remove earlier test exposure.
The original cohorts remain the basis of the original tables and figures.

\begin{table}[!htbp]
\centering\small
\setlength{\tabcolsep}{4pt}
\begin{tabular}{@{}lllrrrr@{}}
\toprule
Setting & Cohort & Seeds & $n$ & E$-$DFA & K$-$A & Nominal $p$ \\
\midrule
MNIST/tanh & Extension & 55--59 & 5 & +4.21 & +0.38 & 0.06250 \\
MNIST/tanh & Pooled & 50--59 & 10 & +4.48 & +0.39 & 0.00195 \\
Fashion/tanh & Extension & 75--79 & 5 & +1.54 & +0.46 & 0.06250 \\
Fashion/tanh & Pooled & 70--79 & 10 & +1.65 & +0.48 & 0.00195 \\
MNIST/ReLU & Extension & 108--112 & 5 & +9.05 & +0.91 & 0.06250 \\
MNIST/ReLU & Pooled & 100--112 & 13 & +8.11 & +0.91 & 0.00024 \\
\bottomrule
\end{tabular}
\caption{\textbf{Post-hoc extensions of the frozen factor protocols.} Mean paired accuracy differences in percentage points. Each model/data-order seed averages three feedback draws. Every seed favors E over DFA and K over A; the last column is the identical exact two-sided Wilcoxon value for the two contrasts, without multiplicity adjustment. Pooled results include the original cohort and the post-hoc extension; they are descriptive and do not turn the original MNIST cohort into a test-blind confirmation. Explicit original, extension and pooled exports prevent the larger cohorts from silently replacing the original tables and figures.}
\label{tab:factor_seed_extensions}
\end{table}

\FloatBarrier
\section{CIFAR-10: activity geometry and transfer across scale}
\label{app:scale}

\begin{figure}[t]
\centering
\includegraphics[width=\textwidth]{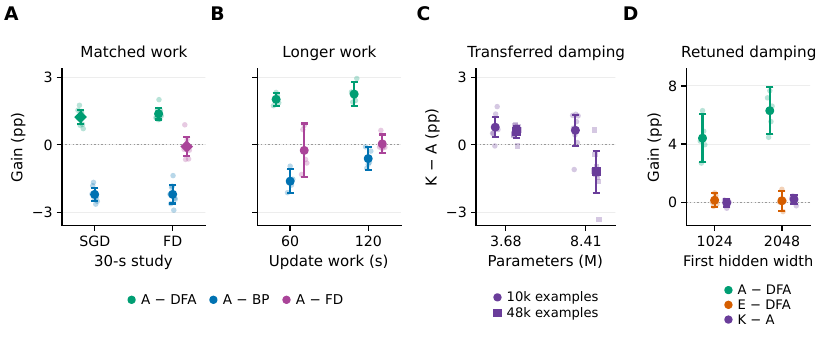}
\caption{\textbf{Earlier CIFAR-10 work and width studies.}
\textbf{A,} A versus DFA, BP, and forward-decorrelated DFA (FD), in separate
optimizer (SGD) and matched-FD studies at 30 seconds of measured update work.
\textbf{B,} Longer matched-work budgets retain the activity gain over DFA;
BP remains stronger and FD is competitive.
\textbf{C,} K versus A at 5,000 updates with transferred damping, across
parameter counts and training-pool sizes (circles: 10k; squares: 48k).
\textbf{D,} Width-specific validation selection retains the activity gain
but provides little additional error-factor benefit.
Faint points show paired seed differences; large markers and whiskers show
means and individual 95\% $t$ intervals (eight seeds in A and C; five in B
and D). Diamonds mark the two prespecified primary comparisons in A;
the remaining contrasts are descriptive. Cohorts are analyzed separately;
different budgets are separate training runs, not checkpoints along a shared trajectory.}
\label{fig:historical_work_width}
\end{figure}

We next ask whether the activity-side gain survives forward normalization
and extends to more training data and wider networks. The studies share
the forward and data protocol below, then vary training-pool size, network
width, and computational budget. Here A, E, and K abbreviate activity-only,
error-only, and two-sided nDFA; CE denotes cross-entropy loss.

\subsection{Shared forward and data protocol}

Holding normalization and data processing fixed helps separate the effect
of conditioning from differences in the forward network.

The shared-BN studies reserve 2,000 of the 50,000 training images for
validation (split seed 80423). Training uses either the remaining 48,000
identities or a nested 10,000-image pool (data seed 9173). Channel means
and population standard deviations are fitted separately on each training
pool. Each update samples 64 identities with replacement and creates two
independently cropped and horizontally flipped views of each. Inputs are
flattened for a ReLU MLP with hidden widths 1024--512 or 2048--1024.
Including BN affine parameters, these models have 3,679,754 and 8,408,074
supervised parameters. BN precedes each hidden ReLU; running statistics
are frozen at validation and test. Feedback scale is 0.1.

The initial activity study selects plain-SGD learning rates of 0.3 for BP
and 0.1 for DFA/A, and relative activity damping $\rho_A=30$.
These settings are transferred unchanged in the scale and factor studies.
The initial search has unequal method-specific trial counts and selected
boundary settings; it is not used to claim equally tuned performance.
The subsequent matched-work studies address search and optimizer fairness
with separate fresh-seed comparisons.

\subsection{Larger training pools, wider models, and longer baselines}

To test transfer without further tuning, we carry the selected settings
across training-pool sizes and widths. Longer unconditioned runs assess
how much of the equal-update advantage can be recovered with extra training.

The scale confirmation uses seeds 33000--33007 in each of the four
training-pool/width cells. A trains for 5,000 updates, while BP and DFA
continue to 8,500; their 5,000-update metrics come from those same
trajectories. The earlier development runs are not pooled into confirmation.
The principal equal-update contrast is A--DFA at 5,000 updates.

On validation, every pair favors A in both accuracy and CE at equal updates. The advantage
persists across the tested pool sizes and widths, while longer raw-DFA
training reduces it. In both larger-pool cells the longer-DFA accuracy
interval spans zero, and longer BP outperforms A. Increasing width enlarges
the A--DFA gap by degrading DFA more than A; it does not improve A's
absolute accuracy. Tables~\ref{tab:scale_cost}--\ref{tab:scale_test_contrasts}
report measured costs and final test comparisons. The complete validation
trajectories are archived with the reproduction records.

\begin{samepage}
Individual paired intervals are descriptive. The four equal-update
accuracy comparisons additionally have exact sign-flip $p=0.0078125$
and Holm-adjusted $p=0.03125$ over that family. These tests concern
training-seed variation within the fixed conditions, not a pooled set
of independent tasks. Measured update times show that the longer raw
baselines still use somewhat less work than A, so the extension is a
nominal allocation rather than an exact equal-time comparison.
\par
\end{samepage}

\begin{table}[!htbp]
\centering\small\setlength{\tabcolsep}{3pt}
\begin{tabular}{@{}llrrr@{}}
\toprule
Cell & Raw rule & $t(A_5)/t(raw_5)$ & $t(raw_{8.5})/t(A_5)$ & 95\% CI (last ratio) \\
\midrule
10k / 3.68M & DFA & $1.805\,\pm\,0.067$ & $0.942\,\pm\,0.033$ & $[0.914,\ 0.969]$ \\
 & BP & $1.795\,\pm\,0.044$ & $0.945\,\pm\,0.026$ & $[0.924,\ 0.966]$ \\
\midrule
48k / 3.68M & DFA & $1.791\,\pm\,0.010$ & $0.948\,\pm\,0.005$ & $[0.943,\ 0.952]$ \\
 & BP & $1.779\,\pm\,0.011$ & $0.954\,\pm\,0.006$ & $[0.949,\ 0.959]$ \\
\midrule
10k / 8.41M & DFA & $1.814\,\pm\,0.012$ & $0.936\,\pm\,0.006$ & $[0.931,\ 0.941]$ \\
 & BP & $1.816\,\pm\,0.012$ & $0.935\,\pm\,0.006$ & $[0.930,\ 0.940]$ \\
\midrule
48k / 8.41M & DFA & $1.810\,\pm\,0.012$ & $0.938\,\pm\,0.006$ & $[0.933,\ 0.943]$ \\
 & BP & $1.804\,\pm\,0.017$ & $0.941\,\pm\,0.009$ & $[0.934,\ 0.949]$ \\
\bottomrule
\end{tabular} \space
\caption{Scale-study measured update-time ratios. Ratios are formed within each seed before averaging; $\pm$ denotes sample SD across seeds. The final interval describes the raw 8,500-update time divided by A at 5,000; values below one mean that the longer raw baseline still uses less measured work.}
\label{tab:scale_cost}
\end{table}
\begin{table}[!htbp]
\centering\small\setlength{\tabcolsep}{3pt}
\begin{tabular}{@{}rrlrr@{}}
\toprule
Train & Params (M) & Rule & Test accuracy (\%) & Test CE \\
\midrule
10k & 3.68 & BP & $53.96\pm0.50$ & $1.4148\pm0.0188$ \\
10k & 3.68 & DFA & $52.13\pm0.90$ & $1.4195\pm0.0206$ \\
10k & 3.68 & A & $54.56\pm0.53$ & $1.3466\pm0.0171$ \\
10k & 8.41 & BP & $54.34\pm0.61$ & $1.4316\pm0.0185$ \\
10k & 8.41 & DFA & $49.57\pm1.72$ & $1.5925\pm0.0726$ \\
10k & 8.41 & A & $54.13\pm0.98$ & $1.4467\pm0.0494$ \\
48k & 3.68 & BP & $58.02\pm0.48$ & $1.1825\pm0.0097$ \\
48k & 3.68 & DFA & $54.03\pm0.83$ & $1.2899\pm0.0202$ \\
48k & 3.68 & A & $55.45\pm1.13$ & $1.2539\pm0.0296$ \\
48k & 8.41 & BP & $58.55\pm0.31$ & $1.1664\pm0.0088$ \\
48k & 8.41 & DFA & $51.76\pm1.42$ & $1.4044\pm0.0608$ \\
48k & 8.41 & A & $53.37\pm2.47$ & $1.3658\pm0.1146$ \\
\bottomrule
\end{tabular} \space
\caption{Scale-study final test endpoints. Mean $\pm$ sample SD across eight seeds per cell. A ends at 5,000 updates; BP and DFA end at 8,500.}
\label{tab:scale_test}
\end{table}
\begin{table}[!htbp]
\centering\small\setlength{\tabcolsep}{3pt}
\begin{tabular}{@{}llrrrr@{}}
\toprule
Cell & Contrast & Accuracy (pp) & 95\% CI & CE & 95\% CI \\
\midrule
10k/3.68 & A$-$BP & +0.596 & $[-0.082,+1.275]$ & -0.0682 & $[-0.0935,-0.0429]$ \\
10k/3.68 & A$-$DFA & +2.430 & $[+1.360,+3.500]$ & -0.0729 & $[-0.0979,-0.0479]$ \\
10k/8.41 & A$-$BP & -0.211 & $[-1.219,+0.796]$ & +0.0151 & $[-0.0326,+0.0628]$ \\
10k/8.41 & A$-$DFA & +4.563 & $[+2.411,+6.714]$ & -0.1458 & $[-0.2365,-0.0551]$ \\
48k/3.68 & A$-$BP & -2.578 & $[-3.539,-1.616]$ & +0.0715 & $[+0.0455,+0.0975]$ \\
48k/3.68 & A$-$DFA & +1.414 & $[+0.388,+2.439]$ & -0.0360 & $[-0.0638,-0.0081]$ \\
48k/8.41 & A$-$BP & -5.181 & $[-7.270,-3.092]$ & +0.1994 & $[+0.1030,+0.2959]$ \\
48k/8.41 & A$-$DFA & +1.606 & $[-0.425,+3.638]$ & -0.0386 & $[-0.1228,+0.0456]$ \\
\bottomrule
\end{tabular} \space
\caption{Activity comparisons at the final test endpoints. Accuracy is in percentage points; CE is unscaled. Positive accuracy and negative CE favor the first method. Individual 95\% intervals are descriptive and unadjusted.}
\label{tab:scale_test_contrasts}
\end{table}
\FloatBarrier
\subsection{Full versus diagonal activity moments}
\label{app:diagonal_control}

A direct control tests whether off-diagonal activity-moment structure
helps beyond separate coordinate scales. In an existing CIFAR-10 confirmation,
full activity conditioning is compared with
$G[\operatorname{diag}(C_A)+\lambda_A I]^{-1}$. Both match the raw DFA hidden
update norms, share BN and a 3072--1024--512--10 ReLU MLP, and use the same
paired training seeds. Each model trains for 500 updates on 10,000 images.
The settings were selected on validation before the eight-seed confirmation:
both use learning rate 0.1, with relative damping 30 for the full moment and
0.3 for its diagonal. This compares separately selected operators, rather
than imposing identical damping or claiming an exactly balanced search.

Full conditioning reaches $46.51\pm1.04\%$ test accuracy and the diagonal
control $42.10\pm1.34\%$ (sample SD). Every paired seed favors the full moment.
The additional full-minus-diagonal contrast is $4.41$ percentage points with
an individual 95\% $t$ interval $[3.71,5.10]$, conditional on the fixed data split
and approximately normal independent seed differences. This contrast was
not one of the original five declared tests and is reported as an additional
analysis. The result supports useful off-diagonal uncentered moment structure
under this recipe. Since $C_A=\operatorname{Cov}(h)+\mu\mu^\top$, with
$\mu=\E[h]$, the diagonal control removes both off-diagonal centered
covariance and off-diagonal mean products. BN before ReLU does not center
the post-ReLU activities. This experiment therefore does not distinguish
centered correlations from mean-activity effects, or establish that
coordinatewise methods generally cannot reproduce the benefit.

\FloatBarrier
\section{CIFAR-10: transfer of the error factor}
\label{app:factors}

The activity-side gain leaves open whether error conditioning adds a
benefit at the same scales. We test this by fixing the activity damping,
selecting error damping using E alone, and evaluating their combination.

A separate study compares DFA, A, E, and K in every training-pool/width
cell, using seeds 35000--35007 and the shared plain-SGD protocol for
5,000 updates. Activity damping remains fixed at $\rho_A=30$.
E-only development on the smaller-model, smaller-pool cell uses seeds
34000 and 34001 to select a single error damping by final mean validation
CE. The selected $\rho_E=30$ is transferred unchanged to E and K in every
cell. The grid is $\{0.03,0.1,0.3,1,3,10,30\}$, and K outcomes
do not enter selection. The selected upper boundary leaves the optimum
beyond the grid unresolved; development scores are archived with the reproduction records.

The activity-side benefit repeats in every pair. K lowers validation CE
relative to A in all pairs of both narrower-model cells, but all final
validation-accuracy intervals span zero. Final testing favors K in the
narrower cells and favors A in the widest, larger-pool cell. In the latter,
the validation trajectory changes from an early K advantage to an endpoint
disadvantage. E alone is not a reliable improvement on DFA. These results
support conditional complementarity (Fig.~\ref{fig:historical_work_width}C),
with all factor-test contrasts treated
as descriptive. Both factors are charged in the measured cost: K requires
more update time than A at the same number of updates.

\begin{table}[!htbp]
\centering\small\setlength{\tabcolsep}{3pt}
\begin{tabular}{@{}lrrrr@{}}
\toprule
Cell & $t(A)/t(DFA)$ & $t(E)/t(DFA)$ & $t(K)/t(A)$ & 95\% CI (last ratio) \\
\midrule
10k / 3.68M & $1.639\,\pm\,0.033$ & $1.638\,\pm\,0.025$ & $1.130\,\pm\,0.017$ & $[1.116,\ 1.145]$ \\
48k / 3.68M & $1.645\,\pm\,0.025$ & $1.648\,\pm\,0.022$ & $1.140\,\pm\,0.028$ & $[1.117,\ 1.163]$ \\
10k / 8.41M & $1.675\,\pm\,0.024$ & $1.681\,\pm\,0.022$ & $1.132\,\pm\,0.019$ & $[1.116,\ 1.147]$ \\
48k / 8.41M & $1.681\,\pm\,0.010$ & $1.688\,\pm\,0.017$ & $1.130\,\pm\,0.005$ & $[1.125,\ 1.134]$ \\
\bottomrule
\end{tabular} \space
\caption{Factor-transfer measured update-time ratios. Within-seed ratios are reported as mean $\pm$ sample SD over eight seeds; the final column gives the individual interval for K/A. These equal-update costs use the factor runner's timing scope.}
\label{tab:factor_cost}
\end{table}
\begin{table}[!htbp]
\centering\small\setlength{\tabcolsep}{3pt}
\begin{tabular}{@{}rrlrr@{}}
\toprule
Train & Params (M) & Rule & Test accuracy (\%) & Test CE \\
\midrule
10k & 3.68 & DFA & $50.86\pm0.62$ & $1.4175\pm0.0152$ \\
10k & 3.68 & A & $55.11\pm0.56$ & $1.3353\pm0.0138$ \\
10k & 3.68 & E & $51.01\pm0.43$ & $1.4212\pm0.0117$ \\
10k & 3.68 & K & $55.89\pm0.33$ & $1.3148\pm0.0134$ \\
10k & 8.41 & DFA & $48.83\pm1.21$ & $1.5669\pm0.0589$ \\
10k & 8.41 & A & $54.46\pm0.63$ & $1.4280\pm0.0321$ \\
10k & 8.41 & E & $47.51\pm1.33$ & $1.6748\pm0.0945$ \\
10k & 8.41 & K & $55.10\pm0.85$ & $1.4192\pm0.0532$ \\
48k & 3.68 & DFA & $50.96\pm0.67$ & $1.3770\pm0.0242$ \\
48k & 3.68 & A & $55.70\pm0.75$ & $1.2514\pm0.0220$ \\
48k & 3.68 & E & $49.84\pm1.00$ & $1.4169\pm0.0326$ \\
48k & 3.68 & K & $56.28\pm0.80$ & $1.2385\pm0.0224$ \\
48k & 8.41 & DFA & $48.00\pm1.02$ & $1.5528\pm0.0669$ \\
48k & 8.41 & A & $53.93\pm1.27$ & $1.3522\pm0.0547$ \\
48k & 8.41 & E & $45.50\pm2.24$ & $1.7419\pm0.1140$ \\
48k & 8.41 & K & $52.73\pm1.92$ & $1.4415\pm0.0753$ \\
\bottomrule
\end{tabular} \space
\caption{Complete factor-transfer test endpoints at 5,000 updates. Mean $\pm$ sample SD over eight seeds per cell.}
\label{tab:factor_test}
\end{table}
\begin{table}[!htbp]
\centering\small\setlength{\tabcolsep}{3pt}
\begin{tabular}{@{}llrrrr@{}}
\toprule
Cell & Contrast & Accuracy (pp) & 95\% CI & CE & 95\% CI \\
\midrule
10k/3.68 & A$-$DFA & +4.245 & $[+3.890,+4.600]$ & -0.0822 & $[-0.0930,-0.0713]$ \\
10k/3.68 & E$-$DFA & +0.149 & $[-0.484,+0.782]$ & +0.0037 & $[-0.0061,+0.0135]$ \\
10k/3.68 & K$-$A & +0.784 & $[+0.332,+1.235]$ & -0.0205 & $[-0.0266,-0.0144]$ \\
10k/8.41 & A$-$DFA & +5.626 & $[+4.898,+6.355]$ & -0.1389 & $[-0.1746,-0.1032]$ \\
10k/8.41 & E$-$DFA & -1.323 & $[-2.101,-0.544]$ & +0.1079 & $[+0.0528,+0.1630]$ \\
10k/8.41 & K$-$A & +0.646 & $[-0.054,+1.346]$ & -0.0088 & $[-0.0469,+0.0292]$ \\
48k/3.68 & A$-$DFA & +4.736 & $[+4.144,+5.328]$ & -0.1256 & $[-0.1406,-0.1107]$ \\
48k/3.68 & E$-$DFA & -1.120 & $[-1.559,-0.681]$ & +0.0399 & $[+0.0225,+0.0572]$ \\
48k/3.68 & K$-$A & +0.576 & $[+0.316,+0.836]$ & -0.0129 & $[-0.0203,-0.0055]$ \\
48k/8.41 & A$-$DFA & +5.929 & $[+5.197,+6.661]$ & -0.2006 & $[-0.2375,-0.1636]$ \\
48k/8.41 & E$-$DFA & -2.504 & $[-4.020,-0.987]$ & +0.1891 & $[+0.1029,+0.2753]$ \\
48k/8.41 & K$-$A & -1.200 & $[-2.139,-0.261]$ & +0.0893 & $[+0.0566,+0.1219]$ \\
\bottomrule
\end{tabular} \space
\caption{Factor comparisons at the final test endpoints. Positive accuracy and negative CE favor the first method. Individual 95\% intervals are descriptive and unadjusted; K--A measures the incremental error factor, as in Fig.~\ref{fig:historical_work_width}C.}
\label{tab:factor_test_contrasts}
\end{table}

\FloatBarrier
\section{CIFAR-10: comparisons at matched measured work}
\label{app:matched_work}

Because conditioning increases the cost of each update, equal update counts
do not imply equal computational cost. These studies compare methods at
the same measured update time and with equal numbers of tuning trials.

\subsection{Common optimizer and work budget}

A shared optimizer and time-based schedule make the computational comparison
interpretable even when methods complete different numbers of updates.

Both matched-work studies use the larger training pool and narrower MLP,
shared BN, and momentum SGD with momentum 0.9, no Nesterov acceleration,
and weight decay $5\times10^{-4}$ on weight matrices only. Learning rate
warms linearly over the first 5\% of measured work from 0.1 times the
peak, then follows a cosine decay to 0.01 times the peak. The schedule
advances by synchronized update time divided by a 30-second budget.
Sampling, augmentation, gradient construction, conditioning or decorrelation,
and optimization are charged. Validation, loading, export, and full-state
diagnostics are excluded; total process time is reported separately.
The final update's budget overshoot is included. Both original
30-second studies ran on an NVIDIA H100 80GB HBM3, as did the 60- and
120-second follow-ups. Training tensors are float32 and no mixed-precision
autocast is used. The follow-up code explicitly disables TF32 for matmul and
cuDNN; the original 30-second receipts do not record these runtime flags.
Each budget uses a separate cohort, so the columns do not describe
checkpoints from one trajectory or an otherwise identical budget intervention.

\subsection{Equal learning-rate search for BP, DFA, and A}

This first comparison asks whether the activity-side gain remains when raw
DFA receives its own learning-rate search under the same work budget.

Development seeds 36000--36001 evaluate peak rates
$\{0.01,0.03,0.1\}$ for every method. Mean final validation CE selects
the setting, followed by mean accuracy and then the smaller peak rate.
Activity damping is inherited from the earlier study. Selection chooses
BP/DFA/A peak rates 0.1/0.03/0.1; BP and A reach the upper boundary.
Confirmation uses eight fresh paired seeds, 37000--37007.

A improves DFA in validation and test accuracy and CE across all paired
seeds, despite completing fewer updates in the common work budget.
The final-test primary comparison remains significant under the declared
two-comparison Holm adjustment (Fig.~\ref{fig:historical_work_width}A, SGD rows).
BP is stronger than A. This establishes
an activity-side benefit under the specified finite optimizer search and
work budget; boundary selections leave fully tuned convergence unresolved.

\begin{table}[!htbp]
\centering\small\setlength{\tabcolsep}{3pt}
\begin{tabular}{@{}lrrr@{}}
\toprule
Rule & Updates & Measured update seconds & Total child seconds \\
\midrule
BP & $14429.5\,\pm\,88.4$ & $30.001131\,\pm\,0.000724$ & $43.67\,\pm\,3.39$ \\
DFA & $14485.1\,\pm\,75.1$ & $30.000759\,\pm\,0.000425$ & $44.69\,\pm\,3.26$ \\
A & $8032.3\,\pm\,65.5$ & $30.001770\,\pm\,0.001249$ & $43.10\,\pm\,3.45$ \\
\bottomrule
\end{tabular} \space
\caption{Optimizer-study updates and costs at the work endpoint. Mean $\pm$ sample SD over eight seeds. Total process time includes overhead excluded from the update clock.}
\label{tab:optimizer_cost}
\end{table}
\begin{table}[!htbp]
\centering\small\setlength{\tabcolsep}{3pt}
\begin{tabular}{@{}rrlrr@{}}
\toprule
Train & Params (M) & Rule & Test accuracy (\%) & Test CE \\
\midrule
48k & 3.68 & BP & $63.52\pm0.20$ & $1.0328\pm0.0031$ \\
48k & 3.68 & DFA & $60.07\pm0.26$ & $1.1339\pm0.0043$ \\
48k & 3.68 & A & $61.31\pm0.29$ & $1.0979\pm0.0036$ \\
\bottomrule
\end{tabular} \space
\caption{Optimizer-study test endpoints. Mean $\pm$ sample SD across eight paired seeds at the measured-work endpoint.}
\label{tab:optimizer_test}
\end{table}

\subsection{Equal-budget comparison with forward decorrelation}

Forward decorrelation provides a closely related way to address activity
correlations. We compare it with update conditioning while allowing both
methods to tune their method-specific parameter.

Development seeds 38000--38001 evaluate nine settings each for BP, DFA,
A, and FD-DFA; confirmation uses seeds 39000--39007. BP and DFA search
nine peak learning rates from 0.003 to 1. A crosses
$\{0.03,0.1,0.3\}$ with activity dampings $\{3,30,300\}$;
FD-DFA crosses the same three peaks with decorrelator adaptation rates
$\{10^{-6},10^{-5},10^{-4}\}$. Mean final validation CE selects one
setting per method before confirmation. Candidate-level development
scores are archived with the reproduction records. Equal trial counts do not equate coordinate
resolution across these different searches.

FD-DFA adapts the dense forward-decorrelation mechanism of
\citet{ahmad2024correlations}, pinned to official-code revision
\texttt{00cf470}. It does not reproduce that work's complete FA/Adam
protocol. Dense decorrelators precede every linear map, including the input
and classifier maps. During training they center activities, apply the
current matrix, then update it once using the reference gain and
left-multiplied covariance correction. An algebraically equivalent
low-rank multiplication avoids a cubic matrix product. Evaluation freezes
matrices and running means. Direct feedback is injected at the original
post-ReLU activity before the outgoing decorrelator; gradients then use
the local ReLU/BN derivative and decorrelated presynaptic inputs. Independent
local derivative checks verify this credit coordinate.

All decorrelator transformations and updates are charged to the work budget.
They add stored state during training, although the resulting affine
transforms can be folded into the following linear maps for inference.
The state count is therefore not a deployment lower bound.
Selected settings are peak 0.1 for BP, peak 0.03 for DFA, peak 0.03 with
$\rho_A=3$ for A, and peak 0.03 with adaptation rate $10^{-5}$ for FD-DFA.
A and FD-DFA retain boundary selections.

A again improves DFA, while its primary contrast with FD-DFA spans zero
on validation and final test. BP remains stronger. Neither an A advantage
nor equivalence to FD-DFA follows
(Fig.~\ref{fig:historical_work_width}A, FD-study rows). Table~\ref{tab:fd_test} reports all four methods, with the primary
final-test comparisons collected in Table~\ref{tab:test_primary}.

\begin{table}[!htbp]
\centering\small\setlength{\tabcolsep}{3pt}
\begin{tabular}{@{}lrrr@{}}
\toprule
Rule & Update seconds & Peak GPU MB & Extra decorrelator state (M) \\
\midrule
BP & $30.00117\pm0.00066$ & 111.14 & 0.00000 \\
DFA & $30.00085\pm0.00059$ & 110.88 & 0.00000 \\
A & $30.00157\pm0.00123$ & 135.26 & 0.00000 \\
FD-DFA & $30.00175\pm0.00131$ & 348.92 & 10.75251 \\
\bottomrule
\end{tabular} \space
\caption{Measured work, allocated GPU memory, and additional stored state. Every method has the same supervised parameter count. Decorrelators can be folded into linear maps at inference; their stored state is not a deployment lower bound. MB uses decimal units.}
\label{tab:fd_costs}
\end{table}
\begin{table}[!htbp]
\centering\small\setlength{\tabcolsep}{3pt}
\begin{tabular}{@{}rrlrr@{}}
\toprule
Train & Params (M) & Rule & Test accuracy (\%) & Test CE \\
\midrule
48k & 3.68 & BP & $63.68\pm0.33$ & $1.0313\pm0.0040$ \\
48k & 3.68 & DFA & $60.10\pm0.20$ & $1.1362\pm0.0043$ \\
48k & 3.68 & A & $61.48\pm0.25$ & $1.1000\pm0.0058$ \\
48k & 3.68 & FD-DFA & $61.56\pm0.55$ & $1.0938\pm0.0127$ \\
\bottomrule
\end{tabular} \space
\caption{Forward-decorrelation test endpoints. Mean $\pm$ sample SD across eight paired seeds at the measured-work endpoint.}
\label{tab:fd_test}
\end{table}

\begin{table}[!htbp]
\centering\small\setlength{\tabcolsep}{3pt}
\begin{tabular}{@{}llrr@{}}
\toprule
Cohort & Primary accuracy contrast & Exact $p$ & Holm $p$ \\
\midrule
optimizer & A$-$DFA & 0.0078125 & 0.0156250 \\
forward & A$-$FD-DFA & 0.7343750 & 0.7343750 \\
\bottomrule
\end{tabular} \space
\caption{Both prespecified final-test accuracy comparisons. Holm adjustment covers this fixed family of two exact sign-flip tests. Secondary results do not replace either primary comparison.}
\label{tab:test_primary}
\end{table}

\FloatBarrier
\begin{samepage}
\section{Targeted tests of work, damping and normalization}
\label{app:revision_followups}

The preceding results leave three specific questions: whether activity gains
persist with longer measured work, whether error damping should change with
width, and how normalization affects predictive stability. We specified these
follow-ups before execution, retaining every planned case. They use the same
48,000/2,000 CIFAR-10 training/validation split, paired augmentation, per-example
error moments and raw-update norm matching as the shared-BN studies. Training uses an H100 with float32 tensors and TF32 disabled. No test endpoint or intermediate checkpoint
selects a setting. All 202 cases finish; 32 are validation-only development
runs, and 170 final models each receive one saved test evaluation. Independent
float64 log-sum-exp reproduces their recorded prediction metrics. These are
follow-up analyses with five paired seeds per condition, not additions to an
earlier confirmatory family. Individual $t$ intervals assume approximately
normal seed differences and do not provide simultaneous coverage.
\par
\end{samepage}

\subsection{Transfer to longer measured work}
\begin{table}[!htbp]
\centering\small
\begin{tabular}{lrrr}
\toprule
Rule & 30 s & 60 s & 120 s\\
\midrule
BP & $63.68\pm0.33$ & $65.82\pm0.40$ & $67.54\pm0.19$\\
DFA & $60.10\pm0.20$ & $62.18\pm0.17$ & $64.67\pm0.21$\\
A & $61.48\pm0.25$ & $64.21\pm0.39$ & $66.92\pm0.30$\\
FD-DFA & $61.56\pm0.55$ & $64.45\pm0.81$ & $66.88\pm0.52$\\
\bottomrule
\end{tabular}
\caption{Test accuracy (\%, mean $\pm$ sample SD) at matched measured update
work in the 3.68M-parameter CIFAR-10 MLP. The 30-second FD study uses eight
paired seeds; the longer budgets share five new paired seeds and transfer
the validation-selected settings with separately stretched schedules.
They are separate training runs, not checkpoints of one trajectory.
Timing excludes validation and export.
Appendices~\ref{app:matched_work} and~\ref{app:revision_followups} give complete
cost, selection and endpoint records.}
\label{tab:historical_work_budgets}
\end{table}

The BP/DFA/A/FD-DFA comparison extends the earlier work budget to 60 and 120
synchronized update seconds. Both budgets reuse the same five new paired
seeds (61000--61004) in separate runs. We transfer the
previously selected settings unchanged and stretch the same warmup/cosine
schedule. This tests transfer across budgets, rather than a newly tuned
convergence envelope. The timing definition remains unchanged: sampling,
augmentation, conditioning and optimizer work count; validation and export do
not. Table~\ref{tab:historical_work_budgets} gives all absolute test accuracies; the
paired contrasts appear in Fig.~\ref{fig:historical_work_width}B.

Activity conditioning improves raw DFA at both longer budgets. Its observed
gap to BP narrows as the budget increases, although BP remains ahead; forward
decorrelation remains competitive. The paired A--FD intervals span zero at
both budgets, establishing neither superiority nor equivalence. These results
extend the activity gain beyond the original short budget without establishing
converged performance or transfer beyond this architecture and dataset.

\subsection{Width-specific error damping and a diagonal control}
Full and diagonal error conditioning receive the same validation grid:
relative damping $\{3,30,300,3000\}$ at each hidden width, evaluated over two
development seeds. Final validation CE selects each operator independently;
both development runs must be finite, and ties favor larger damping. Full E
selects 3000 at both widths, the upper boundary; diagonal E selects 30 and 3,
respectively. Five fresh seeds then compare DFA, A, E, diagonal E and K at
5,000 SGD updates, with learning rate 0.1, BN and feedback scale 0.1. A uses
relative damping 30; K adds the selected full-E factor without joint search.

Activity gains persist at both widths, while the selected E and diagonal-E
rules remain close to DFA, and K remains close to A
(Table~\ref{tab:width_retuning}; Fig.~\ref{fig:historical_work_width}D).
Sparse checkpoint spectra explain what this selected operator does:
strong damping makes full E nearly scalar. At the final update its cosine
with the raw update exceeds 0.999 in every recorded layer and seed; its damped
condition number stays below 1.10. Thus the retuned study largely suppresses
the error transformation rather than establishing a robust error-side gain.
It neither proves equivalence nor excludes a useful error factor elsewhere.

For an error moment of exact rank $r$ in dimension $d$, the ridge
$\rho\,\operatorname{tr}(C_E)/d$ divided by the mean positive eigenvalue is
$\rho r/d$. Fixed relative damping therefore need not preserve conditioning
when width changes and rank is batch-limited. The recorded spectra,
participation ranks and update angles make this issue observable, but do not
establish it as the sole cause of the earlier transfer failure.

\begin{table}[!htbp]
\centering\small
\begin{tabular}{lrrrr}
\toprule
& \multicolumn{2}{c}{1024--512} & \multicolumn{2}{c}{2048--1024}\\
Rule & Accuracy (\%) & CE & Accuracy (\%) & CE\\
\midrule
DFA & $51.19\pm1.97$ & 1.363 & $47.21\pm3.19$ & 1.557\\
A & $55.60\pm0.73$ & 1.253 & $53.51\pm2.24$ & 1.353\\
E & $51.34\pm1.62$ & 1.363 & $47.32\pm2.80$ & 1.548\\
Diagonal E & $51.30\pm1.57$ & 1.365 & $47.58\pm3.42$ & 1.527\\
K & $55.58\pm0.92$ & 1.253 & $53.71\pm2.41$ & 1.344\\
\bottomrule
\end{tabular}
\caption{Width-specific damping follow-up. Accuracy is mean $\pm$ sample SD
across five paired seeds; CE is mean test cross-entropy. The same validation
grid selects full and diagonal E independently. K combines A with full E's
choice. No test result selects damping.}
\label{tab:width_retuning}
\end{table}

\begin{samepage}
\subsection{Crossing normalization and feedback scale}
\label{app:bn_scale_intervention}
To isolate two differences between earlier protocols, we cross BN off/on
with feedback scale 0.1/1 for DFA, A, E and K, keeping every other setting
fixed: hidden widths 1024--512, 5,000 SGD updates, learning rate 0.1 and
relative activity/error damping 30. This deliberately fixed recipe is not
separately tuned for each condition. BN restores ordinary predictive-loss
scales for all four rules, and activity conditioning retains an accuracy gain
within the normalized conditions (Table~\ref{tab:bn_scale_followup}). Reducing
feedback scale without BN does not prevent extreme losses. Raw DFA is also
unstable, so this behavior is not unique to inverse conditioning.
\par
\end{samepage}

This intervention identifies a normalization effect under one common recipe;
it does not explain every difference between historical protocols. Finite
losses are retained regardless of their magnitude. The saved histories report
validation loss distributions and logit scales, while sparse checkpoints
record update and weight norms. No failing or unfavorable endpoint is replaced.

\begin{figure}[t]
\centering
\includegraphics[width=\textwidth]{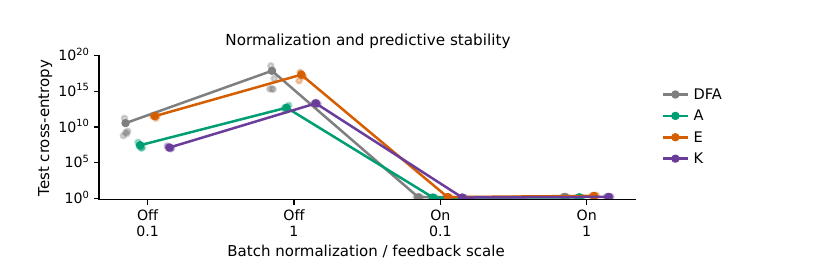}
\caption{\textbf{Normalization restores ordinary predictive-loss scales.}
Test cross-entropy across the fixed-recipe normalization/feedback intervention
(log scale). Faint points show all five seed endpoints per condition;
larger markers and lines show their means. The extreme losses without BN
also occur for raw DFA. All conditions and endpoints were specified before
execution; there is no test-based selection. The work and width follow-ups
appear in Fig.~\ref{fig:historical_work_width}B and D.}
\label{fig:revision_followups}
\end{figure}

\begin{table}[!htbp]
\centering\small
\begin{tabular}{llrrrr}
\toprule
BN & Feedback & DFA & A & E & K\\
\midrule
Off & 0.1 & \shortstack{24.55\\$3.31\times10^{10}$} & \shortstack{36.71\\$2.69\times10^{7}$} & \shortstack{23.93\\$3.20\times10^{11}$} & \shortstack{38.78\\$1.28\times10^{7}$}\\[3pt]
Off & 1 & \shortstack{24.80\\$7.17\times10^{17}$} & \shortstack{38.23\\$4.85\times10^{12}$} & \shortstack{28.53\\$2.03\times10^{17}$} & \shortstack{37.77\\$1.95\times10^{13}$}\\[3pt]
On & 0.1 & \shortstack{51.23\\1.36} & \shortstack{55.72\\1.24} & \shortstack{49.33\\1.42} & \shortstack{56.20\\1.24}\\[3pt]
On & 1 & \shortstack{46.51\\1.55} & \shortstack{53.48\\1.34} & \shortstack{42.06\\2.12} & \shortstack{50.39\\1.55}\\[3pt]
\bottomrule
\end{tabular}
\caption{Normalization/feedback intervention. Each cell gives mean test
accuracy (\%, upper line) and mean CE (lower line) across five paired seeds.
All other settings are fixed. Individual endpoints and predictive-loss
summaries are archived with the reproduction records.}
\label{tab:bn_scale_followup}
\end{table}

\FloatBarrier
\section{Predictive stability under a separate selected protocol}
\label{app:boundary}

To establish the scope of the shared-BN findings, we retain a separate
CIFAR-10 study with different normalization and training choices. It tests
whether the ordering of methods persists under that protocol; it does not
isolate the effect of any one changed choice.

This comparison uses 45,000 training and 5,000 validation images from a
fixed split (seed 15104), training-only channel standardization, and a
3072--1024--512--10 ReLU MLP. Each epoch visits every training identity
once with reflected padding of four pixels, random cropping, and horizontal
flipping. Batch size is 256, with the partial final batch retained.
All methods train for 30 epochs (5,280 updates), using constant-rate SGD
without momentum or weight decay and mean softmax CE. A/E/K have no BN
and use feedback scale 1.0. These and other differences from the shared-BN
studies prevent attributing their different outcomes to one isolated factor.

The ten methods are BP and DFA; each with BN or forward decorrelation;
activity-conditioned BP; and A-, E-, K-nDFA. Conditioned hidden gradients
match their own raw norms. BN precedes ReLU and freezes running statistics
at evaluation. Forward decorrelation uses the same reference mechanism
as above, adapted to this study's credit rules. Classifier and bias updates
remain supervised/local as appropriate. Training uses float32 with
TensorFloat-32 (TF32) disabled and higher-precision reductions as specified
in the code.

\paragraph{Selection and numerical failures.}
Every rule receives twelve initial full-horizon trials and twelve additional
distinct trials in a lower-rate grid declared after the first grid's failures.
This is adaptive validation development, not a single predeclared joint grid.
Raw and BN rules search twelve rates per grid. A and E cross three rates
with four dampings; K crosses three rates with four activity/error pairs;
decorrelation crosses three rates with four adaptation rates. Thus K's
activity damping can differ from A's, and their contrast does not isolate
adding E at fixed A. The complete grids are archived with the reproduction records.
Final validation accuracy selects the setting, with loss and then case
identifier breaking ties; no intermediate or test score enters selection.

Across the completed initial grid and lower-rate grid, 199 trials finish with
finite endpoints and 41 are numerical failures. An interrupted first attempt
at the initial grid had 92 completed runs, 20 numerical failures, two incomplete
cases and six unlaunched cases. The completed rerun resolves the latter eight
as numerical failures; it is this rerun, not the interrupted attempt, that
enters the development comparison. Several selected settings lie on boundaries.
Some conditioned candidates have extreme finite losses despite competitive
accuracy; these remain eligible under the declared accuracy-first rule.
Table~\ref{tab:submission_cifar10_settings} lists every selected setting and
Table~\ref{tab:submission_cifar10} retains all final losses.

Five fresh paired replicates follow selection. Development model, feedback,
order, and augmentation seeds are 15100--15103. Replicate $r=0,\ldots,4$
uses the corresponding seeds $15200+10r$ through $15203+10r$.
All 50 final models are evaluated on the official test set after the
complete cohort is fixed. Reported uncertainty is over training seeds
conditional on the fixed split and selection procedure.

\paragraph{Interpretation.}
A does not improve on raw DFA in this protocol, and E performs worse.
K exceeds A but uses a different activity damping; it cannot establish
the same isolated complementary effect as the main factor studies.
Forward-normalized DFA methods outperform the update conditioners.
Meanwhile, activity conditioning improves BP, demonstrating that its
benefit can also occur with exact gradients. The extreme finite CE values
show that norm matching does not guarantee stable predictive losses.
These findings qualify the successful shared-BN results without identifying
a single cause for their protocol dependence.

\begin{table}[!htbp]
\centering
\small
\setlength{\tabcolsep}{5pt}
\begin{tabular}{lrrrr}
\toprule
Method & Test accuracy (\%) & Mean CE & ms/update & Peak MiB \\
\midrule
BP & $56.83\pm0.71$ & 1.233 & 1.84 & 253 \\
BP + activity & $60.27\pm0.82$ & 1.146 & 4.17 & 268 \\
BP + BatchNorm & $59.65\pm0.77$ & 1.138 & 2.54 & 254 \\
BP + forward decorrelation & $58.35\pm0.34$ & 1.178 & 4.16 & 495 \\
\midrule
DFA & $43.48\pm0.55$ & 1.634 & 1.70 & 253 \\
A-nDFA & $41.01\pm3.42$ & 757.607 & 3.98 & 268 \\
E-nDFA & $28.93\pm2.60$ & $1.18\times10^{14}$ & 4.01 & 268 \\
K-nDFA & $48.30\pm2.02$ & 35.140 & 5.42 & 268 \\
DFA + BatchNorm & $57.15\pm0.53$ & 1.203 & 2.51 & 254 \\
DFA + forward decorrelation & $58.33\pm0.18$ & 1.190 & 3.94 & 495 \\
\bottomrule
\end{tabular}
\caption{\textbf{Full-pool CIFAR-10 confirmation.} A 3072--1024--512--10 ReLU MLP trains for 30 epochs (5,280 updates). Each rule receives 24 distinct development trials across the initial grid and a prospectively declared adaptive lower-rate grid and five fresh paired training seeds. Accuracy is mean $\pm$ sample SD, not SEM; CE is mean cross-entropy. Settings are selected by final validation accuracy, with loss breaking ties. Time is the mean synchronized training-step duration on one H200 GPU, including the update and numerical checks; peak memory is the maximum allocated CUDA memory over the five runs. These measurements exclude allocation setup and validation time and are not an equal-wall-clock comparison. Forward decorrelation adapts the reference mechanism to this study's BP/DFA rules.}
\label{tab:submission_cifar10}
\end{table}
\begin{table}[!htbp]
\centering
\small
\setlength{\tabcolsep}{3pt}
\begin{tabular}{lrrrrrrr}
\toprule
Method & $\eta$ & $\rho_A$ & $\rho_E$ & $\eta_{\rm dec}$ & Val. (\%) & Val. CE & Fail/24 \\
\midrule
BP & 0.096 & -- & -- & -- & 56.80 & 1.230 & 6 \\
BP + activity & 0.1 & 1 & -- & -- & 61.18 & 1.137 & 8 \\
BP + BatchNorm & 0.512 & -- & -- & -- & 58.96 & 1.162 & 0 \\
BP + forward decorrelation & 0.03 & -- & -- & 1e-06 & 57.82 & 1.192 & 12 \\
DFA & 0.0005 & -- & -- & -- & 43.24 & 1.665 & 0 \\
A-nDFA & 0.01 & 1 & -- & -- & 48.24 & 429.522 & 0 \\
E-nDFA & 0.03 & -- & 10 & -- & 33.38 & $1.50\times10^{14}$ & 0 \\
K-nDFA & 0.01 & 3 & 3 & -- & 52.62 & 18.424 & 0 \\
DFA + BatchNorm & 0.016 & -- & -- & -- & 56.72 & 1.221 & 0 \\
DFA + forward decorrelation & 0.003 & -- & -- & 1e-05 & 58.00 & 1.193 & 15 \\
\bottomrule
\end{tabular}
\caption{\textbf{Complete development selection.} Final validation accuracy selects the configuration; ties use lower cross-entropy and then case ID. A dash denotes an inactive parameter. Every numerical failure counts in the 24-trial budget. Relative damping is applied separately as $\max(\rho\,\operatorname{mean}(X^2),10^{-6})$. Values are from the single development seed; they are not confirmation test estimates.}
\label{tab:submission_cifar10_settings}
\end{table}
\begin{table}[!htbp]
\centering
\small
\setlength{\tabcolsep}{3pt}
\begin{tabular}{lrrrr}
\toprule
Contrast & Mean (pp) & Individual 95\% CI & Holm $p_t$ & Exact $p$ \\
\midrule
A-nDFA $-$ DFA & -2.47 & [-6.96, +2.01] & 0.2002 & 0.2500 \\
E-nDFA $-$ DFA & -14.55 & [-18.16, -10.94] & 0.0018 & 0.0625 \\
K-nDFA $-$ A-nDFA & +7.29 & [+2.75, +11.84] & 0.0224 & 0.0625 \\
A-nDFA $-(\mathrm{DFA}+\mathrm{FD})$ & -17.32 & [-21.67, -12.98] & 0.0018 & 0.0625 \\
A-nDFA $-(\mathrm{DFA}+\mathrm{BN})$ & -16.14 & [-20.93, -11.36] & 0.0022 & 0.0625 \\
BP + activity $-$ BP & +3.44 & [+2.82, +4.07] & 0.0006 & 0.0625 \\
\bottomrule
\end{tabular}
\caption{\textbf{All six prespecified paired comparisons.} Intervals and $t$ tests use five paired differences (four degrees of freedom) and an approximate-normality assumption that is difficult to assess at this sample size. Holm correction covers the six $t$ tests; intervals are individual, not simultaneous. The separate two-sided sign-flip test enumerates all 32 assignments and assumes sign exchangeability. Its minimum attainable $p$ is 0.0625 for five nonzero pairs. Constant differences give undefined $t$ statistics and intervals, which are omitted rather than reported as certainty.}
\label{tab:submission_cifar10_contrasts}
\end{table}

\paragraph{Independent predictive-loss audit.}
To distinguish rare extreme errors from broad miscalibration, we recomputed
per-example CE from every saved test logit using float64 log-sum-exp. All
50 endpoint means agree with the original evaluation. Table~\ref{tab:loss_tails}
reports seed-averaged distribution summaries for the raw and conditioned DFA
rules. The large median losses for A and E show that their instability extends
well beyond a small tail. The highest-loss one percent of examples account
for a minority of the total loss. This confirms a predictive failure without
establishing that any one implementation choice caused it.

\begin{table}[!htbp]
\centering\small
\begin{tabular}{lrrr}
\toprule
Rule & Median CE & 99th-percentile CE & Top 1\% loss share (\%)\\
\midrule
DFA & 1.38 & 5.78 & 4.3\\
A & 119.44 & 7179.52 & 12.5\\
E & $4.81\times10^{13}$ & $8.52\times10^{14}$ & 8.3\\
K & 2.69 & 273.07 & 9.7\\
\bottomrule
\end{tabular}
\caption{Loss distributions in the separate full-pool protocol. Each entry
averages the indicated per-seed statistic over all five frozen test endpoints;
quantiles are not computed from pooled predictions. The top-one-percent share
is the fraction of total CE contributed by those examples.}
\label{tab:loss_tails}
\end{table}

\FloatBarrier
\section{Moment orientation and imperfect credit}
\label{app:factor_mechanism}

This experiment separates two questions: how a local moment changes an
approximate update, and whether that change improves its relation to the true
gradient. We construct finite samples with prescribed activity and error
moments while holding their raw cross moment fixed. The targets use the desired
true gradient explicitly. This is an engineered diagnostic of the operator,
not a local procedure for discovering useful credit or unknown noise.

\paragraph{Constructing the moments within a fixed-feedback network.}
The prescribed moments must be compatible with a single network, rather
than assigned to unrelated matrices. The construction below ensures this
while allowing activity and error orientations to vary independently.

Let $A\in\R^{N\times8}$ contain activities and $D\in\R^{N\times8}$ contain
local DFA teaching signals, with $N\ge16$. In this subsection samples
are rows, so the feedback matrix $B$ is transposed relative to
Eq.~\ref{eq:dfa}. For orthogonal bases $Q_L,Q_R\in\R^{8\times8}$, set
\[
 U=Q_LQ_R^\top/\sqrt8,\qquad
 g=Q_L\diag(1,1,1,1,0,0,0,0)Q_R^\top/2.
\]
Both matrices have Frobenius norm one, $U$ has full rank, and
$\cos(g,U)=1/\sqrt2$. Assign
$C_A=Q_R\diag(a)Q_R^\top$ and $C_E=Q_L\diag(e)Q_L^\top$ independently.
Each eigenvalue vector is either isotropic $(8.5,\ldots,8.5)$, signal-low
$(1,1,1,1,16,16,16,16)$, or signal-high with these halves exchanged.
The two anisotropic states have exactly the same spectrum; the isotropic
control preserves their trace. The first four coordinates contain the task
gradient in all orientation cells.

Choose an orthonormal $N\times16$ matrix $[Q_A,Q_Z]$ and construct
\[
 A=\sqrt N Q_A\diag(\sqrt a)Q_R^\top,\qquad
 D=A C_A^{-1}U^\top+\sqrt N Q_Z R^\top,
 \quad RR^\top=C_E-U C_A^{-1}U^\top.
\]
The residual covariance is positive definite. Direct multiplication gives
$A^\top A/N=C_A$, $D^\top D/N=C_E$, and $D^\top A/N=U$.
Changing the error moment changes a teaching component orthogonal to activity,
leaving the raw update unchanged. Activity interventions use the corresponding
compensating construction of $D$.

\begin{samepage}
The trainable matrix is $W\in\R^{8\times8}$, initialized at zero. A fixed
bottleneck readout and its orthogonal complement are
$V=[I;I]/\sqrt2$ and $V_\perp=[I;-I]/\sqrt2$. Use the same fixed feedback
$B=\tfrac12 V+\sqrt{3/4}V_\perp$ in every condition. With
$D_\star=A C_A^{-1}g^\top$, define the initial output residual $E_0$
and squared loss by
\[
 E_0=D_\star V^\top+
       (D-\tfrac12D_\star)V_\perp^\top/\sqrt{3/4},\qquad
 L(W)=\frac{\|E_0+A W^\top V^\top\|_F^2}{2N}.
\]
Then $E_0V=D_\star$ and $E_0B=D$: the actual BP and DFA weight gradients at
zero are precisely $g$ and $U$. The readout and feedback lie in $\R^{16\times8}$
and have full column rank eight,
with $V^\top V=B^\top B=I$ and $V^\top B=I/2$ throughout. The exact Hessian
acts as $\Delta W\mapsto\Delta W C_A$; its left factor is $I$, independently
of the assigned $C_E$. Any error-side benefit here therefore concerns the
direction of imperfect credit, rather than identification of a Hessian factor.
\par
\end{samepage}

\paragraph{Predicted direction changes.}
Because the raw update and true gradient are fixed at initialization,
any change in their alignment comes from conditioning. The shared
eigenvector bases make that change explicit.

\begin{figure}[t]
    \centering
    \includegraphics[width=0.98\textwidth]{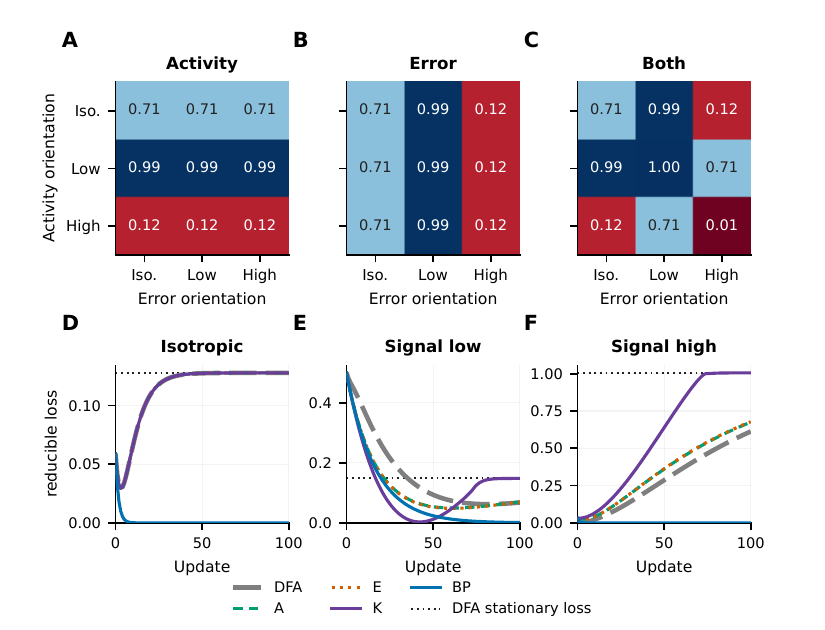}
    \caption{\textbf{Conditioning changes direction without repairing credit.}
    \textbf{A--C,} True-gradient cosines for A, E and K with independently assigned
    moment orientations and damping $1/1$; the raw cosine is $0.707$ throughout.
    Colors use the common cosine range $[0,1]$; labels round to two decimals.
    The displayed $1.00$ is $0.999904$, not exact credit recovery.
    Iso., low and high denote isotropic, signal-low and signal-high moments.
    \textbf{D--F,} Prescribed 100-update trajectories with both moments isotropic,
    signal-low, or signal-high. All methods start from the same weights within
    each cell; the loss shown excludes the constant readout-complement
    component. Initial direction improvement can accelerate approach to a
    suboptimal DFA stationary point in this diagnostic with a fixed bottleneck readout.
    With isotropic moments, the four local trajectories coincide and finish
    above their initial loss.
    These are engineered targets and a fixed
    step size, with all moment/damping cells available in the reproduction records.}
    \label{fig:factor_mechanism}
\end{figure}

Raw, A-, E-, and K-nDFA share each cell's samples and targets. Each conditioned
matrix is matched to $\|U\|_F$. For independently crossed
$\lambda_A,\lambda_E\in\{0.1,1,16\}$, let $w_i$ be the product of the active
reciprocal factors $(a_i+\lambda_A)^{-1}$ and $(e_i+\lambda_E)^{-1}$.
The predicted true-gradient cosine is
\[
 \cos(g,\widetilde U)=
 \frac{\sum_{i=1}^4w_i}{\sqrt{4\sum_{i=1}^8w_i^2}}.
\]
At equal damping, a signal-low factor improves the direction, while
exchanging its eigenspaces worsens it. Two favorable factors reinforce
one another; opposing orientations cancel at equal damping. Thus the
spectrum alone does not determine the sign of a benefit, even with exact
moments (Fig.~\ref{fig:factor_mechanism}A--C).

Eight orthogonal basis seeds and $N\in\{32,64,128\}$ verify the construction
across all nine orientation pairs and nine damping pairs. These realizations
check equivariance and moment identities; they are not independent tasks.
All conditions and direct numerical checks are archived with the reproduction records.

\paragraph{Credit quality and the rotation boundary.}
An update that points partly along the true gradient can still be turned
away from it by conditioning. We now bound how large a rotation is safe,
given a lower bound on the raw update's alignment.

For vectorized $u\ne0$, a symmetric positive-definite (SPD) conditioner
$P$, and its positively norm-matched direction $v\propto Pu$, let
$\theta=\angle(u,v)\in[0,\pi/2)$ and assume
$\cos(g,u)\ge c>0$. Decomposing the unit vectors parallel and perpendicular
to $u$ gives the sharp cone bound
\begin{equation}
 \cos(g,v)\ge c\cos\theta-\sqrt{1-c^2}\sin\theta.
 \label{eq:credit_rotation_bound}
\end{equation}
The lower bound is attained by placing the perpendicular component of $g$
opposite that of $v$. Every gradient in this cone therefore gives descent
along $-v$ exactly when $c>\sin\theta$; equality allows zero derivative.
This is an elementary application of classical operator-angle geometry
\citep{gustafson2010minmax}, not a new angle theorem. For an SPD factor $P$,
$\sin\angle(u,Pu)\le(\kappa(P)-1)/(\kappa(P)+1)$. The vectorized K-nDFA
factor has the product of the two factor condition numbers. A true-gradient
margin is an additional assumption: local covariance estimates do not certify it.

We check the boundary by fixing the signal-low/signal-low construction and
varying the target gradient within this cone. Choosing its perpendicular
component opposite to the conditioned update gives ascent, zero derivative
and descent below, at and above the boundary, while raw DFA remains a descent
direction. Direct loss calculations also verify
\[
 L(-\eta\widetilde U)-L(0)=
 -\eta\langle g,\widetilde U\rangle+
 \frac{\eta^2}{2}\tr(\widetilde U C_A\widetilde U^\top).
\]
Thus a directional guarantee alone does not certify an arbitrary finite
step: at zero derivative, the quadratic term raises the loss.

\paragraph{Short trajectories and preserved stationary mismatch.}
The one-step calculation leaves open what happens over repeated updates.
We therefore follow each constructed network from the same initial weights
within each condition. Figure~\ref{fig:factor_mechanism}D--F displays the
isotropic, signal-low, and signal-high cases. In the isotropic case the
local-rule trajectories coincide; the anisotropic cases show that favorable
initial alignment need not yield a better endpoint.

All 81 orientation/damping cells receive 100 full-batch updates at the
predeclared learning rate $0.03$, with seed 9151500 and $N=64$. Raw DFA, A,
E, K, and BP give 405 trajectories. A/E/K recompute the current error moment
and match their current raw-DFA norm at every step; BP uses its exact gradient.
There are no numerical failures. This is one diagnostic step size, without
optimizer tuning or selection of intermediate checkpoints.

The true and approximate gradients obey
\[
 g(W)=g(0)+W C_A,\qquad U(W)=U(0)+\tfrac12 W C_A.
\]
Their stationary weights differ:
$W_{\rm BP}=-g(0)C_A^{-1}$ and $W_{\rm DFA}=-2U(0)C_A^{-1}$.
Invertible conditioning preserves the zero set of the raw update and cannot
remove this mismatch. The loss in the readout complement remains constant;
we compare the reducible component $\|E(W)V\|_F^2/(2N)$ within each cell.
Initial absolute losses are not comparable measures of difficulty across the
constructed targets.

\paragraph{Scope of the stationary mismatch.}
The mismatch depends on information lost through the fixed readout.
To make that restriction explicit, we compare this construction with a
linear network whose feedback and readout preserve all output directions.

Full column rank here does not make credit injective in the 16-dimensional
output space: $V^\top$ and $B^\top$ each have an eight-dimensional kernel.
The fixed readout cannot change $E(W)V_\perp=E_0V_\perp$, whereas this
residual contributes to DFA because $V_\perp^\top B=\sqrt{3/4}I$.
The mismatch therefore relies on a fixed bottleneck readout and its
unlearnable output complement. For comparison, consider a linear fixed-readout
model with activity dimension $d$, hidden width $h$, and $m\le h$ outputs.
Define the residual--activity
cross moment $R(W)=E(W)^\top A/N\in\R^{m\times d}$. If both
$V^\top,B^\top\in\R^{h\times m}$ have full column rank $m$, then
\[
 g(W)=V^\top R(W)=0
 \quad\Longleftrightarrow\quad R(W)=0
 \quad\Longleftrightarrow\quad U(W)=B^\top R(W)=0.
\]
This condition concerns the cross moment, not necessarily zero sample
residual. Thus the present stationary mismatch is not a general property of
full-rank linear DFA. Neither this counterexample nor the injectivity argument
establishes the stationary structure of nonlinear networks with
activation-dependent gates or trained readouts; their learning behavior
requires separate experiments.

With both signal-low factors, K initially points closest to the true
gradient among the local rules, but approaches the mismatched stationary
point faster and has a higher final reducible loss in the displayed
trajectory. An improved initial direction therefore need not imply a better
endpoint (Fig.~\ref{fig:factor_mechanism}E). The complete orientation/damping
grid remains in the reproduction records.

\FloatBarrier
\section{Connections to other credit-assignment rules}
\label{app:related_expanded}
Conditioning changes how a credit signal is used while retaining DFA's
fixed random feedback. It is complementary to methods that change the
source or routing of credit.

\paragraph{Feedback pathways.}
FA/DFA performance depends on architecture and task
\citep{lillicrap2016feedback,nokland2016dfa,launay2020dfa,bartunov2018scalability}.
The align-then-memorize analysis concerns acquisition of useful feedback
\citep{refinetti2021align}, whereas our spectral calculation assumes alignment.
Weight mirrors and feedback matching adapt feedback directions
\citep{akrout2019weighttransport,lee2026feedbackmatching,moskovitz2018feedbackconv}.
Low-rank parameterizations and orthogonalization address related limitations
\citep{roy2026lowrankfa,boeshertz2026rankcollapse}. Invertible conditioning
preserves algebraic rank but changes singular values and effective rank.

\paragraph{Alternative credit signals.}
Local objectives, target projection, predictive coding, target propagation
and equilibrium propagation change how hidden layers receive credit
\citep{nokland2019local,frenkel2021fixedrandom,whittington2017approximation,
lee2015difference,scellier2017equilibrium}. Perturbation rules estimate credit
from perturbation--error correlations
\citep{williams1992simple,fiete2007model}; NMNC restricts perturbations to an
activity manifold \citep{kang2026nmnc} and is competitive in the exploratory
Fashion-MNIST sweep. Conditioning does not resolve every limitation of
approximate credit.

\paragraph{Curvature and locality.}
Natural-gradient and Kronecker-factored methods motivate the matrix form
\citep{amari1998natural,martens2015kfac,grosse2016convkfac,george2018ekfac}.
Forward decorrelation changes representations
\citep{desjardins2015naturalnn,huang2018decorrelated,huang2019iternorm}
and can help perturbation learning and BP
\citep{dalm2023nodeperturbation,dalm2024decorrbp,ahmad2024correlations}.
nDFA instead conditions updates. Our implementation requires layer-wide
statistics and solves; synaptic locality remains an open implementation question.